\documentclass[preprint,12pt]{elsarticle}

\makeatletter
\def\ps@pprintTitle{%
    \let\@oddhead\@empty
    \let\@evenhead\@empty
    \def\@oddfoot{\footnotesize\itshape
         {}}
    \let\@evenfoot\@oddfoot
    }
\makeatother

\usepackage{hyperref}
\usepackage{amsmath,amssymb,mathtools,float,times,bm,proof,tabularx,capt-of,natbib,siunitx}
\usepackage{pifont}
\usepackage{graphicx,amsfonts,array,url,algorithm,algorithmic,booktabs,hyperref}
\usepackage[flushleft]{threeparttable}
\usepackage[top=1in,bottom=1in,right=1in,left=1in]{geometry}
\usepackage{setspace,multirow,tikz,tikz-cd}
\usepackage{subcaption}
\usepackage{rotating}
\usepackage[section]{placeins}
\usetikzlibrary{calc, positioning}
\usepackage[bottom]{footmisc}
\usepackage{adjustbox}
\usepackage{comment}
\let\Algorithm\algorithm
\renewcommand\algorithm[1][]{\Algorithm[#1]\setstretch{1}}

\graphicspath {{figures/}}

\usepackage[utf8]{inputenc}
\usepackage[english]{babel}
\usepackage[frozencache=true,cachedir=minted-cache]{minted} 
\begin{document}

\begin{frontmatter}

\title{Imbalanced Big Data Oversampling: Taxonomy, Algorithms, Software, Guidelines and Future Directions }

\author[a]{William~C.~Sleeman~IV}\ead{wcsleeman@vcu.edu}
\author[a]{Bartosz Krawczyk}\ead{bkrawczyk@vcu.edu}
\address[a]{Department of Computer Science, Virginia Commonwealth University, Richmond, VA, USA}


\begin{abstract}
Learning from imbalanced data is among the most challenging areas in contemporary machine learning. This becomes even more difficult when considered the context of big data that calls for dedicated architectures capable of high-performance processing. Apache Spark is a highly efficient and popular architecture, but it poses specific challenges for algorithms to be implemented for it. While oversampling algorithms are an effective way for handling class imbalance, they have not been designed for distributed environments. In this paper, we propose a holistic look on oversampling algorithms for imbalanced big data. We discuss the taxonomy of oversampling algorithms and their mechanisms used to handle skewed class distributions. We introduce a Spark library with 14 state-of-the-art oversampling algorithms implemented and evaluate their efficacy via extensive experimental study. Using binary and multi-class massive data sets, we analyze the effectiveness of oversampling algorithms and their relationships with different types of classifiers. We evaluate the trade-off between accuracy and time complexity of oversampling algorithms, as well as their scalability when increasing the size of data. This allows us to gain insight into the usefulness of specific components of oversampling algorithms for big data, as well as formulate guidelines and recommendations for designing future resampling approaches for massive imbalanced data. Our library can be downloaded from https://github.com/fsleeman/spark-class-balancing.git.
\end{abstract}


\end{frontmatter}

\bigskip

\section{Introduction}
\label{sec:int}

Learning from imbalanced data poses significant challenge to most of machine learning algorithms. Standard classifiers were designed to learn from roughly balanced class distributions. Therefore, if one of the classes becomes a predominant one (i.e., majority class), the decision boundary becomes biased towards it. This is rooted in loss functions that assumes uniform costs among all instances. Thus, the more frequent and easier majority class will dominate the training procedure, leading to a poor performance on the minority class.  This learning difficulty can be further augmented by various instance-level characteristics, such as overlapping distributions, borderline examples, small disjuncts, high dimensionality \cite{Korycki:2021} or noisy class labels. When dealing with more than two classes, relationships among them become much more complex, leading to even more challenging learning scenarios. Despite over two decades of progress in this field, imbalanced data problems are as vital as ever. The prevalence in emerging real-world applications, as well as novel problems accompanying class imbalance, such as streaming nature of data, concept drift, or complex data representations, has contributed to challenges with imbalanced data.  The emergence of the deep learning paradigm has only boosted the need for effective ways of handling skewed distributions, while introducing novel challenges, such as long-tailed recognition, or the need for skew-insensitive generative models. 

Class imbalance, while challenging on its own, is often accompanied by big data volume. Having skewed and massive datasets calls for dedicated high-performing architectures, such as Apache Spark. While highly efficient, these computing paradigms require specific ways of implementing algorithms that will harness their full distributed processing power. Oversampling methods are abundant for standard imbalanced data sets, but lack effective implementations for big data problems.

\smallskip
\noindent \textbf{Research goal.} We propose the first, self-contained and holistic work on oversampling for learning from imbalanced big data that encapsulates all the crucial aspects of this domain: (i) taxonomy and details of modern oversampling methods applicable to massive datasets; (ii) software package encompassing the most efficient oversampling algorithms implemented for MapReduce and Apache Spark environments; (iii) a thorough and detailed experimental study on both binary and multi-class imbalanced big data; and (iv) guidelines and future directions for developing novel oversampling algorithms for imbalanced big data. 

\smallskip
\noindent \textbf{Motivation.} While there is a plethora of oversampling algorithms proposed for single-CPU environments, most of them cannot be directly applied to high-performance computing environments, such as Spark. Distributed architecture of such systems poses unique challenges for the design of oversampling algorithms and require specific solutions tuned to work in the MapReduce environment. There is a need to identify which specific components of oversampling that can be easily transferred to a distributed computing setup and how to do this in an efficient manner. Until this point, there were no dedicated software packages that offered any oversampling algorithms more advanced than SMOTE for MapReduce systems. Finally, there is a need to propose an unified taxonomy for imbalanced big data oversampling, as well as outline the directions for future research and challenges awaiting researchers in this important domain.

\smallskip
\noindent \textbf{Summary.} We propose a complete and holistic work on the oversampling of big imbalanced data. We merge both the theoretical, survey-style input with practical, software and empirically oriented contributions. We propose a thorough taxonomy of existing oversampling methods that is designed to identify the most suitable methods that can work in the big data context. Then, we discuss in detail the adaptation of 14 diverse oversampling algorithms to high-performance computing environments, highlighting their useful properties and implementation difficulties. This is followed up with a complete software package that can be used by any researcher. We extend this by a thorough experimental analysis of those oversampling algorithms on 26 binary and multi-class imbalanced big data benchmarks. Finally, we offer design guidelines for future researches on how to create efficient oversampling methods for distributed environments, as well as discuss open challenges and future directions for learning from imbalanced big data.

\smallskip
\noindent \textbf{Main contributions.} This works addresses the discussed challenges in learning from imbalanced big data with the following research contributions:

\begin{itemize}
\item \textbf{Taxonomy of oversampling algorithms.} We propose the first detailed taxonomy of oversampling algorithms that analyzes the main groups of methods based on their instance generation mechanisms and how they utilize inner- and intra-class information.
\item \textbf{Adaptation of popular oversampling algorithms to MapReduce and Spark environments.} We show a detailed way of adapting 14 popular and effective oversampling algorithms to the Spark environment in a way that leverages their usefulness and captures the unique demands of high-performance computing environments. 
\item \textbf{Software package for big data oversampling.} We propose the first complete software package written in the Scala language that offers optimized and efficient implementations of 14 oversampling methods for imbalanced big data. This will increase the reproducibility of future works in this domain and serve as a baseline for researches that will develop their novel algorithms. 
\item \textbf{In-depth experimental study.} We present a thorough experimental study on 26 imbalanced big data benchmarks that capture both binary and multi-class problems. We investigate performance of oversampling techniques based on 4 popular performance metrics, their relationships with different types of data and base classifiers, as well as accuracy/computational complexity trade-offs. 
\item \textbf{Design guidelines for imbalanced big data oversampling.} Based on our experiences with implementing various oversampling methods for high-performance distributed environments, we propose a set of guidelines for researches on what should be taken into account when developing novel oversampling methods and what mechanisms works favorably within the MapReduce framework. 
\item \textbf{Open challenges and future directions for imbalanced big data oversampling.} We outline perspectives and challenges that should be addressed by the research community while designing novel oversampling algorithms for imbalanced big data. 
\end{itemize}

\section{Learning from imbalanced data}
\label{sec:imb}

\subsection{Binary imbalanced problems}
\label{sec:bin}

The strategies for dealing with data imbalance can be divided into two categories. First, the data-level methods: algorithms that perform data preprocessing with the aim of reducing the imbalance ratio, either by decreasing the number of majority observations (undersampling) or increasing the number of minority observations (oversampling). After applying such preprocessing, the transformed data can be later classified using traditional learning algorithms.

By far, the most prevalent data-level approach is the SMOTE \cite{smote} algorithm. It is a guided oversampling technique, in which synthetic minority observations are being created by the interpolation of existing instances. Today it is considered a cornerstone for the majority of the following oversampling methods \cite{smote-variants}. However, due to the underlying assumption about the homogeneity of the clusters of minority observations, SMOTE can inappropriately alter the class distribution when factors such as disjoint data distributions, noise, and outliers are present. Numerous modifications of the original SMOTE algorithm have been proposed in the literature. The most notable include Borderline SMOTE \cite{borderlineSMOTE}, which focuses on the process of synthetic observation generation around the instances close to the decision border; Safe-level SMOTE~\cite{safe_level_smote} and ADASYN \cite{adasyn} which prioritize the difficult instances.

The second category of methods for dealing with data imbalance consists of algorithm-level solutions. These techniques alter the traditional learning algorithms to eliminate the shortcomings they display when applied to imbalanced data problems. Notable examples of algorithm-level solutions include: kernel functions \cite{Raghuwanshi:2020}, splitting criteria in decision trees \cite{Li:2018}, and modifications of the underlying loss function to make it cost-sensitive \cite{Khan:2018}. However, contrary to the data-level approaches, algorithm-level solutions necessitate a choice of a specific classifier. Still, in many cases they are reported to lead to a better performance than sampling approaches \cite{Fernandez:2018}.

\subsection{Multi-class imbalanced problems}
\label{sec:mci}

While performing binary classification, one can easily define the majority and the minority class, as well as quantify the degree of imbalance between the classes. This relationship becomes more convoluted when transferring to the multi-class setting. One of the earlier proposals for the taxonomy of multi-class problems used either: the concept of multi-minority, a single majority class accompanied by multiple minority classes; or multi-majority, a single minority class accompanied by multiple majority classes \cite{Wang:2012}. However in practice, the relationship between the classes tends to be more complicated, and a single class can act as a majority towards some, a minority towards others, and have a similar number of observations to the rest of the classes. Such situations are not well-encompassed by the current taxonomies. Since categorizations such as the one proposed by Napierała and Stefanowski \cite{Napierala:2016} played an essential role in the development of specialized strategies for dealing with data imbalance in the binary setting, the lack of a comparable alternative for the multi-class setting can be seen as a limiting factor for the further research.

The difficulties associated with the imbalanced data classification are also further pronounced in the multi-class setting, where each additional class increases the complexity of the classification problem. This includes the problem of overlapping data distributions, where multiple classes can simultaneously overlap a particular region, and the presence of noise and outliers, where on one hand a single outlier can affect class boundaries of several classes at once, and on the other can cease to be an outlier where some of the classes are excluded. Finally, any data-level observation generation or removal must be done by a careful analysis of how action on a single class influences different types of observations in remaining classes. This leads to a conclusion that algorithms designed explicitly to handle the issues associated with multi-class imbalance are required to adequately address the problem.

The existing methods for handling multi-class imbalance can be divided into two categories, binarization solutions, which decompose a multi-class problem into either $M(M-1)/2$ (one-vs-one, OVO), or $M$ (one-vs-all, OVA) binary sub-problems \cite{Fernandez:2013} where each sub-problem can then be handled individually using a selected binary algorithm. An obvious benefit of this approach is the possibility of utilization existing algorithms \cite{Zhang:2018}, however binarization solutions have several significant drawbacks.

Most importantly, they suffer from the loss of information about class relationships. In essence, we either completely exclude the remaining classes in a single step of OVO decomposition or discard the inner-class relations by merging classes into a single majority in OVA decomposition. Furthermore, especially in the case of OVO decomposition associated computational cost can quickly grow with the number of classes and observations, making the approach ill-suited for dealing with the big data. Among the binarization solutions, the recent literature suggests the efficacy of using ensemble methods with OVO decomposition \cite{Zhang:2016}, augmenting it with cost-sensitive learning \cite{Krawczyk:2016ijcnn}, or applying dedicated classifier combination methods \cite{Fernandez:2018}.

The second category of methods consists of ad-hoc solutions: techniques that treat the multi-class problem natively, proposing dedicated solutions for exploiting the complex relationships between the individual classes. Ad-hoc solutions require either a significant modification to the existing algorithms, or exploring an entirely novel approach to overcoming the data imbalance, both on the data and the algorithm level. However, they tend to significantly outperform binarization solutions, offering a promising direction for further research. Most-popular data-level approaches include extensions of the SMOTE algorithm into a multi-class setting \cite{Saez:2016,Zhu:2017}, strategies using feature selection \cite{Fernandez:2017,Lango:2018}, and alternative methods for instance generation by using Mahalanobis distance \cite{Abdi:2016,Yang:2017}. Algorithm-level solutions include decision tree adaptations \cite{Hoens:2012}, cost-sensitive matrix learning \cite{Bernard:2016}, and ensemble solutions utilizing Bagging \cite{Lango:2018,Collell:2018} and Boosting \cite{Wang:2012}.

\subsection{Big imbalanced data}
\label{sec:big}

When facing the challenge of learning from imbalanced big data the mentioned algorithms are not enough. Their bottleneck lies in the fact that they were designed for small-size datasets. In order to make them scalable to big data classification tasks, one needs to combine them with effective high-performance computing architectures. There are two main trends regarding the choice of hardware.

\begin{itemize}
\item \textbf{Graphic Processing Units}. GPUs offer a powerful high-performance computing environment at a fraction of the cost of a traditional cluster \cite{Cano:2018}. While they are characterized by an excellent data-level parallelism and unbeatable degree of control over each aspect of data scheduling and partitioning, they require highly specialized skills to write efficient and optimized code. Additionally, at the current moment they are unable to handle terabyte-level datasets, due to limitations on the embedded memory. In the context of imbalanced data, the main success of GPUs lies in a highly efficient implementation of the popular SMOTE technique. Due to a full control over data partitioning, one can maintain the meaningful creation of artificial instances in given neighborhoods - a feat that is unreachable in most of CPU clusters. GPUs have been successfully applied to skew-insensitive variants of nearest-neighbor classifiers \cite{Cano:2013}, inducting classification rules using genetic programming \cite{Cano:2015}, as well as efficient large-scale discretization of imbalanced data \cite{Cano:2016}. 

 \item \textbf{CPU clusters}. In recent years one can observe a rise of distributed computing cluster architectures, especially ones using the MapReduce methodology \cite{Ramirez-Gallego18} such as Apache Hadoop and Spark. They offer high elasticity, reliability, and scalability in an user-friendly manner. However, this comes at the cost of increased monetary price of hardware and lack of explicit controlling of scheduling that is available in GPUs. While CPU clusters seem like a highly attractive solution to imbalanced data, one must be aware of their potential limitations. A study on effects of different resampling techniques on Random Forest showed that SMOTE achieves a highly unsatisfactory performance in the MapReduce environment \cite{Rio:2014}. This is caused by the lack of control over data partitioning and in turn creating chunks of data with corrupted spatial relationships. When SMOTE is used in each map with such an unreliable neighborhood, it tends to create artificial instances in incorrect regions, leading to increasing overlapping among classes and shifting true class distributions. Random oversampling and undersampling performs significantly better on CPU clusters. Success stories of MapReduce usage for imbalanced big data include efficient and scalable feature selection \cite{Ramirez-Gallego18b}, rule induction \cite{Fernandez:2017sp}, and data resampling \cite{Triguero:2016sp}. 
\end{itemize}

Mentioned techniques are developed for two-class big imbalanced data. To the best our knowledge, there exists no algorithms dedicated specifically to large-scale multi-class imbalanced data, neither for GPUs nor CPU clusters \cite{Fernandez:2018,Krawczyk:2016}, apart from our preliminary work on Spark-based ensembles \cite{Sleeman:2019}.

\section{MapReduce architecture with Spark}
\label{sec:spk}

Data is now being collected at an accelerated rate as it has become a critical part of research, commercial, and government operations. The effective processing of these large datasets also requires new parallel and distributed computational solutions. Google addressed this challenge with the MapReduce model \cite{dean2008mapreduce} which has lead to the popular open source implementations of Apache Hadoop and Spark. In this section, we give an overview of the MapReduce concept, the Apache Spark architecture, and its distributed data management.

\begin{figure}[ht!]
	\centering
	\includegraphics[width=90mm]{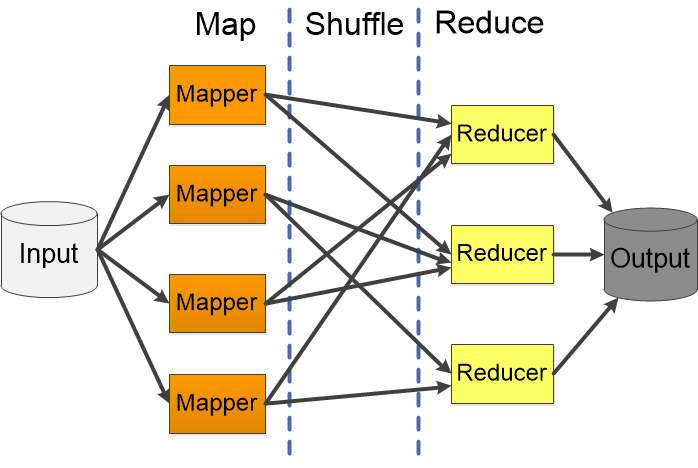}
	\caption{The three phases of the MapReduce model: map, shuffle, reduce.}
	\label{fig:mapreduce}
\end{figure}

The MapReduce distributed computing model uses key/value pairs to distribute data between processing nodes using the three main phases of map, shuffle and reduce as shown in Figure \ref{fig:mapreduce}. In the map phase, data is split by keys and mapped to computing nodes. Intermediate results are then shuffled so that grouped data can be sent to the correct reducer. Finally, the reducer takes incoming intermediate results to compile a final result. In practice, a large number of these map, shuffle, and reduce operations are chained together to build the entire task.

Two popular open source implementations based on the MapReduce concept are Apache Hadoop \cite{Hadoop:2019} and Apache Spark \cite{SparkCluster:2019}. While Hadoop and Spark can use traditional file systems or an object store like Amazon S3, both were designed to use the Hadoop Distributed File System (HDFS) \cite{HDFS}. HDFS provides redundancy, support to handle datasets in the order of terabytes, and distributed data management. The biggest different between these two systems is how they manage data between stages. Spark attempts to keep all intermediate results in main memory while Hadoop has to save the results to disk after one stage and load again for the following stage. Although Spark can be very memory intensive, it can provide superior run time performance with the reduction of disk access.

Spark also uses Directed Acyclical Graphs (DAG) to improve performance. At the beginning of a job, Spark can look at the collection of operations like map, reduce and shuffle to optimize the process. Duplicate operations can be removed, sub-tasks can be reordered, and each step can be performed in a lazy manner in which operations are not started until part of the critical path.

\begin{figure}[ht!]
	\centering
	\includegraphics[width=80mm]{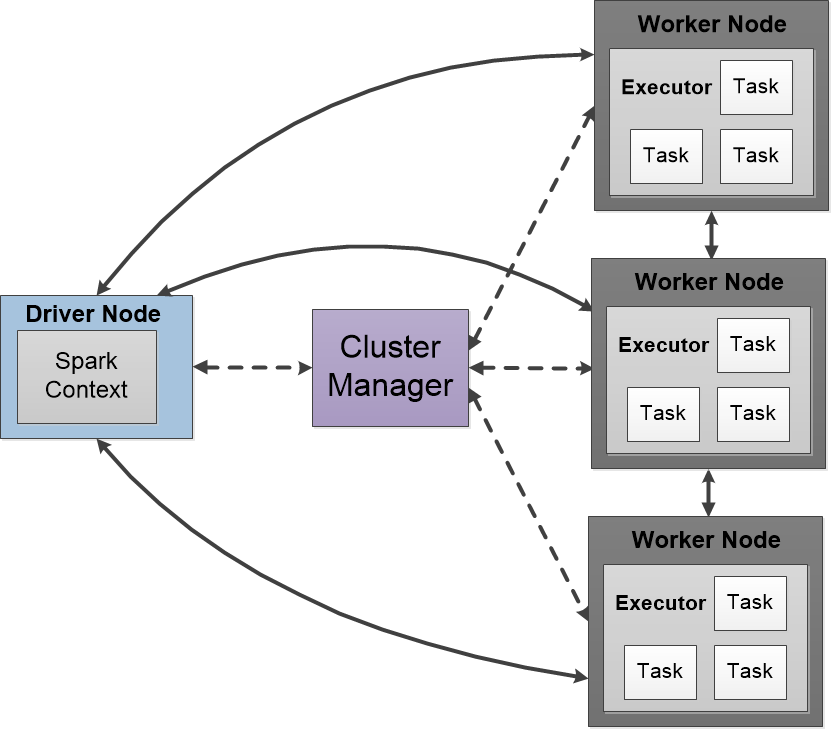}
	\caption{The Spark cluster architecture for resource allocation and data transfer, Resource allocation is shown in dashed lines and data transfer is shown in solid lines.}
	\label{fig:sparkcluster}
\end{figure}

Figure \ref{fig:sparkcluster} shows the architecture of a Spark cluster. In the driver node, a Spark Context object is created which will communicate with the cluster manager, such as Mesos or YARN. When a Spark job starts, the Spark Context object requests resources from the cluster manager which then acquires executors on the worker nodes. Using one or more CPU threads, the executors complete tasks and share partial results with other executors on the same or distributed worker nodes. The driver node sends data to the worker nodes which then send results back to the driver that prepares the final results.

The basic data structure for Spark is the Resilient Distributed Dataset (RDD) \cite{RDD:2019} which provides a layer of abstraction from the parallel operations. RDDs are partitioned so that each portion can be operated on by different worker nodes in parallel. Since Spark manages the parallelism of RDDs under the hood, it allows for the same operations to be performed on different sizes of data and the number of workers without explicitly modifying the code. Spark also includes DataFrames and Datasets which provide the same map, filter, and reduce operations as RDDs but also provides a flexible SQL-like programming interface. 

In addition to the traditional Spark architecture discussed so far, some work has been done to support additional hardware acceleration. One paper \cite{ohno2016accelerating} proposed a Spark modification to support the use of GPUs by invoking CUDA kernels for computationally intensive tasks. Several caching methods were investigated to reduce network overhead, and experiments using both local and remote GPUs showed significant speed improvements. Another project \cite{ghasemi2019accelerating} implemented the \textit{k}-Means algorithm with Field-Programmable Gate Array (FPGA) support. The FPGA access was performed through the executors in such a way that the hardware interface was abstracted from the end user. The HetSpark \cite{hidri2018hetspark} framework extended the Spark architecture to include heterogeneous executors, including both GPU and FPGA hardware support. Some limited support for GPUs has been included in Apache Spark starting with version 3.0.

\section{Oversampling algorithms for imbalanced big data}
\label{sec:ovs}

\subsection{Taxonomy of oversampling algorithms for big data}

\noindent{\textit{Blind vs guided oversampling}}
is the first-level taxonomy of oversampling algorithms that is based on using any data-level information to create artificial instances for minority classes:
\begin{itemize}
\item \textbf{Blind oversampling.} This approach assumes random multiplication of existing instances by either copying them in order to increase the minority class size, or by adding noise / jitter (so-called smearing) to create new artificial instances that are not the exact copies of existing ones. 
\item \textbf{Guided oversampling.} This approach uses various class-level and instance-level properties to generate new artificial instances in a way that increases the size of the minority class. It also decreases the difficulty of learning from this class by making the decision boundary better aligned towards a balanced recognition of all classes.
\end{itemize}

\smallskip
\noindent{\textit{Guided oversampling in details}} is the second-level taxonomy of oversampling algorithms that is based on what and how data-level information is utilized to create artificial instances for minority classes:

\begin{itemize}
\item \textbf{Using global or local information:} 
\begin{itemize}
\item \textbf{Basic local information.} Guided oversampling aims at targeted injection of artificial instances into the minority class in a way that will lower bias, while preserving class characteristics. SMOTE achieved this by incorporating basic information about the class structure by finding $k$ nearest neighbors for each minority instance and injecting artificial samples along hyperplanes between them. This idea was later extended into taking into account the composition of the neighborhood \cite{Sadhukhan:2019} and ratio between minority and majority instances within them \cite{borderlineSMOTE,safe_level_smote}.
\item \textbf{Advanced local information.} With the success of using local information came the observation of non-uniform minority class properties over the feature space. This showed that not only each neighborhood instance may have different importance while using it for oversampling, but also that the size of neighborhood and the resampling ratio should be adapted to each locality \cite{koziarski2019radial}. More advanced local information can be extracted in a form of discrete or continuous taxonomy of minority instance types \cite{Saez:2016}, or by analyzing the strength of each class presence in a given area \cite{Krawczyk:2020}. 
\item \textbf{Basic global information.} An alternative approach postulated that one of the biggest learning difficulties embedded in minority classes may be their potential for being of multi-modal nature \cite{Sharma:2018}. Ignoring this fact would lead to oversampling that increases overlapping between minority and majority classes, thus effectively further enhancing the bias instead of alleviating it. Restricting oversampling to individual modalities (e.g., by clustering or density analysis) will therefore not affect decision boundaries in a negative way. Another type of popular global information includes noise \cite{Saez:2015,Koziarski:2019} and outlier analysis \cite{Koziarski:2020kbs} in order to exclude potentially harmful or incorrect instances from being used as a resampling seed.  
\item \textbf{Advanced global information.} While modalities and data perturbations are important factors, one can gain even deeper insight into minority class structure. This is especially important in the case of multi-class problems, where each individual minority class may have different properties. Therefore, global information can be used to guide the oversampling for each class independently, by taking into account such factors as class compactness, presence of substructures, or substructure size and complexity. 
\item \textbf{Combining local and global information.} Local and global information can be effectively combined, to detect minority class properties and modalities, and then analyze instance-level difficulty within each modality in order to offer locally adaptive resampling that follows global class characteristics \cite{Krawczyk:2020}. 

\end{itemize}

\item \textbf{Using nearest neighbor distances:} 
\begin{itemize}
\item \textbf{Euclidean-based neighborhood.} Most of guided oversampling methods rely on a neighborhood definition that allows for injecting the artificial minority instances. This is usually connected with the $k$-nearest neighborhood that was first utilized by SMOTE and then adapted by other oversampling algorithms. Eucliudean distance is commonly used for vector comparison due to its ease of computation and well-understood properties when dealing with uniform feature types \cite{Tarawneh:2020}. 
\item \textbf{Non-Euclidean neighborhood.} Euclidean distance is subject to various limitations, such as its ineffectiveness in handling mixed feature types or a higher number of dimensions \cite{Pradipta:2021}. Therefore, non-Euclidean alternatives have been used in oversampling, such as Mahalanobis distance (due to its effective properties for modeling more complex manifolds or class boundaries), Hellinger distance \cite{Grzyb:2021}, other $L_k$ norms with $k > 1$ and manifolds \cite{Bellinger:2018,Yang:2019}, kernel density functions \cite{koziarski2019radial}, or metric learning \cite{Feng:2019,Zeng:2019}.
\end{itemize}

\item \textbf {Inter-class vs intra-class relationships:} 
\begin{itemize}
\item \textbf{Inter-class relationships.} This family of methods focuses on extracting information only from the minority class itself. This can either include local or global information, aiming at enriching the minority class in such a way that will empower its presence in the decision space while alleviating its potential difficulties \cite{Liu:2018}, such as noise or the presence of atypical observations \cite{Skryjomski:2017}.  
\item \textbf{Intra-class relationships.} Using only inter-class properties can lead to improper performance of oversampling methods, as while improving the class itself we may harm the remaining classes. To avoid this, modern oversampling methods take into account not only the target minority class itself, but also the remaining classes from the dataset \cite{Krawczyk:2020}. This helps to define class margins or areas with high probability/potential of belonging to a certain class \cite{Koziarski:2019}, allowing for the injection of artificial instances so that the overlapping between classes \cite{Vuttipittayamongkol:2020} will not increase, offering a more balanced performance on all classes instead of replacing one bias with another.  
\end{itemize}

\item \textbf{Oversampling ratio:} 
\begin{itemize}
\item \textbf{Fixed oversampling ratio.} Most of oversampling methods take as an user input how many artificial instances for the minority class should be created (i.e., oversampling ratio). Most studies show that oversampling should aim at balancing minority and majority classes \cite{Rio:2015}. However, in the case of extreme class imbalance the combination of oversampling with majority class undersampling leads to better results \cite{Koziarski:2017}. Other resampling methods learn the oversampling ratio on their own, fixing it for each dataset independently \cite{Li:2021}. However, they still use this fixed ratio over the entire feature space. 
\item \textbf{Adaptive oversampling ratio.} As minority classes may have multi-modal structure, it may be beneficial to adapt the oversampling ratio to each modality. Some of them may be easier to analyze, while the remaining ones require to be more empowered in order to reduce bias imposed on them by a classifier \cite{Koziarski:2019,Chen:2021}. Furthermore, in multi-class settings, the oversampling ratio could be adapted dynamically not only for within class structures, but also for each class independently \cite{Krawczyk:2020}. Finally, adaptive oversampling ratios are necessary under non-stationary data properties, where the resampling algorithm must be able to constantly adapt how many artificial instances are generated at a given moment \cite{Korycki:2020}. 

\end{itemize}
\end{itemize}

\subsection{Details of oversampling algorithms}

In this section, a short description of each sampling method is presented with high level pseudo code. These algorithms were implemented as described in their original publications as closely as possible, although some changes were required to work efficiently within the MapReduce programming model. All of these algorithms were originally designed for binary class problems so multi-class supported was added by performing oversampling on all minority classes using a one-vs-all pattern. Algorithm \ref{alg:transform} shows the process of oversampling the minority classes, where the \textbf{ClassBalance*} method is a placeholder for one of the presented sampling methods. 

\begin{algorithm}
{
	\caption{Transform}
	\label{alg:transform}
	\begin{algorithmic}
		\STATE $\textbf{Input:}$ DataFrame of instances that needs to be class balanced
		\STATE $\textbf{Parameters:}$ DF: DataFrame of instances
		\STATE $\textbf{Output:}$ Oversampled DataFrame
		\STATE
		\STATE $\textbf{function:}$ transform($DF$)
		\STATE $classDFs\leftarrow DF$.groupBy$(classLabel)$
		\STATE $majorityClassLabel, majorityClassCount\leftarrow $\par$ \hskip\algorithmicindent max(classExamples$.map$(x\rightarrow x.count))$
		\STATE $examplesToAdd\leftarrow classExamples$.map$(x\rightarrow $\par$ \hskip\algorithmicindent majorityClassCount - x$.count())
		\STATE $oversampledClasses\leftarrow DFs$.map$(x\rightarrow $\par$ \hskip\algorithmicindent 
		\textbf{ClassBalance*}(x, examplesToAdd(x.label)))$
		\RETURN $union(oversampledClassses)$
	\end{algorithmic}
}
\end{algorithm}

\smallskip
\noindent \textbf{ADASYN} \cite{adasyn} begins with \textit{k}NN applied to the entire dataset and then the nearest neighbors for each minority example is returned. An imbalance ratio  of nearest neighbors is calculated for each minority example and then all of these ratios are summed. The number of synthetic examples to be created from each minority example is proportional to the level of its class imbalance. If a minority example has no minority neighbors, simple example replication is performed, otherwise synthetic examples are created between a target example and one of its minority class neighbors. ADASYN was one of the earliest algorithms that attempted to utilize the local instance difficulty factors in order to introduce artificial instances in a more guided manner that would reduce the complexity of a classifier's decision boundary.

\begin{algorithm}
		\caption{ADASYN}
		\label{alg:adasyn}
		\begin{algorithmic}
            \STATE $\textbf{Input:}$ DataFrame of the full dataset
			\STATE $\textbf{Parameters:}$ \textit{DF}: DataFrame of all examples in the dataset,
			\textit{minorityLabel}: label of the minority class, \textit{totalExamplesToAdd}: number of synthetic examples to create, \textit{k}: neighbor count for \textit{k}NN\STATE $\textbf{Output:}$ Synthetic examples
			\STATE
			\STATE $\textbf{function:}$ ADASYN($DF, minorityLabel, totalExamplesToAdd, k$)
			\STATE $minorityDF\leftarrow DF.$filter$(label = minorityLabel) $
			\STATE $knn\leftarrow $ KNN.fit$(DF)$
			\STATE $nearestNeighbors\leftarrow knn.$transform$(minorityDF)$
			
			\STATE $nearestNeighbors[majorityRatio]\leftarrow nearestNeighbors$\par$
			\hskip\algorithmicindent$.map$(x\rightarrow x.neighbors.tail.$filter$(label \neq minorityLabel)$.count())
			\STATE $majorityRatioSum\leftarrow nearestNeighbors[majorityRatio]$.sum()
			
			\STATE $nearestNeighbors[examplesToAdd]\leftarrow nearestNeighbors$\par$
			\hskip\algorithmicindent$.map$(x\rightarrow (x[majorityRatio] $ / $ majorityRatioSum$)\par
			\hskip\algorithmicindent $ * $ $ totalExamplesToAdd)$
			\RETURN $nearestNeighbors$.map$(x\rightarrow$ createSyntheticExamples$(x))$
			\STATE
			
			\STATE $\textbf{function:}$ createSyntheticExamples($example$)
			\RETURN $(0 $ to $ example[examplesToAdd])$.map$(x\rightarrow $createExample($x, k$))
			
			\STATE
			\STATE $\textbf{function:}$ createExample($example, k$)
			\STATE $example\leftarrow example.neighbors$[0]
			\STATE $randomNeighbor\leftarrow example.neighbors.tail[randomInt(0, k)]$
			\RETURN Array$(example, randomNeighbor)$.transpose\par
			\hskip\algorithmicindent .map($x\rightarrow x[0] $ + $ $randomDouble() $ * $ $ (x[1] - x[0])$)
		\end{algorithmic}
\end{algorithm}

\smallskip
\noindent \textbf{ANS} \cite{ans} starts with a 1-NN of the minority class and then counts the number of majority examples within these radii. Minority examples that are found to have many majority examples nearby are then removed. From the remaining minority examples, the maximum distance between nearest neighbors is determined and used for the radius limit for distance based nearest neighbor calculation. From these nearest neighbor results, the number of majority examples close to each minority class example is counted. Each majority neighbor count is divided by the total sum to produce a probability score that is then used for generating SMOTE style synthetic examples. ANS is an extension of SMOTE-based oversampling that adds a data cleaning step in order to reduce the number of irrelevant or noisy instances that will be used as SMOTE input.

\addtolength{\topmargin}{0in}
\resizebox{4in}{!}{
\begin{minipage}{\textwidth}
\begin{algorithm}[H]
	{
		\caption{ANS}
		\label{alg:ans}
		\begin{algorithmic}
			\STATE $\textbf{Input:}$ DataFrame of the full dataset
			\STATE $\textbf{Parameters:}$ \textit{DF}: DataFrame of all examples in the dataset,
			\textit{minorityLabel}: label of the minority class, \textit{totalExamplesToAdd}: number of synthetic examples to create, \textit{k}: neighbor count for \textit{k}NN, \textit{CMaxRatio}: ratio value for outcast detection
			\STATE $\textbf{Output:}$ Synthetic examples
			\STATE
			\STATE $\textbf{function:}$ ANS($DF, minorityLabel, totalExamplesToAdd, k,CMaxRatio$)
			\STATE $minorityDF\leftarrow DF$.filter$(x\rightarrow x.label = minorityLabel) $
			\STATE $majorityDF\leftarrow DF$.filter$(x\rightarrow x.label \neq minorityLabel) $
			\STATE $CMax\leftarrow DF.$count() $ * $ $cMaxRatio$
			
			\STATE $minorityKnn\leftarrow $KNN.fit$(minorityDF, returnDistances = True) $
			\STATE $minorityNeighbors\leftarrow minorityKNN$.transform$(minorityDF)$
			\STATE $minorityNeighbors[closestMinorityDistance]\leftarrow minorityNeighbors$\par\hskip\algorithmicindent.map$(x\rightarrow x.neighbors$.filter$(y\rightarrow y.label = minorityLabel)[0].distance)$
		
			\STATE $outBorderKnn\leftarrow $KNN.fit$(majorityDF)$
			\STATE $outBorderNeighbors\leftarrow outBorderKnn$.transform$(minorityNeighbors)$
			\STATE $outBorderExamples\leftarrow outBorderNeighbors$\par\hskip\algorithmicindent.map$(x\leftarrow x.neighbors$.filter$(closestMinorityDistance < x.radius))$
			
			\STATE $outBorderExamples[outBorder]\leftarrow outBorderExamples$\par\hskip\algorithmicindent.map$(x\rightarrow x.neighbor$.count())
			\STATE $outBorderCounts\leftarrow outBorderExamples.select(outBorder)$
			
			\STATE 
			\STATE $previousOutcasts\leftarrow -1$
			\FOR{$c=1$ to $CMax$}
				\STATE $numOfOutcasts\leftarrow outBorderCounts$.filter$(x\rightarrow x \geq c)$.sum()
				
				\IF {$|numOfOutcasts - previousOutcasts| = 0$}
				\STATE $C\leftarrow c$
					\IF{$outBorderExamples$.filter$(outborder < C)$.count() $ > 0$}
					\STATE
						break
					\ENDIF
				\ENDIF
				\STATE $previousOutcasts\leftarrow numOfOutcasts$
			\ENDFOR
			\STATE
			
			\STATE $Pused\leftarrow outBorderExamples$.filter$(outBorder < C)$
			\STATE $maxClosestMinorityExample\leftarrow Pused$.select$(closestMinorityDistance)$.max() 
			
			\STATE $PusedKnn\leftarrow $KNN.fit$(Pused, $\par\hskip\algorithmicindent$searchRadius = maxClosestMinorityExample) $
			\STATE $PusedDistances\leftarrow PusedKnn$.transform$(Pused)$
			\STATE $Pused[neighborCount]\leftarrow PusedDistance$\par\hskip\algorithmicindent.map$(x\rightarrow x.neighbors$.count())
			\STATE $neighborCountSum\leftarrow Pused[neighborCount]$.sum()
			\STATE $Pused[examplesToAdd]\leftarrow Pused$.map$(x\rightarrow $\par\hskip\algorithmicindent$(x[neighborCount] $ $ / $ $ neighborCountSum) * totalExamplesToAdd$
			
			\RETURN $Pused$.map$(x\rightarrow $generateExamples($x, x[examplesToAdd, k]$))
			\STATE	
			
			\STATE $\textbf{function:}$ generateExamples($example, samplingRatio, k$)
			\RETURN $(0 $ to $ samplingRatio)$\par\hskip\algorithmicindent.map( \underline{\hspace*{0.3cm}} $\rightarrow $ generateSingleExample($example, k$))
			\STATE $ $ 
			
			\STATE $\textbf{function:}$ generateSingleExample($example, k$)
			\STATE $example\leftarrow example.neighbors[0]$
			\STATE $randomNeighbor\leftarrow example.neighbors.tail$[randomInt($0, k$)]
			\RETURN Array($example, randomNeighbor$)\par
			\hskip\algorithmicindent .map($x\rightarrow x[0] $  $ + $ randomDouble()$ $ $ * $ $ (x[1] - x[0]))$
			
		\end{algorithmic}
	}
\end{algorithm}
\end{minipage}
}
\clearpage

\smallskip
\noindent \textbf{Borderline SMOTE} \cite{borderlineSMOTE} performs oversampling from minority class examples that have multiple majority class neighbors. \textit{k}NN is performed and the minority class data is filtered to only include examples that share \textit{m/2} to \textit{m-1} majority neighbors (labeled as \textit{danger}), where \textit{m} is the number of nearest neighbors. \textit{k}NN is performed again, but only on the minority class and the nearest neighbors for the \textit{danger} examples are returned. Each \textit{danger} example is used to create synthetic examples as in standard SMOTE. Borderline SMOTE utilizes the concept of introducing artificial instances in the uncertainty area between classes, thus aggressively countering the bias towards majority class.

\begin{algorithm}[!]
	{
		\caption{Borderline SMOTE}
		\label{alg:borderlinesmote}
		\begin{algorithmic}
            \STATE $\textbf{Input:}$ DataFrame of the full dataset
			\STATE $\textbf{Parameters:}$ \textit{DF}: DataFrame of all examples in the dataset, \textit{minorityLabel}: label of the minority class, \textit{totalExamplesToAdd}: number of synthetic examples to create, \textit{k}: neighbor count for \textit{k}NN\STATE $\textbf{Output:}$ Synthetic examples
			\STATE
			\STATE $\textbf{function:}$ BorderlineSMOTE($DF, minorityLabel, totalExamplesToAdd, k$)
			\STATE $knnModel\leftarrow$ KNN.fit$(DF, k)$
			\STATE $nearestNeigbors\leftarrow knnModel$.transform($minorityDF$)
			\STATE $dangerDF\leftarrow nearestNeighbors$.filter(isDanger($nearestNeighbors[neighbors.labels], k$))
			\STATE $samplingRate\leftarrow totalExamplesToAdd $ $ / $ $ dangerDF$.count()
			\STATE $dangeKnnModel\leftarrow$ KNN.fit$(minorityDF, k)$
			\STATE $dangerNeigbors\leftarrow dangerKnnModel$.transform($dangerDF$)
			\RETURN $(0$ to $examplesToAdd)$\par$
			\hskip\algorithmicindent$.map($x\rightarrow $ generateExample($dangerNeigbors[x],samplingRate$))
			
			\STATE
			\STATE $\textbf{function:}$ isDanger($labels, k$)
			\STATE $label\leftarrow labels[0]$
			\STATE $majorityNeighbors\leftarrow labels.tail$\par$
			\hskip\algorithmicindent$.map$(x\rightarrow if $ $ x = label: 0; $ $ else: 1)$.sum()
			\IF{$k / 2 \leq majorityNeighbors < k$}
			\RETURN $True$
			\ELSE
			\RETURN $False$
			\ENDIF
			\STATE
			\STATE $\textbf{function:}$ generateExamples($dangerExample, samplingRate$)
			\STATE $example\leftarrow dangerExample.neighbors[0]$
			\STATE $randomNeighbors\leftarrow (0 $ to $ samplingRate)$\par
			\hskip\algorithmicindent.map($x\rightarrow example.neighbors.tail[$randomInt$(0, k)]$ 
			\RETURN $randomNeighbors$.map($x\rightarrow $ generateSingleExample$(example, x)$)
			\STATE
			\STATE $\textbf{function:}$ generateSingleExample($example, neighbor$)
			\RETURN Array$(example, neighbor)$.transpose\par\hskip\algorithmicindent.map
			($x\rightarrow x[0] $ + randomDouble() * $ (x[1] $ - $ x[0]))$)
		\end{algorithmic}
	}
\end{algorithm}
\clearpage

\smallskip
\addtolength{\topmargin}{-.5in}
\noindent \textbf{CCR} \cite{ccr} has two phases for class balancing: cleaning the minority class regions from majority class examples and oversampling in the cleaned regions. First, an energy budget is set and a radius is expanded iteratively until the budget has been spent. Majority class examples in that region are pushed out the distance of the radius. Synthetic examples are created in the cleaned minority class regions proportionally to favor original minority class examples within small radii, thus creating more examples in the most difficult regions. CCR combines data cleaning with oversampling, but instead of removing instances it relocates them. That allows for creating safe regions for oversampling without discarding useful information about majority class.

\resizebox{3.75in}{!}{
\begin{minipage}{\textwidth}
\begin{algorithm}[H]
	{
		\caption{CCR}
		\label{alg:ccr}
		\begin{algorithmic}
			\STATE $\textbf{Input:}$ DataFrame of the full dataset
			\STATE $\textbf{Parameters:}$ \textit{DF}: DataFrame of all examples in the dataset,
			\textit{minorityLabel}: label of the minority class, \textit{totalExamplesToAdd}: number of synthetic examples to create, \textit{engeryBudget}: amount of energy to expend when finding the minority class radii
			\STATE $\textbf{Output:}$ Synthetic examples
			\STATE
			\STATE $\textbf{function:}$ CCR($DF, minorityLabel, totalExamplesToAdd, energyBudget$)
			\STATE $minorityDF\leftarrow DF.filter(x\rightarrow x.label = minorityLabel)$				
			\STATE $majorityDF\leftarrow DF.filter(x\rightarrow x.label \neq minorityLabel)$
			\STATE $minorityDF[radius]\leftarrow minorityDF$.map($x\rightarrow $findRadius($x, energyBudget$))
			
			\STATE $movedMajorityExamples\leftarrow minorityDF$\par$
			\hskip\algorithmicindent.$map$(x\rightarrow $moveMajorityExamples$(x, majorityDF))$
			\STATE $uniqueMajorityExamples\leftarrow movedMajorityExamples.$groupBy$(minorityDF.index)$\par$
			\hskip\algorithmicindent.$map$(x\rightarrow x[$randomInt($0, $ $x.count()$)]))
			\STATE $minorityDF[inverseRadius]\leftarrow 1 $ / $ minorityDF[radius]$
			\STATE $inverseRadiusSum\leftarrow minorityDF[inverseRadius]$.sum()
			\RETURN $minorityDF$.map($x\rightarrow $generateExamples($x, inverseRadiusSum, totalExamplesToAdd$))
			\STATE
			\STATE $\textbf{fu \footskip = 0ptnction:}$ generateExamples($example, inverseRadiusSum, totalExamplesToAdd$)
			\STATE $gi\leftarrow (example[inverseRadius] $ / $ inverseRadiusSum) $\par
			\hskip\algorithmicindent $ * $ $ totalExamplesToAdd$

			\STATE $syntheticExamples\leftarrow (0 $ to $ gi)$.map( $\underline{\hspace*{0.3cm}} \rightarrow (0 $ to $ example$.transpose\par
			\hskip\algorithmicindent.map( $x \rightarrow $ randomChoice(-1, 1) $ * $ randomDouble() \par
			\hskip\algorithmicindent $ * $ $example[radius]) + x$)
			\RETURN $syntheticExamples$

			\STATE
			\STATE $\textbf{function:}$ moveMajorityExamples($example, minorityLabel$)
			\STATE $majorityNeighbors\leftarrow example.neighbors.tail$\par
			\hskip\algorithmicindent.filter($label \neq minorityLabel$)
			\STATE $majorityNeighbors[withinRadius]\leftarrow majorityNeighbors$\par
			\hskip\algorithmicindent.map($x\rightarrow $withinRadius$(example, x, example[radius])$)
			\STATE $majorityNeighbors\leftarrow majorityNeighbors$.filter($withinRadius$ = 1)
			\STATE $majorityNeighbors[distance]\leftarrow majorityNeighbors$\par
			\hskip\algorithmicindent.map($x\rightarrow $Array($examples, x)$.transpose.map($x\rightarrow $abs($x[0] - x[1]$))
			\RETURN $majorityNeighbors$.map($x\rightarrow $ shiftMajorityExample($x$)))
			\STATE	
			
			\STATE $\textbf{function:}$ shiftMajorityExample($majorityExample, minorityExample$)
			\STATE $distance\leftarrow majorityExample[distance]$
			\STATE $radius\leftarrow minorityExample[radius]$
			\RETURN Array$(majorityExample, minorityExample)$.transpose.map($x\rightarrow x[0] + ((radius - distance) $ / $ distance) * (x[1] - x[0])$)	
			
			\STATE
			\STATE $\textbf{function:}$ findRadius($example, energyBudget$)
			\STATE $energy\leftarrow energyBudget$
			\STATE $radius\leftarrow 0$
			\WHILE{$energy > 0$}\STATE $movedMajorityExamples\leftarrow minorityWithRadius$\par$
			\hskip\algorithmicindent.$map$(x\rightarrow $moveMajorityExamples$(x, majorityDF))$
			\STATE $uniqueMajorityExamples\leftarrow movedMajorityExamples.$groupBy$(index)$\par$
			\hskip\algorithmicindent.$map$(x\rightarrow $randomChoice$(x))$
			\STATE $deltaR\leftarrow energy $ $ / $ NoP$(example, radius)$
			\IF{NoP$(example, radius + deltaR) > $NoP$(example, radius)$}
			\STATE $deltaR\leftarrow$ distance to nearest majority example outside of radius
			\ENDIF
			\STATE $radius\leftarrow radius + deltaR$
			\STATE $energy\leftarrow engery - deltaR $ $* $ NoP$(example, radius)$
			\ENDWHILE
			\RETURN $radius$
			\STATE
			
			\STATE
			\STATE $\textbf{function:}$ NoP($example, radius, minorityLabel$)\STATE $movedMajorityExamples\leftarrow minorityWithRadius$\par$
			\hskip\algorithmicindent.$map$(x\rightarrow $moveMajorityExamples$(x, majorityDF))$
			\STATE $uniqueMajorityExamples\leftarrow movedMajorityExamples.$groupby$(index)$\par$
			\hskip\algorithmicindent.$map$(x\rightarrow $randomChoice$(x))$
			\STATE $majorityNeighbors\leftarrow examples.neighbors.tail$\par
			\hskip\algorithmicindent.filter($label \neq minorityLabel$)
			\RETURN $majorityNeighbors$.map($x\rightarrow $withinRadius($x$)).sum() + 1
			\STATE
			\STATE $\textbf{function:}$ withinRadius($example, neighbor, radius$)
			\STATE $distance\leftarrow$ Array$(example, neighbor)$.transpose.map$(x\rightarrow (x[0] - x[1]) *(x[0] - x[1]))$.sum()
			\IF{$distance \leq radius$}
			\RETURN 1
			\ELSE
			\RETURN 0
			\ENDIF
			\RETURN $ $
			
		\end{algorithmic}
	}
\end{algorithm}
\end{minipage}
}

\clearpage

\smallskip
\noindent \textbf{Cluster SMOTE} \cite{cluster_SMOTE} is a straightforward method that utilizes clustering when generating synthetic examples. The minority class is clustered using \textit{k}-Means and \textit{k}NN is performed on each cluster. For each synthetic example to be created, a random cluster is selected and then a random example within that cluster is selected. A new synthetic example is then created on a random point between the selected examples and one of its nearest neighbors. Cluster SMOTE was one of the first solutions addressing a major SMOTE drawback -- poor performance on multi-modal data, where artificial instances should not be introduced between modalities. 

\begin{algorithm}
	{
		\caption{Cluster SMOTE}
		\label{alg:clusterSmote}
		\begin{algorithmic}
			\STATE $\textbf{Input:}$ DataFrame of the full dataset
			\STATE $\textbf{Parameters:}$ \textit{DF}: DataFrame of all examples in the dataset,
			\textit{minorityLabel}: label of the minority class, \textit{totalExamplesToAdd}: number of synthetic examples to create, \textit{k}: number of nearest neighbors to consider, \textit{clusterK}: number of clusters to create
			\STATE $\textbf{Output:}$ Synthetic examples
			\STATE
			\STATE $\textbf{function:}$ ClusterSMOTE($DF, minorityLabel, totalExamplesToAdd, k, clusterK$)
			
			\STATE $minorityDF\leftarrow DF.$filter$(label = minorityLabel)$
			\STATE $kMeans\leftarrow $ KMeans.fit$(minorityDF)$
			\STATE $minorityDF[clusterId]\leftarrow kMeans$.transform$(minorityDF)$
			\STATE $clusters\leftarrow minorityDF.$groupBy$(clusterId)$
			\STATE $clusterKNNs\leftarrow clusters$.map$(x\rightarrow $KNN.fit$(x))$
			\STATE $clusterNearestNeighbors\leftarrow (0 $ to $ clusterK)$\par
			\hskip\algorithmicindent .map($x\rightarrow clusterKNNs[x]$.transform$(clusters[x])$)
			\STATE $randomClusterIds\leftarrow (0 $ to $ totalExamplesToAdd$)\par
			\hskip\algorithmicindent.map($x\rightarrow $randomInt(0, $ clusterK)$)
			\RETURN $randomClusterIds$.map$(x\leftarrow $generateExample$(clusterNearestNeighbors[x]))$
			\STATE
			
			\STATE $\textbf{function:}$ generateExample($cluster$)
			\STATE $example\leftarrow cluster[randomInt(0, cluster$.count())]
			\STATE $randomNeighbor\leftarrow example.neighbors.tail$[randomInt($0, k$)]
			\RETURN Array($example, randomNeighbor$)\par
			\hskip\algorithmicindent .map($x\rightarrow x[0] $  $ + $ randomDouble()$ $ $ * $ $ (x[1] - x[0]))$
			
		\end{algorithmic}
	}
\end{algorithm}

\smallskip
\noindent \textbf{Gaussian SMOTE} \cite{gaussianSmote} works very similar to standard SMOTE but uses a Gaussian instead of a uniform distribution when picking a point between two minority class examples. Gaussian SMOTE aims at enriching the representation of the minority class by introducing more spread within the artificial instances. This helps to avoid the patterns introduced with the original SMOTE algorithm where synthetic examples were only placed on the hyperplanes between existing examples.

\begin{algorithm}[!]
	{
		\caption{Gaussian SMOTE}
		\label{alg:gaussiansmote}
		\begin{algorithmic}
			\STATE $\textbf{Input:}$ DataFrame of the full dataset
			\STATE $\textbf{Parameters:}$ \textit{DF}: DataFrame of all examples in the dataset,
			\textit{minorityLabel}: label of the minority class, \textit{totalExamplesToAdd}: number of synthetic examples to create, \textit{k}: number of nearest neighbors to consider, \textit{sigma}: standard deviation for Gaussian sampling
			\STATE $\textbf{Output:}$ Synthetic examples
			\STATE
			\STATE $\textbf{function:}$ GaussianSMOTE($DF, minorityLabel, totalExamplesToAdd, k, sigma$)
			\STATE $minorityDF\leftarrow DF.$filter$(label = minorityLabel)$
			\STATE $knnModel\leftarrow$ KNN.fit$(minorityDF, k)$
			\STATE $nearestNeigbors\leftarrow knnModel$.transform($minorityDF$)
			\STATE $randomIndexes\leftarrow (0$ to $nearestNeigbors$.count())\par
			\hskip\algorithmicindent .map($x\rightarrow$ randomInt($0, totalExamplesToAdd$))
			\RETURN $randomIndexes$.map($x\rightarrow$ createSyntheticExample($nearestNeigbors[x], k$))
			\STATE
			
			\STATE $\textbf{function:}$ createSyntheticExample($example, k$)
			\STATE $example\leftarrow nearestNeighbors$[0]
			\STATE $randomNeighbor\leftarrow example.neighbors.tail($randomInt(0, $k$))
			\STATE $gap\leftarrow $ nextGaussian(0, $sigma$)
			\RETURN Array($example, randomNeighbor$).map($x\rightarrow x[0] $ \par
			\hskip\algorithmicindent $ + $ $ gap * (x[1] - x[0]))$
		\end{algorithmic}
	}
\end{algorithm}

\smallskip
\noindent \textbf{\textit{k}-Means SMOTE} \cite{kMeansSmote} starts by performing \textit{k}-Means and keeps clusters that have more minority than majority examples. For each cluster, the density of the minority examples is calculated to create a sampling weight. Proportional to the weight of each cluster, SMOTE is used to create synthetic examples. This is a direct extension of Cluster SMOTE, where additional information regarding the minority class distribution in each cluster us used to control the SMOTE oversampling.

\clearpage
\addtolength{\topmargin}{0in}
\resizebox{5in}{!}{
\begin{minipage}{\textwidth}
\begin{algorithm}[H]
	{
		\caption{\textit{k}-Means SMOTE}
		\label{alg:kmeanssmote}
		\begin{algorithmic}
			\STATE $\textbf{Input:}$ DataFrame of the full dataset
			\STATE $\textbf{Parameters:}$ \textit{DF}: DataFrame of all examples in the dataset,
			\textit{minorityLabel}: label of the minority class, \textit{totalExamplesToAdd}: number of synthetic examples to create, \textit{k}: number of nearest neighbors to consider, \textit{clusterK}: number of clusters to create
			\STATE $\textbf{Output:}$ Synthetic examples
			\STATE
			\STATE $\textbf{function:}$ kMeansSMOTE($DF, minorityLabel, totalExamplesToAdd, k, clusterK$)
			\STATE $minorityDF\leftarrow DF.$filter$(x\leftarrow x.label = minorityLabel)$
			\STATE $majorityDF\leftarrow DF.$filter$(x\leftarrow x.label \neq minorityLabel)$	
			\STATE $kMeans\leftarrow $ KMeans.fit$(DF)$
			\STATE $DF[clusterId]\leftarrow kMeans$.transform$(DF)$
			\STATE $clusters\leftarrow predictions$.groupBy$(DF[clusterId]))$
			\STATE $clusters[imbalancedRatio]\leftarrow clusters$.map($x\rightarrow $getImbalancedRatio$(x)$)
			\STATE $filteredClusters\leftarrow clusters$.filter($imbalancedRatio < irt$)
			
			\STATE $filteredClusters[averageDistance]\leftarrow filteredClusters$\par
			\hskip\algorithmicindent.map($x\rightarrow $getAverageDistance($x)$)
			\STATE $filteredClusters[densityFactor]\leftarrow filteredClusters$\par
			\hskip\algorithmicindent.map($x\rightarrow x$.count() / pow$(x[averageDistance], de)$)
			\STATE $filteredClusters[sparsityFactor]\leftarrow filteredClusters$\par
			\hskip\algorithmicindent.map($x\rightarrow 1 / x[densityFactor]$)
			
			\STATE $sparsitySum\leftarrow filteredClusters$.map($x\rightarrow x[sparsityFactor]$).sum()
			\STATE $filteredClusters[samplingWeight]\leftarrow filteredClusters$\par
			\hskip\algorithmicindent.map($x\rightarrow x[sparsityFactor] $ $ / $ $ sparsitySum$) 
			\STATE $clusterKNNs\leftarrow filteredClusters$.map$(x\rightarrow $KNN.fit$(x))$
			\STATE $clusterNearestNeighbors\leftarrow (0 $ to $ clusterK)$\par
			\hskip\algorithmicindent .map($x\rightarrow clusterKNNs[x]$.transform$(filteredClusters[x])$)

			\RETURN $clusterNearestNeighbors$.map($x\rightarrow $generateExamples$(x)$)

			\STATE $ $
			
			\STATE $\textbf{function:}$ getImbalancedRatio($clusterDF$)
			\STATE $minorityCount\leftarrow clusterDF$\par
			\hskip\algorithmicindent.filter$(x\leftarrow x.label = minorityLabel)$.count()
			\STATE $majorityCount\leftarrow clusterDF$\par
			\hskip\algorithmicindent.filter$(x\leftarrow x.label \neq minorityLabel)$.count()
			\RETURN $ (majorityCount + 1) $ $ / $ $ (minorityCount + 1)$
			\STATE $ $
			
			\STATE $\textbf{function:}$ getAverageDistance($clusterDF$)
			\RETURN $clusterDF$.map$(x\leftarrow clusterDF$.map$(y\leftarrow $getDistance($x, y$)).sum()).sum() \par\hskip\algorithmicindent / ($clusterDF$.count() * $clusterDF$.count())
			\STATE $ $ 
			
			\STATE $\textbf{function:}$ getDistance($x, y$)
			\RETURN Array$(x, y)$.transpose.map$(x\rightarrow (x[0] - x[1]) *(x[0] - x[1]))$.sum() 
			\STATE $ $ 
			
			\STATE $\textbf{function:}$ generateExamples($cluster, totalExamplesToAdd$)
			\STATE $examplesToAdd\leftarrow cluster[samplingWeight] * totalExamplesToAdd$ 
			\RETURN $(0 $ to $ examplesToAdd)$.map($x\rightarrow $ generateSingleExample($x$))
			\STATE $ $ 

			\STATE $\textbf{function:}$ generateSingleExample($neighbors, k$)
			\STATE $example\leftarrow neighbors.neighbors[0]$
			\STATE $randomNeighbor\leftarrow example.neighbors.tail$[randomInt($0, k$)]
			\RETURN Array($example, randomNeighbor$)\par
			\hskip\algorithmicindent .map($x\rightarrow x[0] $  $ + $ randomDouble()$ $ $ * $ $ (x[1] - x[0]))$
					
		\end{algorithmic}
	}
\end{algorithm}
\end{minipage}
}
\clearpage

\smallskip
\noindent \textbf{MWMOTE} \cite{mwmote} has three main phases: identify the most difficult minority class examples to learn from, assign a weight to each example based on its importance, and create synthetic examples. 
In the first phase, 1-NN is performed on the dataset and minority class examples with at least one other minority neighbor is kept. MWMOTE follows the idea that difficult instances are the ones that pose highest challenge to a classifier and are most likely to introduce errors during training. Therefore, MWMOTE focuses on decreasing the difficulty of such instances by creating uniform minority class regions around them. 

\addtolength{\topmargin}{0in}
\resizebox{4.5in}{!}{
\begin{minipage}{\textwidth}
\begin{algorithm}[H]
	{
		\caption{MWMOTE}
		\label{alg:mwmote}
		\begin{algorithmic}
			\STATE $\textbf{Input:}$ DataFrame of the full dataset
			\STATE $\textbf{Parameters:}$ \textit{DF}: DataFrame of all examples in the dataset,
			\textit{minorityLabel}: label of the minority class, \textit{totalExamplesToAdd}: number of synthetic examples to create, \textit{k1, k2, k3}: number of nearest neighbors to consider,
			\STATE $\textbf{Output:}$ Synthetic examples
			\STATE
			\STATE $\textbf{function:}$ MWMOTE($DF, minorityLabel, totalExamplesToAdd, k1, k2, k3$)
			\STATE $Smin\leftarrow DF.$filter$(label = minorityLabel)$
			\STATE $Smaj\leftarrow DF.$filter$(label \neq minorityLabel)$
			\STATE $knn1\leftarrow KNN.$fit$(DF, k1)$
			\STATE $SminNN\leftarrow knn1.$transform$(Smin)$
			\STATE $Sminf\leftarrow SminNN.$filter(hasMinorityNeighbors() $ > $ 0)
			\STATE $SmajNN\leftarrow $KNN.fit$(Smaj, k2)$		
			\STATE $Nmaj\leftarrow SmajNN.$transform$(Sminf)$
			\STATE $Sbmaj\leftarrow Nmaj.$map$(x\rightarrow x.neighbors).$distinct()
			\STATE $SminNN\leftarrow $KNN.fit$(Smin, k3)$
			\STATE $Nmin\leftarrow SminNN.$transform$(Sbmaj)$
			\STATE $Simin\leftarrow Nmin.$map$(x\rightarrow x.neighbors).$distinct()
			\STATE $Iw\leftarrow Sbmaj$.map$(y\rightarrow y$.map$(x\rightarrow $Cf$(y, x) $ $ * $ $ $Df$(y, x))$
			\STATE $Sw\leftarrow Simin$.map($x\rightarrow Sbmaj$.map$(y\rightarrow $Iw$(y, x))$.sum())
			\STATE $Sp\leftarrow Sw.$map$(x\rightarrow x $ / $Sw$.sum())
			\STATE $kMeans\leftarrow $ KMeans.fit$(Smin)$
			\STATE $Smin[clusterId]\leftarrow kMeans$.transform$(Smin)$
				
			\RETURN $(0 $ to $ examplesToAdd)$.map( $\underline{\hspace*{0.3cm}}\rightarrow $ generateExample($Smin$))
			
			\STATE
			\STATE $\textbf{function:}$ generateExample($Smin$)
					\STATE $example\leftarrow chooseByProbability(Smin)$
			\STATE $clusterExamples\leftarrow Smin.filter(clusterId = example.clusterId)$
			\STATE $neighbor\leftarrow clusterExamples[randomInt(0, clusterExamples.count()))]$
			\RETURN Array$(example, neighbors)$\par\hskip\algorithmicindent .transpose.map$(x\rightarrow x $ $ + $ randomDouble() $ * $ $ (y - x))$
			
			\STATE
			\STATE $\textbf{function:}$ hasMinorityNeighbors($example$)
			\STATE $label\leftarrow example.neighbors[0].label$
			\RETURN $example.neighbors$\par$\hskip\algorithmicindent$.map$(x\rightarrow $if $ x.label = label$: $ 1; $ else$: 0)$.sum()

			\STATE
			\STATE $\textbf{function:}$ Df($y, x, Simin$)
			\STATE $cf\leftarrow $ Cf($y, x$)
			\STATE $denominator\leftarrow Simin$.map($q\rightarrow $ Cf($y, q$)).sum()
			\RETURN $distance $ / $ denominator$

			\STATE
			\STATE $\textbf{function:}$ Cf($y, x$)
			\STATE $featureLength\leftarrow y$.length()
			\STATE $distance\leftarrow $ Array($y, x)$.transpose.map($x\rightarrow (x[0] - x[1]) * (x[0] - x[1])) $ \par\hskip\algorithmicindent / $ featureLength$
			\IF{$1 / distance \leq CfThreshold$}
			\STATE $cutoff\leftarrow 1 $ / $ difference$
			\ELSE 
			\STATE $cutoff\leftarrow CfThreshold$
			\ENDIF
			\RETURN $ (cutoff $ / $ CfThrehold) * CMAX$
		
		\end{algorithmic}
	}
\end{algorithm}
\end{minipage}
}
\clearpage
\smallskip
\noindent \textbf{NRAS} \cite{nras} works be removing noisy minority examples before creating SMOTE style synthetic examples. Using linear regression, the probability that each example in the dataset belongs to the minority class is determined. All minority class examples that have a membership probability less a threshold parameter value are removed from the dataset and SMOTE style oversampling is performed using the remaining minority class examples. NRAS was designed to handle highly noisy datasets that may be subject to various types of noise (both affecting class labels and features). A potential drawback may lie in forcefully trying to find noise even when it is not present in the training data.

\begin{algorithm}
	{
		\caption{NRAS}
		\label{alg:nras}
		\begin{algorithmic}
			\STATE $\textbf{Input:}$ DataFrame of the full dataset
			\STATE $\textbf{Parameters:}$ \textit{DF}: DataFrame of all examples in the dataset,
			\textit{minorityLabel}: label of the minority class, \textit{totalExamplesToAdd}: number of synthetic examples to create, \textit{k}: neighbor count for \textit{k}NN, \textit{threshold}: minimum number of minority neighbors needed to keep an example
			\STATE $\textbf{Output:}$ Synthetic examples
			\STATE
			\STATE $\textbf{function:}$ NRAS($DF, minorityLabel, totalExamplesToAdd, k, threshold$)
			\STATE $minorityDF\leftarrow DF.$filter$(x\rightarrow x.label = minorityLabel)$
			\STATE $minorityDF\leftarrow DF.$filter$(x\rightarrow x.label \neq minorityLabel)$
			
			\STATE $DF[propensityScore]\leftarrow $ LinearRegression$(DF)$
			\STATE $knn\leftarrow $KNN.fit$(DF, k)$
			\STATE $nearestNeighbors\leftarrow knn.$transform$(minorityDF)$  
			\STATE $samplingRatio\leftarrow nearestNeighbors$.count() $ / $ $ totalExamplesToAdd$
			
			\STATE $keepMinority\leftarrow nearestNeighbors.$filter$(neighbors.tail$\par\hskip\algorithmicindent.map$(x\rightarrow if $ $ x.label = minorityLabel$: $1; $ $ else$: $ 0)$.sum() $ \geq $ $ threshold))$

			\RETURN $keepMinority$.map($x\rightarrow $generateExamples($x, samplingRatio, k$))
			\STATE 
			 
			\STATE $\textbf{function:}$ generateExamples($example, samplingRatio, k$)
			\RETURN $(0 $ to $ samplingRatio)$\par\hskip\algorithmicindent.map( \underline{\hspace*{0.3cm}} $\rightarrow $ generateSingleExample($example, k$))
			\STATE $ $ 
			
			\STATE $\textbf{function:}$ generateSingleExample($example, k$)
			\STATE $example\leftarrow example.neighbors[0]$
			\STATE $randomNeighbor\leftarrow example.neighbors.tail$[randomInt($0, k$)]
			\RETURN Array($example, randomNeighbor$)\par
			\hskip\algorithmicindent .map($x\rightarrow x[0] $  $ + $ randomDouble()$ $ $ * $ $ (x[1] - x[0]))$
		\end{algorithmic}
	}
\end{algorithm}

\smallskip
\noindent \textbf{Random Oversampling} duplicates existing examples instead of creating synthetic ones. For each minority class, random examples are duplicated until the desired level of class balance is achieved.

\begin{algorithm}
	{
		\caption{Random Oversample}
		\label{alg:nras}
		\begin{algorithmic}
			\STATE $\textbf{Input:}$ DataFrame of the full dataset
			STATE $\textbf{Parameters:}$ \textit{DF}: DataFrame of all examples in the dataset \textit{minorityLabel}: label of the minority class, \textit{examplesToAdd}: number of synthetic examples to create
			\STATE $\textbf{Output:}$ Synthetic examples
			\STATE
			\STATE $\textbf{function:}$ RandomOversample($DF, minorityLabel, examplesToAdd$)
			\STATE $minorityDF\leftarrow DF.$filter$(x\rightarrow x.label = minorityLabel)$
			\STATE $samplingRate\leftarrow examplesToAdd $ /  $ minorityDF.$count()
			\RETURN $minorityDF.$sample($withReplacement=True, samplingRate)$
		\end{algorithmic}
	}
\end{algorithm}

\smallskip
\noindent \textbf{RBO} \cite{koziarski2019radial} begins by picking a random minority example for each synthetic example to be created. Next, a randomized translation vector is created and applied to a copy of the picked example. The \textit{phi} function is run on both examples and if the result corresponding to the translated example is smaller, the translation vector is applied to the original example. This process is repeated until the maximum number of iterations have completed or a random stopping probability condition has been met. Finally, the translated point is added as a new synthetic example. RBO offers a powerful alternative to SMOTE-based oversampling that is free of nearest neighbor search. Additionally, RBO automatically identifies difficult regions within each minority class and adjusts its local oversampling ratio accordingly. 

\addtolength{\topmargin}{0in}
\resizebox{6in}{!}{
	\begin{minipage}{\textwidth}
		\begin{algorithm}[H]
			{
		\caption{RBO}
		\label{alg:rbo}
		\begin{algorithmic}
			\STATE $\textbf{Input:}$ DataFrame of the full dataset
			\STATE $\textbf{Parameters:}$ \textit{DF}: DataFrame of all examples in the dataset, \textit{minorityLabel}: label of the minority class, \textit{examplesToAdd}: number of synthetic examples to create, \textit{gamma}: spread of radial basis function, \textit{stepSize}: step size for point translation, \textit{iterations}: number of iterations per synthetic sample, \textit{probability}: probability of stopping early
			\STATE $\textbf{Output:}$ Synthetic examples
			\STATE  
			\STATE $\textbf{function:}$ RBO($DF, minorityLabel, examplesToAdd, gamma, stepSize, iterations, probability$)\STATE $minorityDF\leftarrow DF.$filter$(x\rightarrow x.label = minorityLabel)$
			\STATE $majorityDF\leftarrow DF.$filter$(x\rightarrow x.label \neq minorityLabel)$
			\RETURN $(0 $ to $ examplesToAdd)$\par$\hskip\algorithmicindent$.map$($ $\underline{\hspace*{0.3cm}} \rightarrow $createdExample$(minorityDF, majorityDF, gamma,$\par$\hskip\algorithmicindent\hskip\algorithmicindent stepSize, iterations, probability))$
			
			\STATE
			\STATE $\textbf{function:}$ createExample($majorityDF, minorityDF, gamma,$\par$\hskip\algorithmicindent stepSize, iterations, probability$)
			\STATE $point\leftarrow minortyDF.$getRandomExample()
			\STATE $pointPhi\leftarrow calculatePhi(point, majorityDF, minorityDF, gamma)$
			
			\STATE $randomStopIndex\leftarrow $getRandomStopIndex($iterations, probability$) 
			\FOR{$i=0$ to $randomStopIndex$}
			\STATE $randomDirection\leftarrow features.indicies$.map$(x\rightarrow $randomChoice$(-1, 1))$
			\STATE $randomVector\leftarrow features.indicies.$map$(x\rightarrow $randomDouble()$)$
			\STATE $translated\leftarrow point +  randomDirection * randomVector * stepSize$
			\IF{ $|calculatePhi(translated, majorityDF, minorityDF, gamma)| < $\par$\hskip\algorithmicindent |calculatePhi(point, majorityDF, minorityDF, gamma)|$}
			\STATE $point\leftarrow translated$
			\ENDIF 
			\ENDFOR
			\RETURN $point$
			\STATE
			
			\STATE $\textbf{function:}$ getRandomStopIndex($iterations, probability$)
		  	\IF{$probability = 1.0$}
		  	\RETURN $iterations$
		  	\ELSE
		  	\RETURN $iterations * probability$ $ + $ randomGaussian() $ * $ $ iterations $ $ * $\par$ \hskip\algorithmicindent probability)$
		  	\ENDIF
			\STATE
			
			\STATE $\textbf{function:}$ calculatePhi($x, K, k, gamma$)
			\STATE $majorityValue\leftarrow K.$map$(Ki\rightarrow$\par$\hskip\algorithmicindent$exp(-power(pointDifference$(Ki, x)/gamma$, 2))).collect().sum()
			\STATE $minorityValue\leftarrow k.$map$(ki\rightarrow$\par$\hskip\algorithmicindent$exp(-power(pointDifference$(ki, x)/gamma$, 2))).collect().sum()
			\RETURN $majorityValue - minorityValue$
			\STATE
			\STATE $\textbf{function:}$ pointDifference($x, y$)
			\RETURN $(x, y).$transpose.map$(x\rightarrow$$\lvert \lvert x[0]-x[1] \rvert \rvert$$)$.sum()
		\end{algorithmic}
	}
\end{algorithm}
\end{minipage}
}

\smallskip
\noindent \textbf{Safe Level SMOTE} \cite{safe_level_smote} performs \textit{k}NN and the nearest neighbors for each minority example is returned. For each of these examples (\textit{p}), a random neighbor (\textit{n}) is selected the number of neighboring minority examples are counted. A safe level ratio is created by dividing the \textit{p} minority nearest neighbor count by the \textit{n} minority nearest neighbor count. Based on this ratio, different explicit rules are applied to generate new synthetic examples. Safe Level SMOTE aims at avoiding enhancement of noisy or difficult instances. However, very often those difficult instances are the core concepts of the minority class and therefore avoiding their oversampling may lead to a drop of performance on testing sets.

\begin{algorithm}
	{
		\caption{Safe Level SMOTE}
		\label{alg:safelevelsmote}
		\begin{algorithmic}
			\STATE $\textbf{Input:}$ DataFrame of the full dataset
			\STATE $\textbf{Parameters:}$ \textit{DF}: DataFrame of all examples in the dataset,
			\textit{minorityLabel}: label of the minority class,
			\textit{examplesToAdd}: number of synthetic examples to create, \textit{k}: neighbor count for \textit{k}NN
			\STATE $\textbf{Output:}$ Synthetic examples
			\STATE
			\STATE $\textbf{function:}$ SafeLevelSMOTE($DF, minorityLabel, examplesToAdd, k$)
			\STATE $minorityDF\leftarrow DF$.filter$(x\rightarrow x.label = minorityLabel)$
			\STATE $knn\leftarrow $ KNN.fit$(DF, k)$
			\STATE $minorityNearestNeighbors\leftarrow knn$.transform$(minorityDF)$
			\STATE $randomIndexes\leftarrow (0$ to $examplesToAdd$)\par$\hskip\algorithmicindent$.map($x\rightarrow$ randomInt($0, minorityNearestNeighbors$.count()))
			\RETURN $randomIndexes$.map($x\rightarrow $generateExample$(minorityNearestNeighbors[x]))$
	
			\STATE
			\STATE $\textbf{function:}$ generateExample($nearestNeighbors$)
			\STATE $example\leftarrow nearestNeighbors.head$
			\STATE $randomNeighbor\leftarrow nearestNeighbors.tail($randomInt(0, $k$))
			\STATE $safeLevel\textsubscript{p}\leftarrow example.neighbors$\par$\hskip\algorithmicindent.$filter$(label = minorityLabel)$.count()
			\STATE $safeLevel\textsubscript{n}\leftarrow randomNeighbor.neighbors$\par$\hskip\algorithmicindent.$filter$(x\rightarrow label = minorityLabel)$.count()
			\STATE $safeLevelRatio\leftarrow 0$
			\IF{$safeLevel\textsubscript{n} \neq 0$}
			\STATE $safeLevelRatio = safeLevel\textsubscript{p} $ / $ safeLevel\textsubscript{n}$
			\ELSE
			\STATE $safeLevelRatio\leftarrow \infty$
			\ENDIF
			\IF{$safeLevelRatio = \infty $ and $ saveLevel\textsubscript{p} = 0$}
			\RETURN $\varnothing$
			\ELSE
			\IF{$safeLevelRatio = \infty $ and $ safeLevel\textsubscript{p} \neq 0$}
			\STATE $gap\leftarrow 0$ 
			\ELSIF{$safeLevelRatio = 1$}
			\STATE $gap\leftarrow $randomDouble$(0, 1) $
			\ELSIF{$safeLevelRatio > 1$}
			\STATE $ gap\leftarrow $randomDouble$(0, 1$ $/$ $safeLevelRatio)$
			\ELSIF{$safeLeveRatio < 1$}
			\STATE{$gap\leftarrow $randomDouble$(1$ $-$ $safeLevelRatio, 1))$}
			\ENDIF
			\ENDIF
			\RETURN $features.$map$(x\rightarrow x[0] + gap * (x[1] - x[0]))$
		\end{algorithmic}
	}
\end{algorithm}

\smallskip
\noindent \textbf{SMOTE} \cite{smote} performs \textit{k}NN on the minority class and then randomly chooses existing examples as the basis for new synthetic examples. Each synthetic example is created on a random point between the previously chosen example and one of its random nearest neighbors.
\begin{algorithm}
	{
		\caption{SMOTE}
		\label{alg:smote}
		\begin{algorithmic}
			\STATE $\textbf{Input:}$ DataFrame of the full dataset
			\STATE $\textbf{Parameters:}$ \textit{DF}: DataFrame of all examples in the dataset,
			\textit{minorityLabel}: label of the minority class, \textit{examplesToAdd}: number of synthetic examples to create, \textit{k}: neighbor count for \textit{k}NN
			\STATE $\textbf{Output:}$ Synthetic examples
			\STATE
			\STATE $\textbf{function:}$ SMOTE($DF, minorityDF, examplesToAdd, k$)
			\STATE $knnModel\leftarrow$ KNN.fit$(minorityDF, k)$
			\STATE $nearestNeigbors\leftarrow knnModel$.transform($minorityDF$)
			\STATE $randomIndexes\leftarrow (0$ to $examplesToAdd$).map($x\rightarrow$ randomInt($0, k$))
			\RETURN $randomIndexes$.map($x\rightarrow$ createSmoteStyleExample($nearestNeighbors[x], k$))
			\STATE
	
			\STATE $\textbf{function:}$ createSmoteStyleExample($nearestNeighbors, k$)
			\STATE $example\leftarrow nearestNeighbors.head$
			\STATE $randomNeighbor\leftarrow nearestNeighbors.tail($randomInt(0, $k$))
			\STATE $gap\leftarrow $ randomDouble()
			\STATE $features\leftarrow $Array($example.features, randomNeighbor.features$)
			\RETURN $features$.map($x\rightarrow x[0] + gap * (x[1] - x[0]))$
		\end{algorithmic}
	}
\end{algorithm}

\smallskip
\noindent \textbf{SMOTE-D} \cite{smote_d} is designed to be a deterministic version of the standard SMOTE algorithm. \textit{k}NN is performed on the minority examples and the standard deviation between their nearest neighbors is calculated. Each minority class examples is oversampled proportional to its standard deviation divided by the sum of all standard deviations. For each of these minority examples, synthetic examples are evenly created between itself and each nearest neighbor such that neighbors further away are used to generate the most examples.

\begin{algorithm}
	{
		\caption{SMOTE-D}
		\label{alg:smoted}
		\begin{algorithmic}
			\STATE $\textbf{Input:}$ DataFrame of the full dataset
			\STATE $\textbf{Parameters:}$ \textit{DF}: DataFrame of all examples in the dataset,
			\textit{minorityLabel}: label of the minority class, \textit{examplesToAdd}: number of synthetic examples to create, \textit{k}: neighbor count for \textit{k}NN
			\STATE $\textbf{Output:}$ Synthetic examples
			\STATE
			\STATE $\textbf{function:}$ getSmoteSamples($DF, minorityLabel, examplesToAdd, k$)
			\STATE $knn\leftarrow$ KNN.fit($minorityDF, k$)
			\STATE $nearestNeighbors\leftarrow$ knn.transform($minorityDF$)
			\STATE $nearestNeighbors[std]\leftarrow minorityNeighbors$\par$\hskip\algorithmicindent$.map$(x\rightarrow$ getDistanceSTD($x$))
			\STATE $nearestNeighbors[stdWeights]\leftarrow neighborDistanceSTDs$\par$\hskip\algorithmicindent$.map($x\rightarrow x[std] / neighborNeighbors[std].sum$)
			\STATE $nearestNeighbors[examplesToAdd]\leftarrow$\par$\hskip\algorithmicindent nearestNeighbors[stdWeights] * examplesToAdd$
			\RETURN $nearestNeighbors$.map$(x\rightarrow$ sampleCurrentExample($x$))
			\STATE
			
			\STATE $\textbf{function:}$ sampleCurrentExample($example$)
			\STATE $neighbors\leftarrow example.neighbors$
			\STATE $neighbors[distances]\leftarrow neighbors$\par$\hskip\algorithmicindent$.map($x\rightarrow $distance$(example, x))$
			\STATE $neighbors[examplesToAdd]\leftarrow neighbors[distances]$\par$\hskip\algorithmicindent$.map($x\rightarrow (x * example[examplesToAdd])$ $ / $ $ distances.sum)$
			\RETURN $neighbors.map(x\rightarrow $createExamples($example, x$))
			\STATE
			\STATE $\textbf{function:}$ createExamples($example, neighbor$)
			\STATE $examplesToAdd\leftarrow neighbor[examplesToAdd]$
			\STATE $syntheticExamples\leftarrow (0$ to $neighbor[examplesToAdd])$\par$\hskip\algorithmicindent$.map$(x\rightarrow example + (neighbor - examples)$\par$\hskip\algorithmicindent *$ $ (x + 1) / neighbor[examplesToAdd])$
			\RETURN $syntheticExamples$
			
		\end{algorithmic}
	}
\end{algorithm}

\begin{figure}[h!]
	\centering
	\includegraphics[width=40mm]{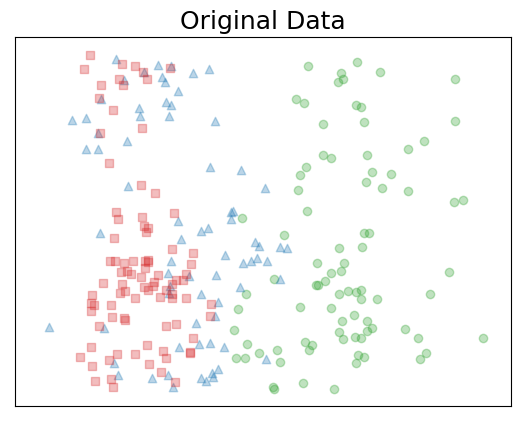}
	\includegraphics[width=40mm]{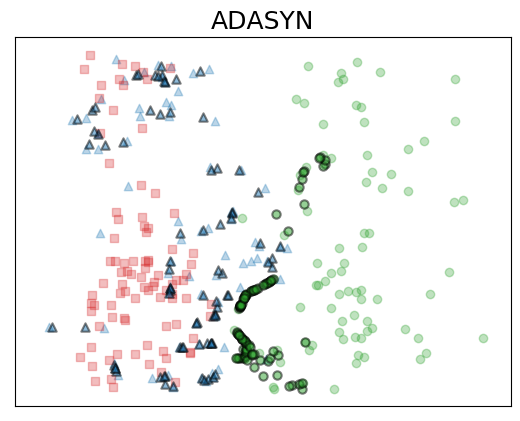}
	\includegraphics[width=40mm]{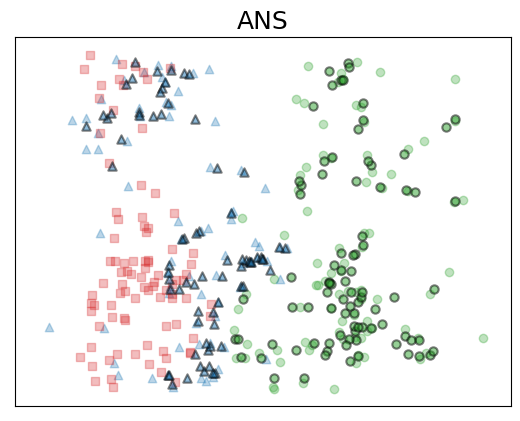}
	\vspace{1mm}
	
	\includegraphics[width=40mm]{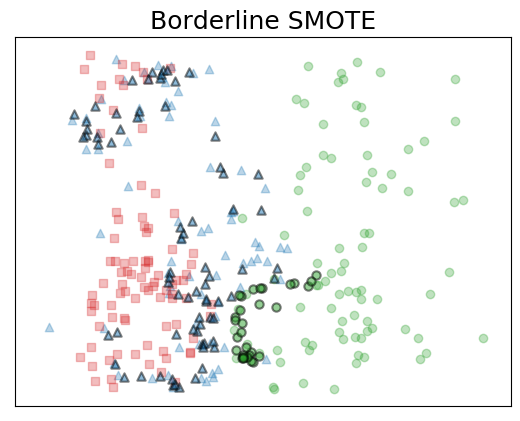}
	\includegraphics[width=40mm]{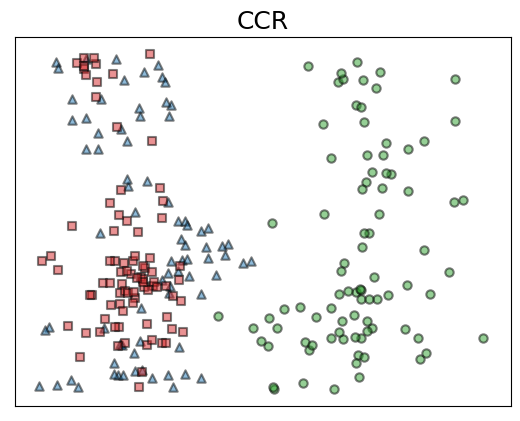}
	\includegraphics[width=40mm]{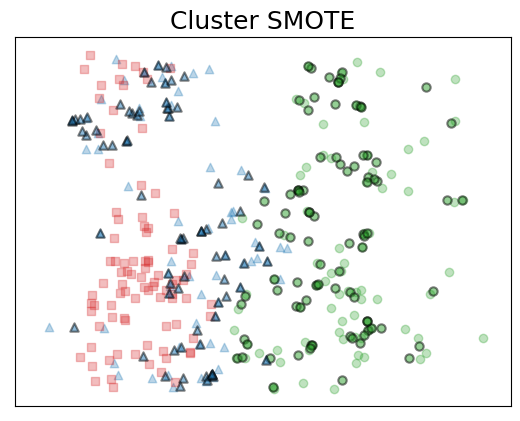}
	\vspace{1mm}

    \includegraphics[width=40mm]{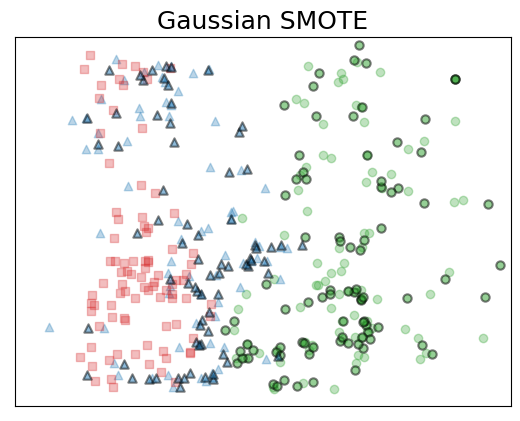}
	\includegraphics[width=40mm]{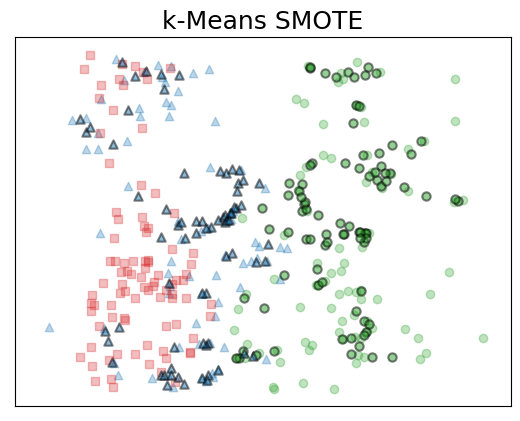}
	\includegraphics[width=40mm]{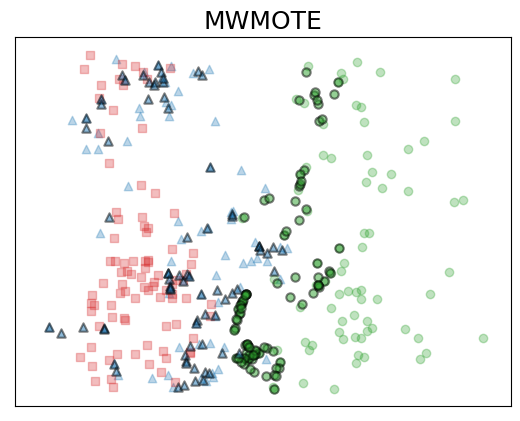}
	\vspace{1mm}
	
	\includegraphics[width=40mm]{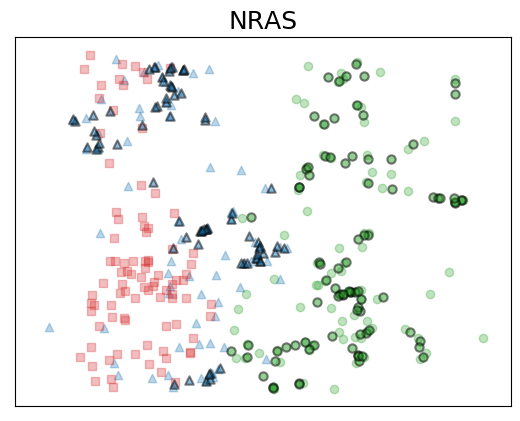}
	\includegraphics[width=40mm]{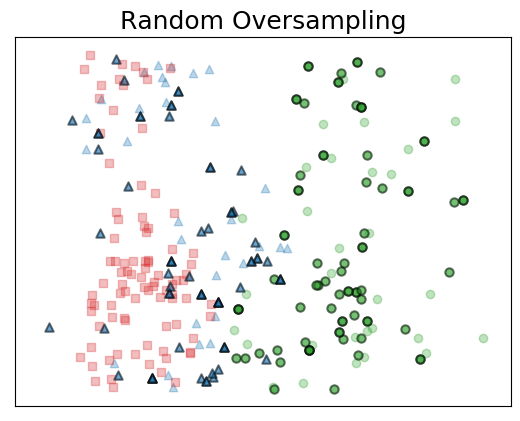}
	\includegraphics[width=40mm]{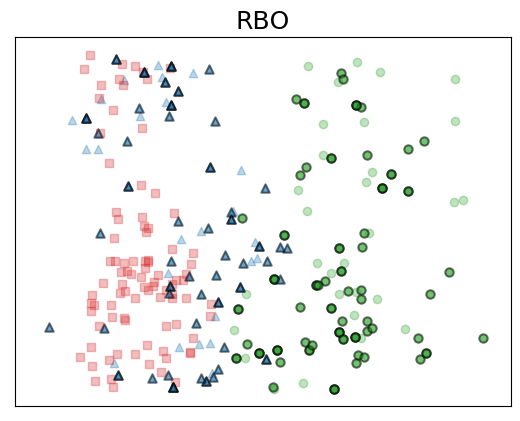}
	\vspace{1mm}
	
	\includegraphics[width=40mm]{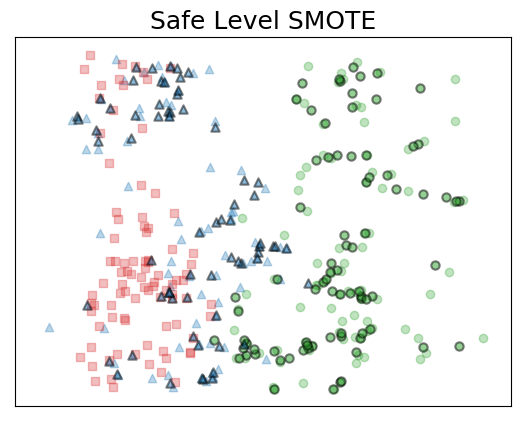}
	\includegraphics[width=40mm]{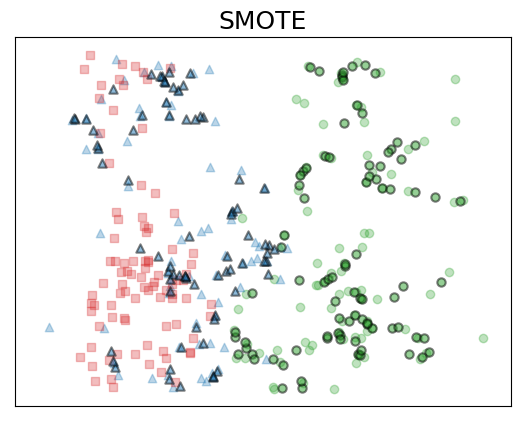}
	\includegraphics[width=40mm]{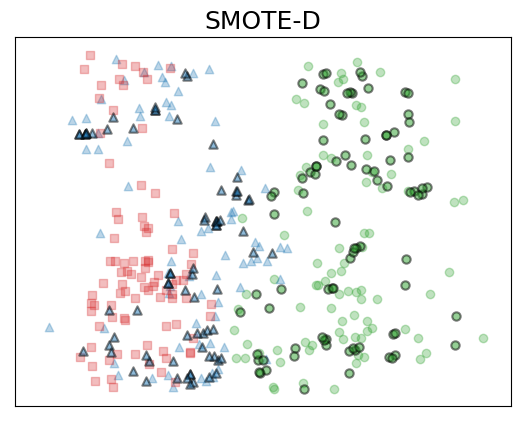}
	
	\caption{Shown with black borders, synthetic examples from each sampling method are combined with the original data. The upper left plot shows the original data without oversampling.}
	\label{fig:sampling_examples}
\end{figure}

\section{Software package}
\label{sec:sof}

The proposed library (https://github.com/fsleeman/spark-class-balancing) was developed to both compare the original oversampling methods and provide the first Spark implementation written in Scala. These oversampling algorithms include: ADASYN, ANS, Borderline SMOTE, CCR, Cluster SMOTE, Gaussian SMOTE, \textit{k}-Means SMOTE, MWMOTE, NRAS, Random Oversampling, RBO, Safe Level SMOTE, SMOTE, SMOTE-D. Since each of these algorithms were designed for binary class problems, this Spark implementation has added multi-class support. Each minority class is processed separately, but for algorithms that take in account the majority class a one-vs-rest approach was used. In this case, all examples not part of the minority class was treated as if they were part of a single majority class. The sampling methods were developed in the style of the Spark Machine Learning Library (MLlib) methods using the \textit{fit} and \textit{transform} pattern.

Each of these oversampling methods were originally presented as serial algorithms which required some modifications when implemented for Spark. However, we have kept these implementation as close to the original as possible and have not made major improvements on performance or accuracy. In Section \ref{sec:fut}, we discuss some ways the running time of these algorithms could be improved and a discussion on the types of algorithms that likely will be more successful when implemented using Spark.

\subsection{Dependencies}
The spark-class-balancing library was developed using Spark 3.0.1 with Scala 2.12.1 and currently has following dependencies: spark-core, spark-sql, spark-mllib, breeze-natives, breeze, spark-knn (from the fsleeman fork). The original spark-knn implementation was based on Spark 2 and so the forked version was updated to be compatible with Spark 3.

\subsection{spark-knn}
Many of these sampling algorithms, such as SMOTE variants, include a \textit{k}-Nearest Neighbors search. Instead of creating that method from scratch, an efficient implementation from spark-knn (https://github.com/saurfang/spark-knn) was used. However, the spark-knn implementation did not support nearest neighbor search by distance radius which was required by ANS and CCR. To address this limitation, spark-knn was forked (https://github.com/fsleeman/spark-knn) and the distance search feature was added. 

\subsection{Using the spark-class-balancing library}
The spark-class-balancing library is designed to build a fat jar using the \textit{sbt} tool, with the following commands:
\begin{minted}{c}
	sbt compile
	sbt assembly
\end{minted}

Since the spark-knn (fsleeman fork) library is a dependency of spark-class-balancing, it must also be built as far jar using a similar method and the \textit{build.sbt} file must be updated to include that generated file. The resulting spark-class-balancing jar file can then be used for running a Spark job, both in the local or cluster mode.

\subsection{Invoking sampling methods}
Invoking one of the oversampling methods is straightforward as shown below in this example using standard SMOTE.

\begin{minted}{scala}
	val method = new SMOTE
	val model = method.fit(trainData).setK(5)
	val sampledData = model.transform(trainData)
\end{minted}

Using the \textit{fit/transform} pattern, a new SMOTE estimator is instantiated and then a model is created with the \textit{fit} function. The resulting model is set with prediction parameters, in this case setting the \textit{k} value to 5, and finally the \textit{trainData} DataFrame is transformed to produce oversampled data. The same process is done for every other oversampling methods, although available parameters may change.

\subsection{Limitations and Future Extensions}
These algorithms were implemented in the spirit of their original design to better assess how they translate to the Spark platform in terms of performance and accuracy. While some of these implementations have limitations in efficiency, they can now be updated based on the algorithm design guidelines presented in Section \ref{sec:fut}. While this initial version of spark-class-balancing has 14 oversampling methods, it can be extended to include other types of sampling methods including undersampling, hybrid methods and ensembles.

The \textit{k}NN algorithm from the spark-knn package can run into errors depending on the settings used and properties of the data. The spark-knn implementation uses a hybrid spill tree which switches between spill and metric trees based on data metrics at each node split. If data is not well balanced between the two child notes, the tree generation may fail. However, this can be avoided by forcing the use of metric trees at all nodes. Other issues can occur with small datasets or few examples for a given class, but in that case spark-class-balancing includes a fallback to use the brute force \textit{k}NN mode.

In the future, this library can be built as a Spark package with a better integration with the custom fork of the spark-knn package. The spark-class-balancing library can also be updated to the newest versions of Spark as they come out. These Spark updates can include new built-in machine learning tools or general features which can be utilized in future sampling methods.

\subsection{Adding new algorithms}
A Scala class named \textit{samplingTemplate} has been added as a blank slate for developing new algorithms. After this class is copied to a new file, the placeholder names can be replaced and the core code can be added, using the other algorithms as an example.

\section{Experimental study}
\label{sec:exp}

We have designed the experimental study to answer the following research questions:

\begin{itemize}
    \item[RQ1:] Which of the proposed big data oversampling approaches adapted to the Spark architecture offers the best predictive performance?
    \item[RQ2:] What is the computational cost of the proposed big data oversampling approaches?
    \item[RQ3:] Which of the proposed big data oversampling approaches offers the best trade-off between predictive accuracy and computational complexity?
 \item[RQ4:] How does the examined algorithms handle binary and multi-class problems, as well as different types of features?
    \item[RQ5:] How well do the examined oversampling methods scale with the size of the data and which algorithms become prohibitively expensive to be used efficiently?
\end{itemize}

\subsection{Setup}
\label{sec:set}

\noindent \textbf{Hardware}. All experiments were performed using 32 threads on a large shared machine with Intel Xeon E7-8894 cores.

\noindent \textbf{Datasets}. We have chosen 26 datasets that include two-classes with binary features, two-classes with continuous features and multi-class with continuous features. These datasets provide a wide range of features and class imbalance ratios which gives some insight on how different oversampling-classifier combinations perform on these types of data. Initial tests showed that using the full datasets was not feasible with the time required for each test, so we decided to select a sub-section of 100,000 examples from each dataset which is still significantly more than used in the original algorithm publications. In addition to making the running time achievable, it also allows for the direct time comparisons between each dataset. Table \ref{table:datasets} shows high level parameters of these datasets.

\begin{table}[ht!]
	\centering
	\caption{High level properties of the twenty six datasets included in the following experiments.}
	\label{table:datasets}
	\renewcommand{\arraystretch}{1.25}
	\begin{adjustbox}{center,max width = 115mm}
		\footnotesize
		\begin{tabular}{lcccr}\hline
			\multicolumn{5}{c}{\textbf{Dataset Descriptions}}\\
			\bottomrule
			\textbf{Dataset} & \textbf{Class Count} & \textbf{Feature Count} & \textbf{Feature Type} & \textbf{Max Imbalanced Ratio}\\
			\bottomrule
			Bitcoin & 2 &  8 & Continuous & 69.42\\
			Cover Type & 7 &  54 & Continuous & 103.08\\
			Fuzzing & 2 &  115 & Continuous & 4.19\\
			HIGGS: Imbalanced Ratio 16:1 & 2 &  28 & Continuous & 16.00\\
			HIGGS: Imbalanced Ratio 4:1 & 2 &  28 & Continuous & 4.00\\
			HIGGS: Imbalanced Ratio 8:1 & 2 &  28 & Continuous & 8.00\\
			IoT & 11 &  115 & Continuous & 4.82\\
			MIRAI & 2 &  116 & Continuous & 5.28\\
			Poker: 0 vs 2 & 2 &  85 & Discrete & 10.53\\
			Poker: 0 vs 3 & 2 &  85 & Discrete & 23.73\\
			Poker: 0 vs 4 & 2 &  85 & Discrete & 129.04\\
			Poker: 0 vs 5 & 2 &  85 & Discrete & 251.53\\
			Poker: 0 vs 6 & 2 &  85 & Discrete & 352.36\\
			Poker: 1 vs 2 & 2 &  85 & Discrete & 8.87\\
			Poker: 1 vs 3 & 2 &  85 & Discrete & 20.00\\
			Poker: 1 vs 4 & 2 &  85 & Discrete & 108.77\\
			Poker: 1 vs 5 & 2 &  85 & Discrete & 211.77\\
			Poker: 1 vs 6 & 2 &  85 & Discrete & 296.62\\
			SEER & 10 &  11 & Continuous & 5.69\\
			SSL Renegotiation  & 2 &  115 & Continuous & 22.83\\
			SUSY: Imbalanced Ratio 16:1 & 2 &  18 & Continuous & 16.00\\
			SUSY: Imbalanced Ratio 4:1 & 2 &  18 & Continuous & 4.00\\
			SUSY: Imbalanced Ratio 8:1 & 2 &  18 & Continuous & 8.00\\
			SYN DOS & 2 &  115 & Continuous & 392.70\\
			Traffic & 7 &  26 & Continuous & 7.60\\
			Video Injection & 2 &  115 & Continuous & 23.12\\
			\bottomrule
		\end{tabular}
	\end{adjustbox}
\end{table}

\noindent \textbf{Classifiers}. We have used three popular classifiers that are available in the Spark Machine Learning Library (MLlib) - Random Forest, Support Vector Machines (SVM) and Naive Bayes. Since each classifier uses a different base algorithm, our experiments give some insight on how sampling methods may affect classifiers from different families.

\noindent \textbf{Parameters of classifier}. Table \ref{table:classifier_hyperparams} shows the hyperparameters used for each classifier. These values were chosen by a manual search that provided good accuracy compared to running time on several datasets.

\begin{table}[ht!]
	\centering
	\caption{Classifier hyperparameters used in the sampling experiments.}
	\label{table:classifier_hyperparams}
	\footnotesize
	\renewcommand{\arraystretch}{1.2}
	\begin{tabular}{ll}
		\toprule
		\textbf{Algorithm} & \textbf{Hyperparameters} \\
		\midrule
		\multirow{2}{*}{Support Vector Machine}&\textit{number of iterations} = 100\\
		&\textit{regularization parameter (C)} = 10.0\\
		\hline
		\multirow{2}{*}{Random Forest }&\textit{number of trees} = 100\\
		&\textit{maximum depth} = 20\\
		\hline		
		\multirow{1}{*}{Naive Bayes}&None\\
		\hline
		\bottomrule
	\end{tabular}
\end{table}

\noindent \textbf{Oversampling methods}.
We have implemented 14 oversampling methods detailed in Section \ref{sec:ovs} and were used with all combinations of classifiers and datasets. However, the sampling algorithms ANS, MWMOTE and RBO were significantly slower and full experiments were not feasible. Section \ref{sec:fut} provides a discussion on the properties of these algorithms which lead to limits in scalability.

\noindent \textbf{Parameters of oversampling methods}. Table \ref{table:sampling_hyperparameters} shows the list of hyperparameters used for each oversampling algorithm. Since the following experiments included a combination of datasets, classifiers and oversampling methods, an exhaustive hyperparameter search was not practical and so we have chosen a single set of hyperparameters for each oversampling method. These values were chosen from the original papers and a limited manual search was performed if no values were presented. As 5 is often used as the value of \textit{k} and since it was chosen in the original SMOTE algorithm, we have kept that value as well. The hyperparameters for RBO were set in such a way that only a single iteration per synthetic example was run, which deviates from the experiments in the original paper. This value was chosen in an attempt to compensate for the slow running time which is further discussed in Section \ref{sec:fut}.
 
\begin{table}[ht!]
	\centering
	\caption{Hyperparameters used for each sampling method.}
	\label{table:sampling_hyperparameters}
	\footnotesize
	\renewcommand{\arraystretch}{1.2}
	\begin{tabular}{ll}
		\toprule
		\textbf{Algorithm} & \textbf{Hyperparameters} \\
		\hline
		\multirow{1}{*}{ADASYN}&\textit{k} = 5 \\
		\hline
		\multirow{3}{*}{ANS}&\textit{k} = 5\\
		& \textit{C max ratio} = 0.25\\
		& \textit{distance neighbor limit} = 100\\
		\hline
		\multirow{1}{*}{Borderline SMOTE}&\textit{k} = 5\\
		\hline
		\multirow{2}{*}{CCR}&\textit{k} = 5\\
		& \textit{energy} = 1.0\\
		& \textit{distance neighbor limit} = 100\\
		\hline
		\multirow{2}{*}{Cluster SMOTE}&\textit{k} = 5 \\
		& \textit{cluster k} = 5\\
		\hline
		\multirow{2}{*}{Gaussian SMOTE}&\textit{k} = 5\\
		& \textit{sigma} = 0.5\\
		\hline
		\multirow{3}{*}{\textit{k}-Means SMOTE}&\textit{k} = 5\\
		& \textit{cluster k} = 5\\
		& \textit{imbalancedThreshold} = 10.0\\
		\hline		
		\multirow{4}{*}{MWMOTE}&\textit{k1} = 5 \\
		& \textit{k2} = 5\\
		& \textit{k3} = 5\\
		& \textit{cluster k} = 10\\
		& \textit{C max} = 3.0\\
		& \textit{$C_{f}$(th)} = 50.0\\
		\hline
		\multirow{2}{*}{NRAS}&\textit{k} = 5\\
		& \textit{threshold} = 3\\
		\hline
		\multirow{1}{*}{Random Oversampling} & None\\
		\hline
		\multirow{4}{*}{RBO}&\textit{gamma} = 1.0\\
		& \textit{iterations} = 1\\
		& \textit{step size} = 0.01\\
		& \textit{stopping probability} = 1.0\\
		\hline
		\multirow{2}{*}{Safe Level SMOTE}&\textit{k} = 5\\
		& \textit{sampling correction rate} = 0.05\\
		\hline
		\multirow{1}{*}{SMOTE}&\textit{k} = 5 \\
		\hline
		\multirow{1}{*}{SMOTE-D}&\textit{k} = 5 \\
		\hline
		\bottomrule
	\end{tabular}
\end{table}

 \noindent \textbf{Adaptation to multi-class problems}.  Each minority class was oversampled to the size of the majority class individually, resulting in a fully balanced dataset.

\noindent \textbf{Training and testing}. We have used a stratified 5-fold cross validation in order to maintain the original class distribution among the folds.

\noindent \textbf{Evaluation metrics}. In order to properly evaluate the performance of classifiers on imbalanced domains, skew-insensitive metrics are required. As multi-class imbalanced problems were not popular until recently among researchers, there is a lack of uniform approach to which measures should be considered as a standard. In order to gain an in-depth insight into the performance of the analyzed classifiers, Figure \ref{fig:metrics} shows four popular metrics for multi-class imbalanced data used to evaluate the presented oversampling algorithms.

\begin{figure}
\centering
\begin{equation*}
\begin{aligned}
AvAcc = \sum\limits_{i=1}^C \frac{tp_{i}+tn_{i}}{tp_{i}+tn_{i}+fp_{i}+fn_{i}} \\[0pt]
MAvG = \sqrt[C]{\prod_{i=0}^{C} recall_{i}}\\[0pt]
AvF_{\beta} = \frac{1}{C}\sum\limits_{i=1}^C \frac{(1+\beta^2) \cdot precision_{i} \cdot recall_{i}}{\beta^2 \cdot precision_{i} + recall_{i}}  \\[0pt]
CBA = \sum\limits_{i=1}^C \frac{mat_{i,i}}{max(\sum\limits_{j=1}^C mat_{i,j},\sum\limits_{j=1}^C mat_{j,i})}  
\end{aligned}
\end{equation*}
\caption{Formulas for the metrics used to evaluate the classifier and sampling combinations.}
\label{fig:metrics}
\end{figure}

\noindent \textbf{Experiments}. The experiments performed include all of the combinations of datasets, classifier and sampling method with 5-fold cross validation. The ANS, MWMOTE and RBO algorithms were not included in the full experiments because of the excessive run time and this limitation is further discussed in Section \ref{sec:fut}.

\begin{table}[t!]
	\centering
	\caption{Classifier hyperparameters used in the sampling experiments.}
	\label{table:classifier_hyperparameters}
	\footnotesize
	\renewcommand{\arraystretch}{1.2}
	\begin{tabular}{ll}
		\toprule
		\textbf{Algorithm} & \textbf{Hyperparameters} \\
		\midrule
		\multirow{2}{*}{Support Vector Machine}&\textit{number of iterations} = 100\\
		&\textit{regularization parameter (C)} = 10.0\\
		\hline
		\multirow{2}{*}{Random Forest }&\textit{number of trees} = 100\\
		&\textit{maximum depth} = 20\\
		\hline		
		\multirow{1}{*}{Naive Bayes}&None\\
		\hline
		\bottomrule
	\end{tabular}
\end{table}

\begin{table}[h!]
	\centering
	\caption{Comparison of 11 oversampling methods using four different metrics with Random Forest as the base classifier.}
	\label{tab: resnb}
	\renewcommand{\arraystretch}{1.25}
	\begin{adjustbox} {center, max width = 115mm}
		\scalebox{0.48} {
			\begin{tabular}{llcccccccccccc}\toprule
				& & ADASYN & Borderline SMOTE & CCR & Cluster SMOTE & Gaussian SMOTE & \textit{k}-Means SMOTE & NRAS & Random Oversample & Safe Level SMOTE & SMOTE & SMOTE-D\\
				\midrule
				
				Bitcoin & AvAcc & 98.08 & 98.56 & \textbf{98.60} & 96.02 & 95.32 & 97.88 & 98.59 & 96.35 & 97.42 & 96.32 & 96.74\\
				& MAvG & 48.69 & 68.14 & \textbf{75.80} & 40.51 & 38.52 & 43.86 & 71.02 & 42.07 & 38.28 & 41.16 & 40.29\\
				& AvFb & 58.03 & 55.22 & 53.53 & 65.26 & 65.14 & 57.60 & 53.09 & \textbf{65.64} & 57.75 & 64.81 & 62.32\\
				& CBA & 57.28 & 53.89 & 52.36 & 56.66 & 55.47 & 57.15 & 52.00 & \textbf{57.49} & 56.78 & 57.11 & 57.04\\
				
				Cover Type & AvAcc & 95.00 & \textbf{95.21} & 93.54 & 94.86 & 94.34 & 94.69 & 94.87 & 94.78 & 94.40 & 94.74 & 94.76\\
				& MAvG & 67.87 & \textbf{81.76} & 35.15 & 67.62 & 61.40 & 61.76 & 67.84 & 66.39 & 61.40 & 65.78 & 65.94\\
				& AvFb & \textbf{81.39} & 77.21 & 41.31 & 81.03 & 79.06 & 70.40 & 78.17 & 80.69 & 77.78 & 80.69 & 80.94\\
				& CBA & 68.84 & \textbf{73.28} & 36.19 & 68.82 & 63.29 & 64.10 & 68.74 & 67.85 & 63.21 & 67.33 & 67.49\\
				
				Fuzzing & AvAcc & 99.99 & 99.98 & 99.87 & 99.98 & 99.99 & 99.99 & 99.87 & 99.99 & 99.94 & \textbf{100.00} & 99.99\\
				& MAvG & 99.97 & 99.96 & 99.68 & 99.95 & 99.99 & 99.99 & 99.67 & 99.99 & 99.83 & \textbf{99.99} & 99.99\\
				& AvFb & 99.99 & 99.97 & 99.86 & 99.98 & \textbf{99.99} & 99.98 & 99.87 & 99.98 & 99.93 & 99.99 & 99.99\\
				& CBA & 99.96 & 99.95 & 99.61 & 99.94 & 99.98 & 99.97 & 99.59 & 99.97 & 99.79 & \textbf{99.99} & 99.98\\
				
				HIGGS: 4:1 & AvAcc & 77.53 & 74.38 & 80.58 & 77.64 & 79.68 & 77.50 & 78.63 & \textbf{82.21} & 72.02 & 76.35 & 74.83\\
				& MAvG & 62.68 & 60.14 & \textbf{76.93} & 62.86 & 65.70 & 60.82 & 63.54 & 70.90 & 57.42 & 61.61 & 59.88\\
				& AvFb & 67.81 & 67.66 & 50.79 & 67.87 & \textbf{68.88} & 63.63 & 66.65 & 65.95 & 65.30 & 67.85 & 66.80\\
				& CBA & 64.21 & 59.16 & 43.13 & 64.41 & \textbf{68.02} & 61.35 & 66.37 & 61.39 & 56.30 & 62.19 & 60.03\\
				
				HIGGS: 8:1 & AvAcc & 84.91 & 86.69 & 88.99 & 82.64 & 85.05 & 83.29 & 87.81 & \textbf{89.06} & 76.46 & 80.97 & 79.02\\
				& MAvG & 56.03 & 59.89 & \textbf{81.16} & 53.63 & 57.73 & 48.99 & 60.32 & 68.65 & 46.00 & 52.01 & 49.81\\
				& AvFb & 64.19 & 64.31 & 49.70 & 65.04 & \textbf{66.39} & 59.00 & 58.33 & 61.18 & 60.64 & 64.90 & 63.72\\
				& CBA & 62.40 & \textbf{63.47} & 45.27 & 59.12 & 63.05 & 57.55 & 54.92 & 57.29 & 51.51 & 56.98 & 54.61\\
				
				HIGGS: 16:1 & AvAcc & 91.56 & 94.01 & \textbf{94.13} & 86.93 & 89.21 & 88.73 & 93.92 & 93.82 & 85.42 & 85.08 & 83.48\\
				& MAvG & 49.53 & \textbf{66.11} & 33.56 & 43.38 & 47.89 & 38.45 & 47.54 & 63.71 & 36.99 & 41.54 & 39.66\\
				& AvFb & 59.66 & 55.34 & 49.51 & 61.91 & \textbf{63.60} & 56.39 & 50.43 & 57.48 & 57.25 & 61.46 & 60.35\\
				& CBA & \textbf{59.17} & 52.58 & 47.17 & 54.76 & 58.15 & 54.28 & 48.06 & 54.84 & 51.64 & 52.83 & 51.16\\
				
				IoT & AvAcc & 99.68 & 99.78 & 99.77 & \textbf{99.85} & 99.74 & 99.83 & 99.71 & 99.64 & 99.78 & 99.67 & 99.69\\
				& MAvG & 98.86 & 99.14 & 99.01 & \textbf{99.42} & 98.98 & 99.33 & 98.93 & 98.63 & 99.13 & 98.78 & 98.93\\
				& AvFb & 98.66 & 99.15 & 98.95 & \textbf{99.40} & 98.96 & 99.29 & 98.82 & 98.57 & 99.15 & 98.68 & 98.74\\
				& CBA & 97.61 & 98.37 & 98.11 & \textbf{98.86} & 98.03 & 98.69 & 97.84 & 97.34 & 98.36 & 97.57 & 97.74\\
				
				MIRAI & AvAcc & \textbf{99.91} & 99.83 & 99.62 & 99.87 & 99.87 & 99.81 & 99.84 & 99.90 & 99.87 & 99.86 & 99.85\\
				& MAvG & 99.72 & 99.60 & 98.90 & 99.74 & 99.68 & 99.67 & 99.60 & \textbf{99.79} & 99.60 & 99.71 & 99.70\\
				& AvFb & \textbf{99.89} & 99.75 & 99.55 & 99.76 & 99.81 & 99.61 & 99.77 & 99.82 & 99.84 & 99.76 & 99.72\\
				& CBA & 99.67 & 99.55 & 98.72 & 99.70 & 99.64 & 99.55 & 99.55 & \textbf{99.78} & 99.53 & 99.70 & 99.64\\
				
				Poker: 0 vs 2 & AvAcc & 97.41 & 92.60 & 91.50 & 97.15 & 91.32 & 94.33 & 95.62 & \textbf{99.87} & 99.37 & 97.56 & 96.92\\
				& MAvG & 98.61 & 96.18 & 95.65 & 96.38 & 0.00 & 84.69 & 97.67 & \textbf{99.93} & 99.66 & 98.69 & 98.36\\
				& AvFb & 87.00 & 58.01 & 50.38 & 86.33 & 49.07 & 72.98 & 76.89 & \textbf{99.37} & 97.00 & 87.77 & 84.24\\
				& CBA & 83.71 & 53.59 & 46.78 & 83.45 & 45.66 & 69.29 & 72.45 & \textbf{99.16} & 96.05 & 84.64 & 80.60\\
				
				Poker: 0 vs 3 & AvAcc & 96.35 & 96.10 & 96.01 & 96.76 & 95.96 & 95.71 & 97.04 & \textbf{99.27} & 98.96 & 96.06 & 96.14\\
				& MAvG & 98.15 & 98.03 & 78.39 & 93.61 & 0.00 & 60.11 & 98.35 & \textbf{99.62} & 99.46 & 98.01 & 98.05\\
				& AvFb & 55.59 & 51.72 & 50.37 & 61.76 & 49.58 & 51.85 & 65.38 & \textbf{92.37} & 89.04 & 51.16 & 52.42\\
				& CBA & 53.08 & 49.76 & 48.63 & 58.71 & 47.98 & 50.06 & 61.98 & \textbf{90.53} & 86.61 & 49.30 & 50.37\\
				
				Poker: 0 vs 4 & AvAcc & 99.23 & 99.23 & 99.23 & 99.03 & \textbf{99.23} & 99.22 & 96.65 & 95.86 & 98.90 & 99.23 & 99.23\\
				& MAvG & 0.00 & 0.00 & 0.00 & \textbf{63.96} & 0.00 & 5.14 & 9.64 & 32.87 & 41.41 & 34.01 & 0.00\\
				& AvFb & 49.92 & 49.92 & 49.92 & 56.21 & 49.92 & 50.00 & 52.91 & \textbf{64.21} & 51.10 & 50.16 & 49.92\\
				& CBA & 49.62 & 49.62 & 49.62 & \textbf{54.08} & 49.62 & 49.68 & 49.55 & 53.49 & 50.34 & 49.81 & 49.62\\
				
				Poker: 0 vs 5 & AvAcc & 99.60 & 99.60 & 99.87 & 99.60 & 99.60 & 99.57 & 99.76 & \textbf{99.96} & 99.93 & 99.56 & 99.61\\
				& MAvG & 39.92 & 0.00 & 99.94 & 90.94 & 0.00 & 42.75 & 99.54 & \textbf{99.98} & 99.97 & 23.00 & 39.92\\
				& AvFb & 50.27 & 49.96 & 86.18 & 57.08 & 49.96 & 51.92 & 71.04 & \textbf{96.34} & 92.64 & 50.24 & 50.28\\
				& CBA & 50.06 & 49.80 & 84.14 & 55.84 & 49.80 & 51.45 & 69.45 & \textbf{95.56} & 91.26 & 50.05 & 50.06\\
				
				Poker: 0 vs 6 & AvAcc & 99.72 & 99.73 & 99.72 & 99.72 & 99.72 & 99.72 & \textbf{99.91} & 99.78 & 99.74 & 99.72 & 99.72\\
				& MAvG & 19.97 & 99.87 & 0.00 & 39.95 & 0.00 & 39.95 & 79.98 & \textbf{99.89} & 98.13 & 39.94 & 19.97\\
				& AvFb & 50.19 & 53.44 & 49.97 & 51.06 & 49.97 & 50.85 & \textbf{85.05} & 63.55 & 55.33 & 50.41 & 50.19\\
				& CBA & 50.03 & 52.69 & 49.86 & 50.74 & 49.86 & 50.57 & \textbf{83.97} & 61.39 & 54.28 & 50.21 & 50.04\\
				
				Poker: 1 vs 2 & AvAcc & 89.88 & 89.87 & 89.87 & 89.89 & 89.87 & 87.60 & 89.86 & \textbf{91.00} & 89.92 & 89.87 & 89.88\\
				& MAvG & 71.57 & 0.00 & 0.00 & 80.00 & 0.00 & 29.94 & 55.25 & 87.05 & \textbf{92.92} & 56.88 & 89.25\\
				& AvFb & 48.95 & 48.90 & 48.90 & 49.46 & 48.90 & 50.48 & 48.93 & \textbf{57.27} & 49.25 & 48.92 & 48.94\\
				& CBA & 44.98 & 44.94 & 44.94 & 45.44 & 44.94 & 47.47 & 44.97 & \textbf{52.52} & 45.24 & 44.95 & 44.97\\
				
				Poker: 1 vs 3 & AvAcc & 95.26 & 95.24 & 95.24 & 95.20 & 95.24 & 94.84 & 95.26 & \textbf{96.10} & 95.66 & 95.24 & 95.24\\
				& MAvG & 78.08 & 19.52 & 0.00 & 83.61 & 0.00 & 32.38 & 94.99 & 96.14 & \textbf{97.80} & 19.52 & 19.52\\
				& AvFb & 49.74 & 49.52 & 49.51 & 50.22 & 49.51 & 50.02 & 49.81 & \textbf{60.76} & 55.00 & 49.52 & 49.53\\
				& CBA & 47.82 & 47.63 & 47.62 & 48.25 & 47.62 & 48.23 & 47.87 & \textbf{57.44} & 52.29 & 47.63 & 47.64\\
				
				Poker: 1 vs 4 & AvAcc & 99.09 & 99.09 & 99.09 & 98.77 & \textbf{99.09} & 99.05 & 99.09 & 96.24 & 99.06 & 99.08 & 99.09\\
				& MAvG & 0.00 & 0.00 & 0.00 & 52.58 & 0.00 & 19.46 & 0.00 & 38.34 & \textbf{54.83} & 28.81 & 28.81\\
				& AvFb & 49.91 & 49.91 & 49.91 & 58.61 & 49.91 & 50.15 & 49.91 & \textbf{67.69} & 51.30 & 50.11 & 50.04\\
				& CBA & 49.54 & 49.54 & 49.54 & \textbf{57.32} & 49.54 & 49.77 & 49.54 & 55.72 & 50.71 & 49.71 & 49.66\\
				
				Poker: 1 vs 5 & AvAcc & 99.53 & 99.53 & \textbf{99.94} & 99.59 & 99.53 & 99.53 & 99.56 & 99.94 & 99.53 & 99.53 & 99.53\\
				& MAvG & 0.00 & 0.00 & \textbf{99.97} & 96.13 & 0.00 & 0.00 & 59.87 & 99.97 & 19.95 & 0.00 & 0.00\\
				& AvFb & 49.95 & 49.95 & \textbf{94.78} & 56.68 & 49.95 & 49.95 & 53.85 & 94.42 & 50.09 & 49.95 & 49.95\\
				& CBA & 49.77 & 49.77 & \textbf{93.69} & 55.75 & 49.77 & 49.76 & 53.08 & 93.27 & 49.87 & 49.77 & 49.77\\
				
				Poker: 1 vs 6 & AvAcc & 99.66 & 99.66 & 99.66 & 99.66 & 99.66 & 99.65 & \textbf{99.67} & 99.66 & 99.66 & 99.66 & 99.66\\
				& MAvG & 0.00 & 0.00 & 0.00 & 0.00 & 0.00 & 0.00 & \textbf{74.02} & 39.93 & 0.00 & 0.00 & 0.00\\
				& AvFb & 49.97 & 49.97 & 49.97 & 49.97 & 49.97 & 49.96 & \textbf{51.07} & 50.52 & 49.97 & 49.97 & 49.97\\
				& CBA & 49.83 & 49.83 & 49.83 & 49.83 & 49.83 & 49.83 & \textbf{50.73} & 50.28 & 49.83 & 49.83 & 49.83\\
				
				SEER & AvAcc & 93.71 & 93.76 & 92.18 & 93.77 & 93.73 & 93.88 & \textbf{94.30} & 93.94 & 93.94 & 93.74 & 93.99\\
				& MAvG & 55.73 & 56.50 & 0.00 & 56.43 & 55.98 & 57.58 & \textbf{60.83} & 57.12 & 57.58 & 55.90 & 57.30\\
				& AvFb & 61.30 & 60.55 & 37.32 & 61.09 & 61.66 & 58.27 & 61.35 & \textbf{62.29} & 62.08 & 61.48 & 60.59\\
				& CBA & 57.09 & 57.51 & 30.72 & 56.62 & 56.82 & 53.47 & 57.51 & 58.13 & \textbf{58.65} & 56.94 & 55.64\\
				
				SSL Renegotiation & AvAcc & 100.00 & 100.00 & 99.99 & 100.00 & 100.00 & 99.99 & 99.98 & 100.00 & 99.95 & 100.00 & \textbf{100.00}\\
				& MAvG & 99.96 & 99.96 & 99.94 & 99.98 & 99.96 & 99.97 & 99.73 & \textbf{99.98} & 99.43 & 99.98 & 99.96\\
				& AvFb & 99.98 & 99.98 & 99.98 & 99.98 & 99.98 & 99.97 & 99.93 & 99.98 & 99.86 & 99.98 & \textbf{99.99}\\
				& CBA & 99.95 & 99.95 & 99.93 & 99.96 & 99.95 & 99.94 & 99.72 & 99.96 & 99.41 & 99.96 & \textbf{99.96}\\
				
				SUSY: 4:1 & AvAcc & 85.12 & 84.24 & 84.87 & 84.31 & 86.24 & 84.39 & 86.75 & \textbf{87.01} & 83.45 & 83.96 & 83.25\\
				& MAvG & 75.22 & 73.68 & \textbf{85.96} & 73.80 & 77.53 & 74.08 & 79.03 & 80.01 & 72.46 & 73.24 & 72.17\\
				& AvFb & 78.09 & 78.10 & 64.40 & 77.78 & \textbf{78.19} & 76.24 & 77.68 & 77.26 & 77.56 & 77.70 & 77.44\\
				& CBA & 75.63 & 73.34 & 56.72 & 73.74 & \textbf{77.87} & 75.11 & 76.01 & 74.80 & 71.80 & 72.93 & 71.42\\
				
				SUSY: 8:1 & AvAcc & 90.34 & 90.92 & 90.91 & 87.52 & 89.81 & 87.92 & 91.50 & \textbf{91.70} & 86.81 & 86.92 & 86.20\\
				& MAvG & 73.09 & 75.11 & \textbf{89.38} & 66.00 & 71.41 & 66.56 & 77.95 & 79.46 & 64.67 & 64.85 & 63.68\\
				& AvFb & 75.74 & 76.15 & 61.37 & 75.62 & \textbf{76.41} & 74.41 & 75.13 & 74.27 & 75.27 & 75.26 & 75.13\\
				& CBA & 75.41 & \textbf{75.63} & 56.11 & 68.00 & 73.57 & 69.07 & 73.21 & 71.65 & 66.61 & 66.82 & 65.46\\
				
				SUSY: 16:1 & AvAcc & 94.25 & 94.95 & 94.95 & 90.41 & 92.36 & 91.61 & \textbf{95.08} & 94.80 & 91.03 & 89.70 & 89.08\\
				& MAvG & 70.36 & 75.86 & \textbf{91.77} & 56.65 & 62.25 & 59.12 & 77.46 & 74.87 & 57.96 & 55.13 & 53.93\\
				& AvFb & 72.99 & 72.13 & 59.10 & 72.62 & \textbf{74.21} & 71.38 & 71.26 & 70.79 & 72.68 & 72.22 & 71.83\\
				& CBA & \textbf{72.60} & 70.29 & 55.60 & 62.60 & 67.16 & 65.07 & 68.99 & 68.83 & 63.79 & 61.30 & 60.26\\
				
				SYN DOS & AvAcc & 100.00 & 99.99 & 99.99 & 99.99 & 99.98 & 99.99 & 99.99 & 100.00 & 99.01 & 100.00 & \textbf{100.00}\\
				& MAvG & 99.61 & 99.42 & 99.21 & 99.42 & 95.81 & 98.48 & 99.60 & 99.61 & 45.08 & 99.81 & \textbf{100.00}\\
				& AvFb & 99.76 & 99.25 & 98.41 & 99.41 & 98.95 & 99.22 & 99.12 & 99.61 & 77.41 & 99.49 & \textbf{99.84}\\
				& CBA & 99.42 & 98.64 & 98.03 & 98.84 & 95.91 & 98.50 & 98.82 & 99.41 & 59.68 & 99.22 & \textbf{99.80}\\
				
				Traffic & AvAcc & 84.87 & 84.85 & 80.06 & 84.70 & 84.75 & 84.36 & 84.68 & \textbf{85.10} & 84.82 & 84.81 & 84.79\\
				& MAvG & 45.07 & 45.94 & 14.89 & 43.07 & 42.83 & 42.20 & \textbf{46.62} & 45.28 & 44.11 & 43.34 & 43.27\\
				& AvFb & 43.55 & 43.19 & 18.03 & 44.26 & 44.36 & 41.08 & 41.39 & \textbf{45.64} & 43.37 & 44.34 & 44.34\\
				& CBA & 40.20 & 39.58 & 12.86 & 40.96 & 40.75 & 37.81 & 37.77 & \textbf{42.12} & 40.20 & 41.33 & 41.42\\
				
				Video Injection & AvAcc & 99.99 & 99.99 & \textbf{100.00} & 99.99 & 99.98 & 99.99 & 99.99 & 99.99 & 99.97 & 99.99 & 99.99\\
				& MAvG & 99.93 & 99.93 & \textbf{99.97} & 99.88 & 99.81 & 99.93 & 99.95 & 99.91 & 99.63 & 99.88 & 99.89\\
				& AvFb & 99.96 & 99.94 & 99.97 & 99.95 & 99.95 & 99.93 & \textbf{99.97} & 99.96 & 99.90 & 99.94 & 99.95\\
				& CBA & 99.90 & 99.91 & \textbf{99.96} & 99.87 & 99.80 & 99.87 & 99.92 & 99.90 & 99.61 & 99.86 & 99.87\\
				
				\bottomrule
			\end{tabular}
		}
	\end{adjustbox}
\end{table}

\begin{table}[h!]
	\centering
	\caption{Comparison of 11 oversampling methods using four different metrics with SVM as the base classifier.}
\label{tab: resnb}
	\renewcommand{\arraystretch}{1.25}
	\begin{adjustbox} {center, max width = 115mm}
		\scalebox{0.48} {
			\begin{tabular}{llcccccccccccc}\toprule
				& & ADASYN & Borderline SMOTE & CCR & Cluster SMOTE & Gaussian SMOTE & \textit{k}-Means SMOTE & NRAS & Random Oversample & Safe Level SMOTE & SMOTE & SMOTE-D\\
				\midrule

				Bitcoin & AvAcc & 20.85 & 98.58 & 98.58 & 9.79 & \textbf{98.58} & 98.58 & 80.16 & 59.72 & 98.58 & 41.34 & 98.58\\
				 & MAvG & 2.38 & 0.00 & 0.00 & 12.33 & 0.00 & 0.00 & 13.07 & 2.38 & 0.00 & \textbf{13.59} & 0.00\\
				 & AvFb & 12.66 & 49.86 & 49.86 & 8.62 & \textbf{49.86} & 49.86 & 45.23 & 31.26 & 49.86 & 27.37 & 49.86\\
				 & CBA & 10.43 & 49.29 & 49.29 & 5.03 & \textbf{49.29} & 49.29 & 41.36 & 29.86 & 49.29 & 21.34 & 49.29\\

				Cover Type & AvAcc & 81.40 & \textbf{89.35} & 85.36 & 83.95 & 83.27 & 83.98 & 83.27 & 82.87 & 83.62 & 83.11 & 83.04\\
				 & MAvG & 22.76 & 4.75 & 0.00 & 25.58 & \textbf{26.43} & 18.26 & 24.56 & 25.07 & 25.93 & 24.26 & 24.65\\
				 & AvFb & 33.03 & 32.25 & 11.80 & 38.85 & \textbf{40.52} & 31.73 & 38.72 & 38.19 & 39.80 & 37.88 & 37.45\\
				 & CBA & 20.25 & 25.53 & 6.97 & 25.70 & \textbf{26.54} & 22.95 & 24.68 & 24.13 & 25.49 & 24.64 & 24.22\\

				Fuzzing & AvAcc & 60.55 & \textbf{80.99} & 80.72 & 80.94 & 71.80 & 80.90 & 58.78 & 59.19 & 62.19 & 62.79 & 43.98\\
				 & MAvG & 57.43 & \textbf{83.86} & 0.00 & 78.45 & 63.66 & 74.49 & 56.46 & 56.64 & 58.07 & 58.42 & 50.62\\
				 & AvFb & 63.77 & 48.79 & 47.72 & 50.48 & \textbf{73.59} & 48.82 & 62.24 & 62.60 & 65.19 & 65.73 & 49.38\\
				 & CBA & 42.15 & 41.33 & 40.36 & 43.21 & \textbf{52.87} & 41.41 & 40.46 & 40.82 & 43.49 & 44.07 & 28.11\\

				HIGGS: 4:1 & AvAcc & 20.00 & 80.00 & 80.00 & 79.77 & 80.00 & 79.15 & \textbf{80.00} & 44.00 & 80.00 & 80.00 & 80.00\\
				 & MAvG & 0.00 & 0.00 & 0.00 & 17.43 & 0.00 & 20.56 & \textbf{30.61} & 0.00 & 0.00 & 0.00 & 0.00\\
				 & AvFb & 27.78 & 47.62 & 47.62 & 47.83 & 47.62 & \textbf{48.27} & 47.64 & 35.71 & 47.62 & 47.62 & 47.62\\
				 & CBA & 10.00 & 40.00 & 40.00 & 40.32 & 40.00 & \textbf{41.13} & 40.02 & 22.00 & 40.00 & 40.00 & 40.00\\

				HIGGS: 8:1 & AvAcc & 11.11 & 88.89 & 88.89 & 88.77 & \textbf{88.89} & 86.66 & 88.40 & 73.33 & 88.89 & 88.89 & 88.89\\
				 & MAvG & 0.00 & 0.00 & 0.00 & 25.31 & 0.00 & 23.47 & \textbf{54.21} & 0.00 & 0.00 & 0.00 & 0.00\\
				 & AvFb & 19.23 & 48.78 & 48.78 & 48.96 & 48.78 & \textbf{50.08} & 49.90 & 42.87 & 48.78 & 48.78 & 48.78\\
				 & CBA & 5.56 & 44.44 & 44.44 & 44.65 & 44.44 & \textbf{46.76} & 45.71 & 36.67 & 44.44 & 44.44 & 44.44\\

			    HIGGS: 16:1 & AvAcc & 76.47 & 94.12 & 94.12 & 94.09 & \textbf{94.12} & 90.62 & 71.04 & 41.18 & 94.12 & 94.12 & 94.12\\
				 & MAvG & 0.00 & 0.00 & 0.00 & 24.37 & 0.00 & 30.08 & \textbf{32.74} & 0.00 & 0.00 & 0.00 & 0.00\\
				 & AvFb & 41.89 & 49.38 & 49.38 & 49.44 & 49.38 & \textbf{49.85} & 44.05 & 26.90 & 49.38 & 49.38 & 49.38\\
				 & CBA & 38.24 & 47.06 & 47.06 & \textbf{47.12} & 47.06 & 47.12 & 39.52 & 20.59 & 47.06 & 47.06 & 47.06\\

				IoT & AvAcc & 89.93 & 92.91 & 88.49 & 93.60 & 90.26 & 92.76 & 90.39 & 91.39 & 91.71 & 91.06 & \textbf{93.69}\\
				 & MAvG & 0.00 & 0.00 & 0.00 & 0.00 & 0.00 & 0.00 & 0.00 & 0.00 & 0.00 & 0.00 & \textbf{30.78}\\
				 & AvFb & 39.99 & 54.87 & 27.24 & 62.68 & 45.55 & 53.46 & 47.19 & 52.24 & 55.30 & 51.55 & \textbf{62.71}\\
				 & CBA & 29.25 & 45.74 & 19.24 & 52.27 & 35.75 & 42.45 & 37.28 & 41.92 & 45.35 & 41.65 & \textbf{52.70}\\

				MIRAI & AvAcc & 89.79 & 84.60 & 84.29 & 89.72 & \textbf{90.89} & 87.21 & 89.73 & 89.72 & 89.73 & 89.73 & 89.72\\
				 & MAvG & 78.05 & 75.74 & 72.98 & 77.95 & \textbf{81.52} & 77.87 & 77.97 & 77.95 & 77.96 & 77.96 & 77.94\\
				 & AvFb & \textbf{89.33} & 51.88 & 49.83 & 89.27 & 84.26 & 69.99 & 89.28 & 89.27 & 89.27 & 89.27 & 89.26\\
				 & CBA & 74.38 & 45.56 & 43.60 & 74.26 & \textbf{81.74} & 60.80 & 74.29 & 74.27 & 74.28 & 74.28 & 74.26\\

				Poker: 0 vs 2 & AvAcc & 91.32 & 91.32 & 91.32 & 56.19 & \textbf{91.32} & 91.32 & 54.33 & 74.79 & 91.32 & 51.51 & 91.32\\
				 & MAvG & 0.00 & 0.00 & 0.00 & 27.90 & 0.00 & 0.00 & 27.96 & 0.00 & 0.00 & \textbf{28.14} & 0.00\\
				 & AvFb & 49.07 & 49.07 & 49.07 & 42.73 & \textbf{49.07} & 49.07 & 42.06 & 42.47 & 49.07 & 40.97 & 49.07\\
				 & CBA & 45.66 & 45.66 & 45.66 & 33.05 & \textbf{45.66} & 45.66 & 31.93 & 37.40 & 45.66 & 30.25 & 45.66\\

				Poker: 0 vs 3 & AvAcc & 40.81 & 95.96 & 95.96 & 54.22 & \textbf{95.96} & 95.96 & 57.90 & 59.19 & 95.96 & 52.03 & 95.96\\
				 & MAvG & 0.00 & 0.00 & 0.00 & 19.45 & 0.00 & 0.00 & \textbf{19.76} & 0.00 & 0.00 & 19.09 & 0.00\\
				 & AvFb & 25.06 & 49.58 & 49.58 & 37.15 & \textbf{49.58} & 49.58 & 39.08 & 33.23 & 49.58 & 35.88 & 49.58\\
				 & CBA & 20.40 & 47.98 & 47.98 & 29.29 & \textbf{47.98} & 47.98 & 31.33 & 29.60 & 47.98 & 28.07 & 47.98\\

				Poker: 0 vs 4 & AvAcc & 20.46 & 99.23 & 99.23 & 67.24 & \textbf{99.23} & 99.23 & 82.66 & 59.85 & 99.23 & 66.67 & 99.23\\
				 & MAvG & 0.00 & 0.00 & 0.00 & \textbf{12.06} & 0.00 & 0.00 & 9.63 & 0.00 & 0.00 & 11.89 & 0.00\\
				 & AvFb & 11.48 & 49.92 & 49.92 & 39.31 & \textbf{49.92} & 49.92 & 44.62 & 30.70 & 49.92 & 38.96 & 49.92\\
				 & CBA & 10.23 & 49.62 & 49.62 & 34.37 & \textbf{49.62} & 49.62 & 42.06 & 29.92 & 49.62 & 34.06 & 49.62\\

				Poker: 0 vs 5 & AvAcc & 59.92 & 99.60 & 79.76 & 62.00 & \textbf{99.60} & 99.60 & 66.87 & 40.08 & 99.60 & 59.04 & 99.60\\
				 & MAvG & 0.00 & 0.00 & 0.00 & 6.32 & 0.00 & 0.00 & \textbf{7.12} & 0.00 & 0.00 & 6.97 & 0.00\\
				 & AvFb & 30.36 & 49.96 & 40.16 & 34.53 & \textbf{49.96} & 49.96 & 37.07 & 20.57 & 49.96 & 33.33 & 49.96\\
				 & CBA & 29.96 & 49.80 & 39.88 & 31.25 & \textbf{49.80} & 49.80 & 33.74 & 20.04 & 49.80 & 29.78 & 49.80\\

				Poker: 0 vs 6 & AvAcc & 40.06 & 99.72 & 99.72 & 63.10 & \textbf{99.72} & 99.72 & 64.57 & 59.94 & 99.72 & 59.75 & 99.72\\
				 & MAvG & 0.00 & 0.00 & 0.00 & 5.38 & 0.00 & 0.00 & \textbf{5.86} & 0.00 & 0.00 & 5.17 & 0.00\\
				 & AvFb & 20.41 & 49.97 & 49.97 & 34.78 & \textbf{49.97} & 49.97 & 35.17 & 30.26 & 49.97 & 33.16 & 49.97\\
				 & CBA & 20.03 & 49.86 & 49.86 & 31.73 & \textbf{49.86} & 49.86 & 32.51 & 29.97 & 49.86 & 30.04 & 49.86\\

				Poker: 1 vs 2 & AvAcc & 26.08 & 89.87 & 89.87 & 52.24 & \textbf{89.87} & 89.87 & 56.10 & 73.92 & 89.87 & 51.54 & 89.87\\
				 & MAvG & 0.00 & 0.00 & 0.00 & 30.05 & 0.00 & 0.00 & 30.10 & 0.00 & 0.00 & \textbf{30.31} & 0.00\\
				 & AvFb & 24.20 & 48.90 & 48.90 & 42.30 & \textbf{48.90} & 48.90 & 43.94 & 42.72 & 48.90 & 42.21 & 48.90\\
				 & CBA & 13.04 & 44.94 & 44.94 & 31.46 & \textbf{44.94} & 44.94 & 33.88 & 36.96 & 44.94 & 31.05 & 44.94\\

				Poker: 1 vs 3 & AvAcc & 22.86 & 95.24 & 95.24 & 54.68 & \textbf{95.24} & 95.24 & 60.66 & 40.95 & 95.24 & 52.50 & 95.24\\
				 & MAvG & 0.00 & 0.00 & 0.00 & 21.50 & 0.00 & 0.00 & \textbf{21.79} & 0.00 & 0.00 & 21.28 & 0.00\\
				 & AvFb & 17.90 & 49.51 & 49.51 & 38.60 & \textbf{49.51} & 49.51 & 41.50 & 25.80 & 49.51 & 37.44 & 49.51\\
				 & CBA & 11.43 & 47.62 & 47.62 & 29.99 & \textbf{47.62} & 47.62 & 33.33 & 20.48 & 47.62 & 28.76 & 47.62\\

				Poker: 1 vs 4 & AvAcc & 40.18 & 99.09 & 99.09 & 60.92 & \textbf{99.09} & 99.09 & 92.53 & 40.18 & 99.09 & 60.21 & 99.09\\
				 & MAvG & 0.00 & 0.00 & 0.00 & \textbf{12.11} & 0.00 & 0.00 & 11.40 & 0.00 & 0.00 & 12.01 & 0.00\\
				 & AvFb & 21.28 & 49.91 & 49.91 & 36.38 & \textbf{49.91} & 49.91 & 48.97 & 21.28 & 49.91 & 35.99 & 49.91\\
				 & CBA & 20.09 & 49.54 & 49.54 & 31.19 & \textbf{49.54} & 49.54 & 47.29 & 20.09 & 49.54 & 30.81 & 49.54\\

				Poker: 1 vs 5 & AvAcc & 59.91 & 99.53 & 99.53 & 57.71 & \textbf{99.53} & 99.53 & 59.03 & 59.91 & 99.53 & 57.32 & 99.53\\
				 & MAvG & 0.00 & 0.00 & 0.00 & 6.06 & 0.00 & 0.00 & \textbf{7.21} & 0.00 & 0.00 & 6.62 & 0.00\\
				 & AvFb & 30.43 & 49.95 & 49.95 & 32.43 & \textbf{49.95} & 49.95 & 33.41 & 30.43 & 49.95 & 32.41 & 49.95\\
				 & CBA & 29.95 & 49.77 & 49.77 & 29.10 & \textbf{49.77} & 49.77 & 29.81 & 29.95 & 49.77 & 28.92 & 49.77\\

				Poker: 1 vs 6 & AvAcc & 59.93 & 99.66 & 99.66 & 59.86 & \textbf{99.66} & 99.66 & 55.57 & 40.07 & 99.66 & 57.40 & 99.66\\
				 & MAvG & 0.00 & 0.00 & 0.00 & 5.41 & 0.00 & 0.00 & 5.31 & 0.00 & 0.00 & \textbf{5.45} & 0.00\\
				 & AvFb & 30.31 & 49.97 & 49.97 & 33.28 & \textbf{49.97} & 49.97 & 31.20 & 20.48 & 49.97 & 32.12 & 49.97\\
				 & CBA & 29.97 & 49.83 & 49.83 & 30.12 & \textbf{49.83} & 49.83 & 27.96 & 20.03 & 49.83 & 28.88 & 49.83\\

				SEER & AvAcc & 89.00 & \textbf{90.01} & 89.21 & 88.92 & 89.01 & 89.26 & 89.20 & 88.24 & 89.28 & 89.24 & 89.37\\
				 & MAvG & 32.82 & 33.75 & 0.00 & 31.09 & 34.42 & 26.12 & \textbf{37.15} & 31.57 & 35.23 & 33.45 & 32.20\\
				 & AvFb & 38.57 & 39.91 & 18.53 & 37.63 & 40.01 & 32.48 & \textbf{42.33} & 37.45 & 41.07 & 39.90 & 37.87\\
				 & CBA & 31.26 & \textbf{35.54} & 11.39 & 31.56 & 32.36 & 27.11 & 33.47 & 29.98 & 33.96 & 32.92 & 31.47\\

				SSL Renegotiation & AvAcc & 90.08 & 95.80 & \textbf{95.80} & 83.51 & 90.12 & 89.35 & 90.10 & 90.08 & 90.11 & 90.10 & 83.51\\
				 & MAvG & 54.04 & 0.00 & 0.00 & 45.04 & \textbf{54.11} & 16.13 & 54.08 & 54.06 & 54.10 & 54.08 & 45.04\\
				 & AvFb & 78.83 & 49.57 & 49.57 & 70.87 & 78.87 & 48.79 & 78.85 & 78.84 & \textbf{78.87} & 78.85 & 70.87\\
				 & CBA & 59.53 & 47.90 & 47.90 & 51.54 & \textbf{59.60} & 47.92 & 59.57 & 59.54 & 59.59 & 59.56 & 51.54\\

				SUSY: 4:1 & AvAcc & \textbf{81.12} & 80.27 & 80.00 & 80.32 & 80.07 & 80.06 & 80.63 & 80.97 & 80.19 & 80.36 & 80.31\\
				 & MAvG & 78.28 & \textbf{79.26} & 0.00 & 78.27 & 78.82 & 69.25 & 79.22 & 77.91 & 79.06 & 78.09 & 78.46\\
				 & AvFb & \textbf{53.06} & 48.76 & 47.62 & 49.03 & 47.90 & 48.18 & 50.32 & 52.70 & 48.44 & 49.22 & 48.97\\
				 & CBA & \textbf{45.79} & 41.06 & 40.00 & 41.32 & 40.26 & 40.55 & 42.54 & 45.55 & 40.76 & 41.50 & 41.26\\

				SUSY: 8:1 & AvAcc & 89.11 & 88.89 & 88.89 & 89.04 & 88.90 & 88.97 & \textbf{89.17} & 73.99 & 88.96 & 89.05 & 89.04\\
				 & MAvG & \textbf{77.98} & 37.71 & 0.00 & 76.89 & 70.77 & 69.15 & 76.78 & 68.78 & 76.83 & 76.29 & 77.42\\
				 & AvFb & 51.08 & 48.79 & 48.78 & 50.40 & 48.95 & 50.05 & \textbf{51.89} & 46.24 & 49.51 & 50.60 & 50.35\\
				 & CBA & 46.53 & 44.45 & 44.44 & 45.91 & 44.60 & 45.60 & \textbf{47.28} & 39.78 & 45.10 & 46.09 & 45.86\\

                SUSY: 16:1 & AvAcc & \textbf{94.15} & 94.12 & 94.12 & 94.11 & 94.11 & 94.09 & 94.09 & 88.84 & 94.11 & 94.11 & 94.11\\
				 & MAvG & \textbf{70.10} & 0.00 & 0.00 & 66.97 & 57.65 & 59.90 & 67.21 & 57.18 & 66.20 & 67.79 & 68.03\\
				 & AvFb & 52.64 & 49.38 & 49.38 & 51.05 & 49.51 & 51.08 & 52.91 & \textbf{56.24} & 49.96 & 51.28 & 51.10\\
				 & CBA & 49.96 & 47.06 & 47.06 & 48.53 & 47.17 & 48.58 & \textbf{50.23} & 49.73 & 47.57 & 48.74 & 48.57\\

				SYN DOS & AvAcc & 79.35 & 99.75 & \textbf{99.75} & 77.33 & 79.35 & 99.71 & 99.62 & 72.19 & 70.44 & 71.46 & 99.62\\
				 & MAvG & 9.92 & 0.00 & 0.00 & 17.54 & 9.82 & \textbf{52.10} & 47.77 & 9.56 & 9.23 & 9.39 & 47.66\\
				 & AvFb & 43.73 & 49.97 & 49.97 & 44.79 & 43.68 & 59.08 & \textbf{60.67} & 40.39 & 39.46 & 39.99 & 60.50\\
				 & CBA & 40.17 & 49.87 & 49.87 & 41.51 & 40.16 & 58.16 & \textbf{60.27} & 36.52 & 35.61 & 36.13 & 60.08\\

				Traffic & AvAcc & 78.10 & 78.99 & 78.78 & 77.74 & \textbf{79.13} & 77.68 & 78.13 & 78.29 & 78.91 & 78.37 & 78.77\\
				 & MAvG & 20.40 & \textbf{20.79} & 7.67 & 18.35 & 20.68 & 17.89 & 19.85 & 19.91 & 20.40 & 19.46 & 20.37\\
				 & AvFb & 20.84 & \textbf{21.90} & 14.03 & 19.71 & 21.27 & 19.31 & 21.38 & 20.90 & 21.83 & 21.03 & 20.95\\
				 & CBA & 16.67 & 18.33 & 9.05 & 16.12 & 16.65 & 16.11 & \textbf{18.36} & 17.13 & 18.22 & 17.61 & 16.48\\

				Video Injection & AvAcc & 31.77 & 95.85 & \textbf{95.85} & 54.22 & 54.69 & 75.50 & 54.41 & 53.11 & 54.30 & 42.63 & 71.12\\
				 & MAvG & 18.45 & 0.00 & 0.00 & \textbf{34.18} & 20.81 & 19.04 & 20.68 & 20.47 & 20.65 & 25.60 & 24.62\\
				 & AvFb & 27.18 & 49.57 & 49.57 & 40.02 & 39.57 & 47.83 & 39.31 & 38.45 & 39.23 & 35.79 & \textbf{50.40}\\
				 & CBA & 17.27 & 47.93 & \textbf{47.93} & 29.70 & 29.14 & 40.00 & 28.97 & 28.23 & 28.91 & 23.42 & 39.11\\

				\bottomrule
			\end{tabular}
		}
	\end{adjustbox}
\end{table}

\begin{table}[h!]
	\centering
	\caption{Comparison of 11 oversampling methods using four different metrics with Naive Bayes as the base classifier.}

\label{tab: resnb}
	\renewcommand{\arraystretch}{1.25}
	\begin{adjustbox} {center, max width = 115mm}
		\scalebox{0.48} {
			\begin{tabular}{llcccccccccccc}\toprule
				& & ADASYN & Borderline SMOTE & CCR & Cluster SMOTE & Gaussian SMOTE & \textit{k}-Means SMOTE & NRAS & Random Oversample & Safe Level SMOTE & SMOTE & SMOTE-D\\
				\midrule

				Bitcoin & AvAcc & 50.83 & 98.58 & 98.58 & 55.62 & \textbf{98.58} & 70.90 & 61.26 & 50.74 & 63.35 & 49.58 & 68.48\\
				 & MAvG & 13.60 & 0.00 & 0.00 & 12.09 & 0.00 & 11.44 & 12.95 & 13.49 & \textbf{14.19} & 13.37 & 11.29\\
				 & AvFb & 32.24 & 49.86 & 49.86 & 33.39 & \textbf{49.86} & 40.61 & 36.88 & 32.14 & 38.60 & 31.49 & 39.46\\
				 & CBA & 26.24 & 49.29 & 49.29 & 28.61 & \textbf{49.29} & 36.43 & 31.58 & 26.19 & 32.77 & 25.58 & 35.18\\

				Cover Type & AvAcc & 83.99 & \textbf{89.57} & 89.14 & 85.78 & 86.14 & 83.40 & 86.48 & 86.15 & 86.27 & 86.09 & 85.64\\
				 & MAvG & 28.59 & \textbf{37.62} & 0.00 & 31.59 & 31.43 & 18.51 & 34.95 & 31.42 & 32.11 & 31.36 & 30.49\\
				 & AvFb & 42.35 & 43.71 & 29.06 & 44.82 & 46.63 & 30.62 & \textbf{49.94} & 46.59 & 47.46 & 46.54 & 45.52\\
				 & CBA & 28.38 & \textbf{38.64} & 23.56 & 31.56 & 32.90 & 24.37 & 36.93 & 32.87 & 33.67 & 32.79 & 31.38\\

				Fuzzing & AvAcc & 66.17 & \textbf{81.14} & 80.99 & 77.95 & 75.43 & 80.60 & 71.06 & 70.63 & 72.22 & 70.80 & 73.75\\
				 & MAvG & 60.21 & 66.44 & \textbf{89.83} & 52.12 & 64.68 & 61.54 & 62.89 & 62.67 & 63.50 & 62.76 & 64.04\\
				 & AvFb & 68.57 & 57.26 & 48.65 & 57.04 & \textbf{74.78} & 48.67 & 72.67 & 72.36 & 73.51 & 72.48 & 74.21\\
				 & CBA & 47.35 & 50.86 & 41.19 & 49.07 & \textbf{57.75} & 41.38 & 52.21 & 51.75 & 53.52 & 51.93 & 55.47\\

				HIGGS: 4:1 & AvAcc & 47.40 & 54.18 & 80.00 & 52.78 & \textbf{80.00} & 57.95 & 53.64 & 53.73 & 60.76 & 53.93 & 54.93\\
				 & MAvG & 42.67 & 43.22 & 0.00 & 41.54 & 0.00 & 41.75 & 42.48 & 43.35 & 43.25 & \textbf{43.44} & 43.16\\
				 & AvFb & 47.06 & 50.19 & 47.62 & 48.30 & 47.62 & 49.77 & 49.40 & 50.14 & \textbf{51.82} & 50.29 & 50.39\\
				 & CBA & 32.62 & 38.50 & 40.00 & 37.34 & 40.00 & 41.81 & 38.06 & 38.10 & \textbf{44.34} & 38.27 & 39.17\\

				HIGGS: 8:1 & AvAcc & 51.73 & 88.89 & 88.89 & 55.34 & \textbf{88.89} & 57.64 & 55.79 & 56.14 & 67.08 & 54.98 & 56.45\\
				 & MAvG & 33.88 & 0.00 & 0.00 & 32.42 & 0.00 & 32.21 & 33.60 & \textbf{34.36} & 33.78 & 34.17 & 33.89\\
				 & AvFb & 44.70 & 48.78 & 48.78 & 45.03 & 48.78 & 45.55 & 46.07 & 46.90 & \textbf{49.80} & 46.30 & 46.65\\
				 & CBA & 31.89 & 44.44 & 44.44 & 34.10 & \textbf{44.44} & 35.57 & 34.56 & 34.85 & 42.01 & 34.06 & 35.00\\

                HIGGS: 16:1 & AvAcc & 57.67 & 94.12 & 94.12 & 57.24 & \textbf{94.12} & 61.44 & 60.88 & 53.16 & 71.95 & 54.55 & 55.13\\
				 & MAvG & 25.20 & 0.00 & 0.00 & 24.21 & 0.00 & 23.78 & 23.93 & 25.57 & \textbf{26.18} & 25.64 & 25.54\\
				 & AvFb & 42.30 & 49.38 & 49.38 & 41.37 & \textbf{49.38} & 42.81 & 42.57 & 40.57 & 48.36 & 41.27 & 41.46\\
				 & CBA & 32.49 & 47.06 & 47.06 & 32.12 & \textbf{47.06} & 34.47 & 34.17 & 29.95 & 40.85 & 30.75 & 31.08\\

				IoT & AvAcc & 92.30 & 92.34 & \textbf{92.83} & 90.91 & 92.25 & 90.59 & 91.35 & 91.88 & 91.82 & 91.88 & 92.41\\
				 & MAvG & 0.00 & 0.00 & 0.00 & 0.00 & \textbf{0.00} & 0.00 & 0.00 & 0.00 & 0.00 & 0.00 & 0.00\\
				 & AvFb & 54.18 & 54.62 & 54.14 & 49.80 & 55.87 & 41.82 & 54.87 & 56.86 & 56.67 & \textbf{56.88} & 56.32\\
				 & CBA & 42.18 & 44.31 & 42.08 & 35.41 & 43.48 & 29.54 & 40.91 & 43.27 & 42.93 & 43.26 & \textbf{46.25}\\

				MIRAI & AvAcc & 89.88 & 89.89 & 89.89 & 89.87 & 89.89 & \textbf{90.38} & 89.88 & 89.88 & 89.88 & 89.88 & 89.86\\
				 & MAvG & 78.19 & 78.20 & 78.20 & 78.18 & 78.20 & \textbf{79.77} & 78.19 & 78.18 & 78.19 & 78.19 & 78.16\\
				 & AvFb & 89.43 & \textbf{89.44} & 89.44 & 89.42 & 89.44 & 87.16 & 89.42 & 89.42 & 89.42 & 89.42 & 89.41\\
				 & CBA & 74.55 & 74.57 & 74.57 & 74.54 & 74.57 & \textbf{77.83} & 74.55 & 74.54 & 74.55 & 74.55 & 74.52\\

				Poker: 0 vs 2 & AvAcc & 56.38 & 91.32 & 91.32 & 57.05 & \textbf{91.32} & 53.57 & 55.49 & 51.67 & 57.26 & 51.81 & 52.25\\
				 & MAvG & \textbf{28.58} & 0.00 & 0.00 & 28.02 & 0.00 & 28.27 & 27.93 & 28.18 & 28.40 & 28.15 & 28.02\\
				 & AvFb & 43.38 & 49.07 & 49.07 & 43.17 & \textbf{49.07} & 41.36 & 42.54 & 41.07 & 43.61 & 41.11 & 41.21\\
				 & CBA & 33.25 & 45.66 & 45.66 & 33.59 & \textbf{45.66} & 31.53 & 32.64 & 30.35 & 33.77 & 30.43 & 30.68\\

				Poker: 0 vs 3 & AvAcc & 53.34 & 95.96 & 95.96 & 55.17 & \textbf{95.96} & 64.65 & 59.76 & 51.78 & 56.91 & 52.53 & 53.83\\
				 & MAvG & 19.36 & 0.00 & 0.00 & 19.41 & 0.00 & 19.13 & \textbf{19.84} & 19.01 & 19.41 & 19.09 & 19.24\\
				 & AvFb & 36.68 & 49.58 & 49.58 & 37.57 & \textbf{49.58} & 41.47 & 39.98 & 35.71 & 38.40 & 36.12 & 36.84\\
				 & CBA & 28.81 & 47.98 & 47.98 & 29.81 & \textbf{47.98} & 34.92 & 32.35 & 27.93 & 30.75 & 28.34 & 29.06\\

				Poker: 0 vs 4 & AvAcc & 73.05 & 99.23 & 99.23 & 70.75 & \textbf{99.23} & 80.08 & 84.94 & 69.03 & 66.73 & 70.16 & 69.53\\
				 & MAvG & 12.28 & 0.00 & 0.00 & \textbf{12.64} & 0.00 & 9.55 & 8.49 & 12.39 & 10.56 & 12.46 & 12.28\\
				 & AvFb & 42.04 & 49.92 & 49.92 & 41.21 & \textbf{49.92} & 44.26 & 45.21 & 40.30 & 38.35 & 40.85 & 40.47\\
				 & CBA & 37.36 & 49.62 & 49.62 & 36.21 & \textbf{49.62} & 40.71 & 43.13 & 35.31 & 34.00 & 35.90 & 35.56\\

				Poker: 0 vs 5 & AvAcc & 61.16 & 99.60 & \textbf{99.60} & 64.46 & 99.60 & 71.16 & 71.43 & 59.90 & 62.31 & 61.08 & 63.18\\
				 & MAvG & 6.80 & 0.00 & 0.00 & 6.50 & \textbf{11.52} & 5.10 & 7.45 & 7.01 & 7.38 & 6.97 & 6.98\\
				 & AvFb & 34.28 & 49.96 & 49.96 & 35.71 & \textbf{50.11} & 38.40 & 39.25 & 33.75 & 35.01 & 34.30 & 35.29\\
				 & CBA & 30.85 & 49.80 & 49.80 & 32.49 & \textbf{49.93} & 35.79 & 36.06 & 30.21 & 31.45 & 30.81 & 31.87\\

				Poker: 0 vs 6 & AvAcc & 78.69 & 99.72 & 99.72 & 65.91 & \textbf{99.72} & 75.05 & 66.05 & 59.91 & 70.06 & 61.88 & 63.69\\
				 & MAvG & \textbf{5.79} & 0.00 & 0.00 & 5.40 & 0.00 & 5.43 & 4.25 & 5.25 & 5.14 & 5.19 & 5.39\\
				 & AvFb & 41.94 & 49.97 & 49.97 & 36.09 & \textbf{49.97} & 40.24 & 35.74 & 33.26 & 37.94 & 34.17 & 35.07\\
				 & CBA & 39.59 & 49.86 & 49.86 & 33.14 & \textbf{49.86} & 37.74 & 33.16 & 30.12 & 35.22 & 31.11 & 32.03\\

				Poker: 1 vs 2 & AvAcc & 51.13 & 89.87 & 89.87 & 53.05 & \textbf{89.87} & 61.23 & 57.24 & 52.16 & 60.17 & 51.78 & 52.36\\
				 & MAvG & 30.24 & 0.00 & 0.00 & 29.97 & 0.00 & 30.20 & 30.04 & 30.18 & \textbf{30.55} & 30.34 & 30.39\\
				 & AvFb & 41.96 & 48.90 & 48.90 & 42.59 & \textbf{48.90} & 45.72 & 44.34 & 42.37 & 45.79 & 42.33 & 42.62\\
				 & CBA & 30.78 & 44.94 & 44.94 & 31.95 & \textbf{44.94} & 37.12 & 34.58 & 31.42 & 36.50 & 31.20 & 31.57\\

				Poker: 1 vs 3 & AvAcc & 53.38 & 95.24 & 95.24 & 55.58 & \textbf{95.24} & 59.98 & 62.62 & 52.58 & 59.94 & 53.00 & 53.45\\
				 & MAvG & 21.40 & 0.00 & 0.00 & 21.50 & 0.00 & 21.36 & \textbf{21.87} & 21.40 & 21.60 & 21.34 & 21.28\\
				 & AvFb & 37.94 & 49.51 & 49.51 & 39.01 & \textbf{49.51} & 40.67 & 42.39 & 37.56 & 41.07 & 37.72 & 37.88\\
				 & CBA & 29.26 & 47.62 & 47.62 & 30.48 & \textbf{47.62} & 32.90 & 34.43 & 28.82 & 32.91 & 29.04 & 29.28\\

				Poker: 1 vs 4 & AvAcc & 68.29 & 99.09 & 99.09 & 63.78 & \textbf{99.09} & 69.98 & 98.25 & 61.60 & 64.87 & 62.69 & 63.53\\
				 & MAvG & \textbf{12.48} & 0.00 & 0.00 & 12.43 & 0.00 & 11.94 & 7.61 & 12.19 & 12.08 & 12.30 & 12.41\\
				 & AvFb & 39.99 & 49.91 & 49.91 & 37.89 & 49.91 & 40.43 & \textbf{50.22} & 36.75 & 38.22 & 37.31 & 37.77\\
				 & CBA & 34.99 & 49.54 & 49.54 & 32.67 & 49.54 & 35.81 & \textbf{49.63} & 31.54 & 33.21 & 32.11 & 32.54\\

				Poker: 1 vs 5 & AvAcc & 57.04 & 99.53 & 99.53 & 59.41 & \textbf{99.53} & 61.85 & 60.94 & 57.00 & 58.39 & 58.71 & 59.58\\
				 & MAvG & 6.42 & 0.00 & 0.00 & 6.15 & 0.00 & \textbf{7.54} & 7.21 & 6.65 & 6.57 & 6.62 & 6.61\\
				 & AvFb & 32.22 & 49.95 & 49.95 & 33.25 & \textbf{49.95} & 34.86 & 34.32 & 32.27 & 32.91 & 33.08 & 33.49\\
				 & CBA & 28.77 & 49.77 & 49.77 & 29.96 & \textbf{49.77} & 31.25 & 30.77 & 28.76 & 29.46 & 29.62 & 30.06\\

				Poker: 1 vs 6 & AvAcc & 81.85 & 99.66 & 99.66 & 61.75 & \textbf{99.66} & 72.01 & 56.74 & 57.65 & 76.20 & 59.10 & 60.42\\
				 & MAvG & \textbf{5.97} & 0.00 & 0.00 & 5.36 & 0.00 & 5.70 & 5.41 & 5.43 & 5.51 & 5.46 & 5.41\\
				 & AvFb & 43.37 & 49.97 & 49.97 & 34.16 & \textbf{49.97} & 38.95 & 31.79 & 32.23 & 40.78 & 32.93 & 33.54\\
				 & CBA & 41.21 & 49.83 & 49.83 & 31.07 & \textbf{49.83} & 36.24 & 28.55 & 29.01 & 38.34 & 29.74 & 30.40\\

				SEER & AvAcc & 88.63 & \textbf{90.90} & 89.47 & 90.05 & 89.96 & 88.46 & 89.26 & 89.54 & 90.75 & 89.50 & 90.17\\
				 & MAvG & 34.56 & 0.00 & 0.00 & 34.84 & 38.99 & 21.18 & 37.49 & 38.21 & \textbf{41.09} & 38.12 & 36.85\\
				 & AvFb & 34.90 & 31.56 & 19.48 & 35.71 & 38.88 & 26.51 & 37.96 & 38.29 & \textbf{39.56} & 38.17 & 36.21\\
				 & CBA & 29.26 & 25.75 & 12.58 & 31.24 & 33.37 & 22.33 & 32.75 & 32.26 & \textbf{33.73} & 32.16 & 31.29\\

				SSL Renegotiation & AvAcc & 83.45 & 83.46 & 83.46 & 83.49 & 83.46 & \textbf{92.47} & 83.46 & 83.46 & 83.46 & 83.46 & 83.47\\
				 & MAvG & 44.61 & 44.41 & 44.41 & \textbf{44.85} & 44.66 & 24.03 & 44.67 & 44.67 & 44.67 & 44.67 & 44.75\\
				 & AvFb & 70.30 & 70.02 & 70.01 & \textbf{70.61} & 70.35 & 50.08 & 70.37 & 70.37 & 70.37 & 70.37 & 70.47\\
				 & CBA & 51.38 & 51.33 & 51.33 & \textbf{51.48} & 51.40 & 49.31 & 51.40 & 51.40 & 51.40 & 51.40 & 51.43\\

				SUSY: 4:1 & AvAcc & 73.41 & 79.15 & 80.00 & 73.84 & \textbf{82.86} & 70.92 & 77.25 & 74.38 & 77.62 & 73.79 & 74.29\\
				 & MAvG & 59.83 & 65.02 & 17.89 & 60.28 & \textbf{75.34} & 53.61 & 62.61 & 60.55 & 63.15 & 60.10 & 60.44\\
				 & AvFb & 67.89 & \textbf{69.00} & 47.62 & 68.30 & 62.66 & 61.03 & 68.29 & 68.30 & 68.71 & 68.07 & 68.22\\
				 & CBA & 57.69 & \textbf{66.89} & 40.00 & 58.21 & 56.00 & 55.28 & 63.59 & 59.01 & 64.11 & 58.19 & 58.90\\

				SUSY: 8:1 & AvAcc & 77.05 & 88.94 & 88.89 & 74.30 & \textbf{89.31} & 73.79 & 81.28 & 74.43 & 81.39 & 74.16 & 74.91\\
				 & MAvG & 49.78 & \textbf{79.68} & 0.00 & 48.32 & 70.47 & 44.69 & 52.66 & 48.44 & 52.93 & 48.24 & 48.64\\
				 & AvFb & 64.60 & 49.18 & 48.78 & 63.69 & 60.77 & 59.98 & 65.41 & 63.80 & \textbf{65.65} & 63.62 & 63.94\\
				 & CBA & 52.83 & 44.79 & 44.44 & 50.15 & 56.62 & 49.05 & 57.46 & 50.27 & \textbf{57.64} & 50.01 & 50.72\\

                SUSY: 16:1 & AvAcc & 78.56 & 94.12 & \textbf{94.12} & 74.92 & 93.82 & 76.22 & 84.98 & 75.52 & 84.74 & 74.71 & 75.37\\
				 & MAvG & 38.78 & 19.40 & 0.00 & 37.01 & \textbf{64.66} & 33.59 & 42.98 & 37.49 & 42.81 & 37.03 & 37.34\\
				 & AvFb & 60.31 & 49.40 & 49.38 & 58.31 & 59.44 & 55.38 & \textbf{62.99} & 58.86 & 62.95 & 58.31 & 58.70\\
				 & CBA & 47.63 & 47.08 & 47.06 & 44.85 & \textbf{56.87} & 44.94 & 53.22 & 45.35 & 52.99 & 44.72 & 45.21\\

				SYN DOS & AvAcc & 70.49 & 79.39 & 79.39 & 83.63 & 70.51 & 95.65 & \textbf{99.70} & 70.46 & 70.46 & 70.46 & 99.63\\
				 & MAvG & 8.93 & 10.93 & 10.97 & 18.58 & 9.22 & 58.71 & \textbf{60.21} & 9.22 & 9.23 & 9.22 & 48.78\\
				 & AvFb & 39.36 & 44.23 & 44.25 & 48.08 & 39.49 & 58.11 & \textbf{61.29} & 39.46 & 39.47 & 39.46 & 60.58\\
				 & CBA & 35.62 & 40.27 & 40.27 & 44.75 & 35.64 & 56.23 & \textbf{60.32} & 35.62 & 35.62 & 35.62 & 60.08\\

				Traffic & AvAcc & 76.86 & 78.77 & 78.56 & 76.43 & \textbf{79.10} & 77.04 & 77.84 & 76.91 & 78.99 & 76.79 & 77.04\\
				 & MAvG & 20.10 & \textbf{22.06} & 0.00 & 17.66 & 20.85 & 16.31 & 19.61 & 20.72 & 20.71 & 20.58 & 20.25\\
				 & AvFb & 18.06 & \textbf{21.64} & 9.14 & 15.77 & 21.14 & 16.10 & 20.07 & 18.66 & 19.54 & 18.46 & 16.89\\
				 & CBA & 14.52 & 17.67 & 3.77 & 13.32 & \textbf{18.49} & 13.21 & 17.32 & 14.77 & 15.54 & 14.51 & 12.98\\

				Video Injection & AvAcc & 43.72 & 95.85 & \textbf{95.85} & 75.78 & 84.64 & 86.31 & 95.00 & 82.59 & 88.51 & 82.71 & 94.60\\
				 & MAvG & 16.25 & 0.00 & 0.00 & 37.80 & 41.77 & 48.68 & \textbf{66.13} & 40.04 & 47.86 & 40.17 & 64.62\\
				 & AvFb & 30.01 & 49.57 & 49.57 & 56.03 & 66.34 & 55.54 & \textbf{81.39} & 64.77 & 71.57 & 64.94 & 79.86\\
				 & CBA & 23.40 & 47.93 & 47.93 & 45.52 & 51.39 & 50.72 & \textbf{70.06} & 49.53 & 56.11 & 49.65 & 69.07\\

				\bottomrule
			\end{tabular}
		}
	\end{adjustbox}
\end{table}

\begin{figure}
	\includegraphics[width=16cm]{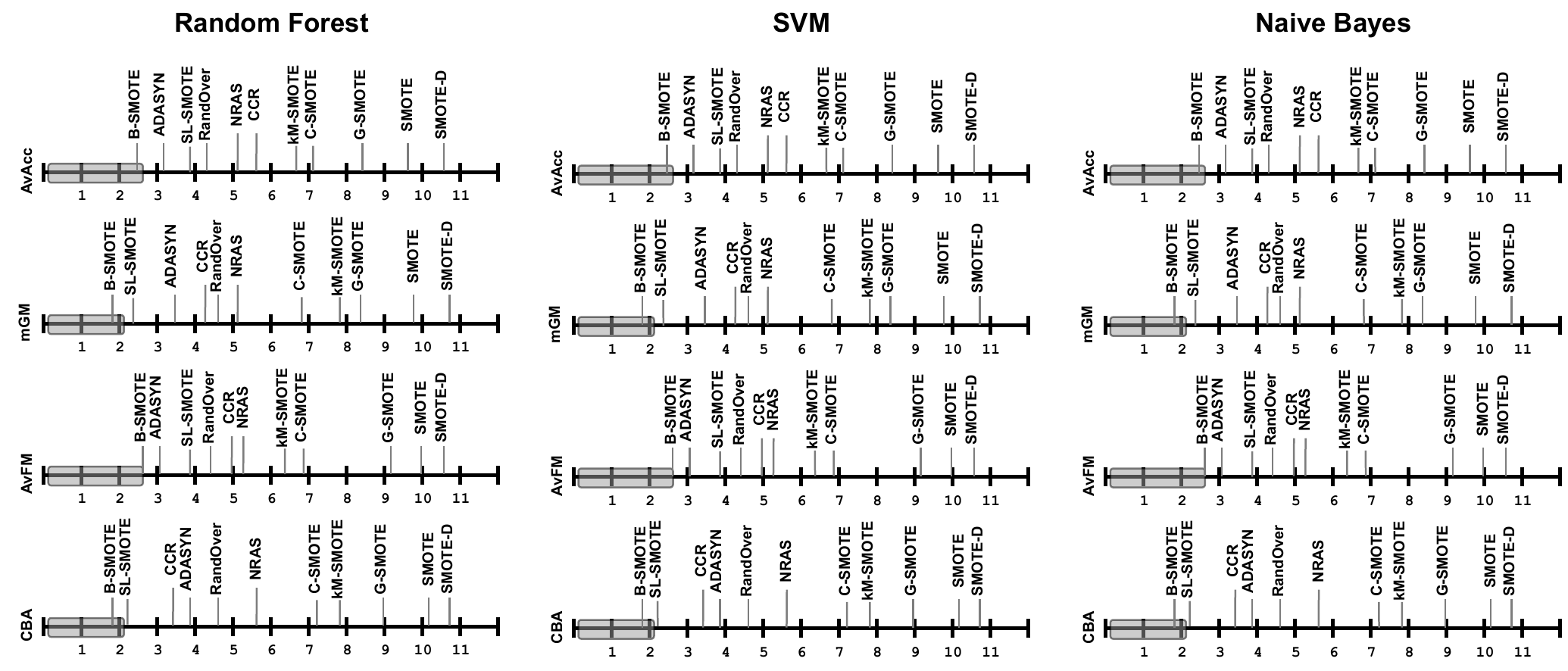}
	\caption{The Bonferroni--Dunn test for comparison among examined oversampling methods with respect to evaluation metric and used classifier.}
	\label{fig:ranks}
\end{figure}

\begin{figure}[h]
\includegraphics[width=16cm]{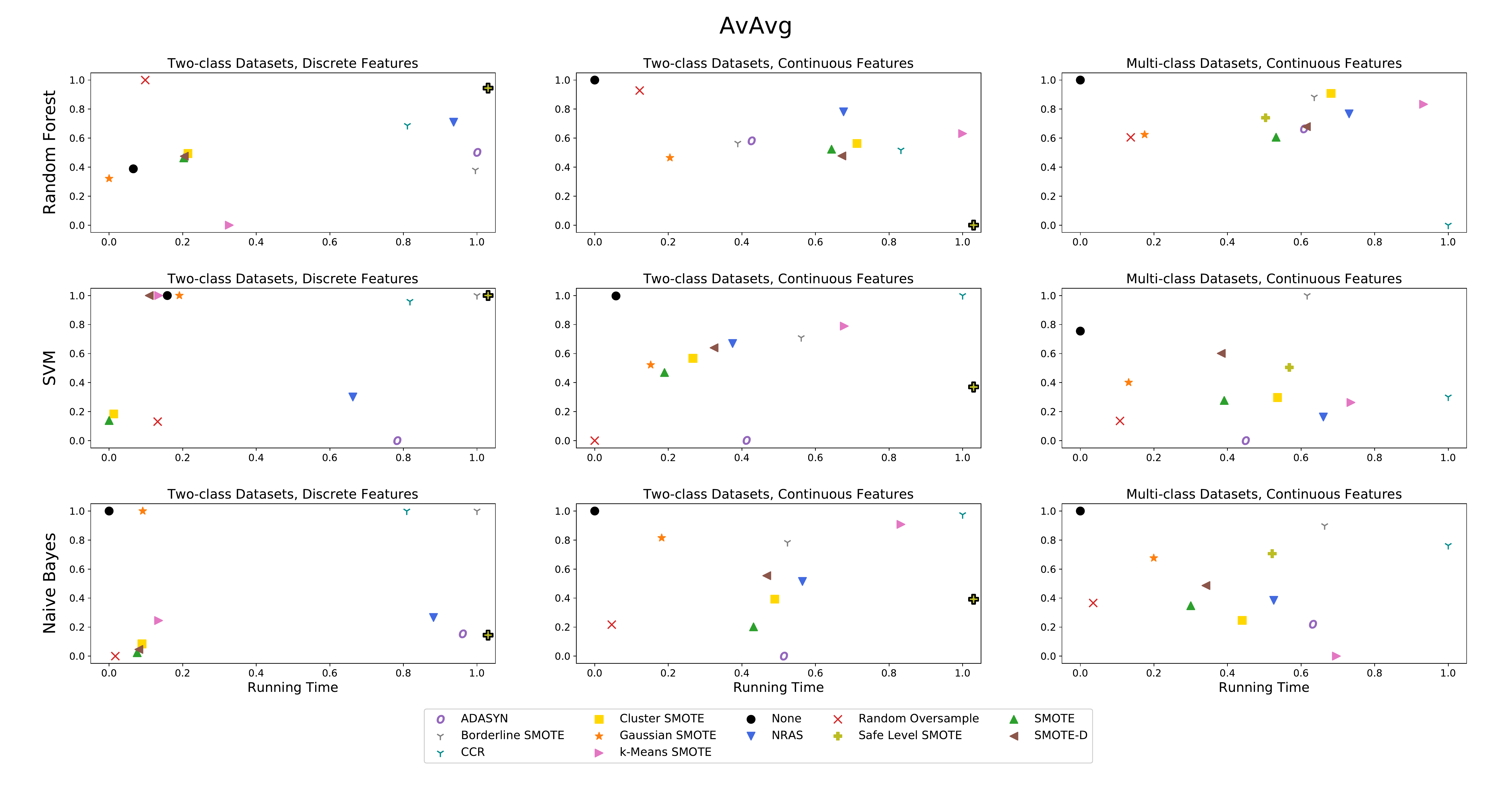}
\caption{The relative run times and AvAvg scores for the different dataset groups, classifiers and sampling methods.}
\label{fig:AvAvg}
\end{figure}

\begin{figure}[h]
\includegraphics[width=16cm]{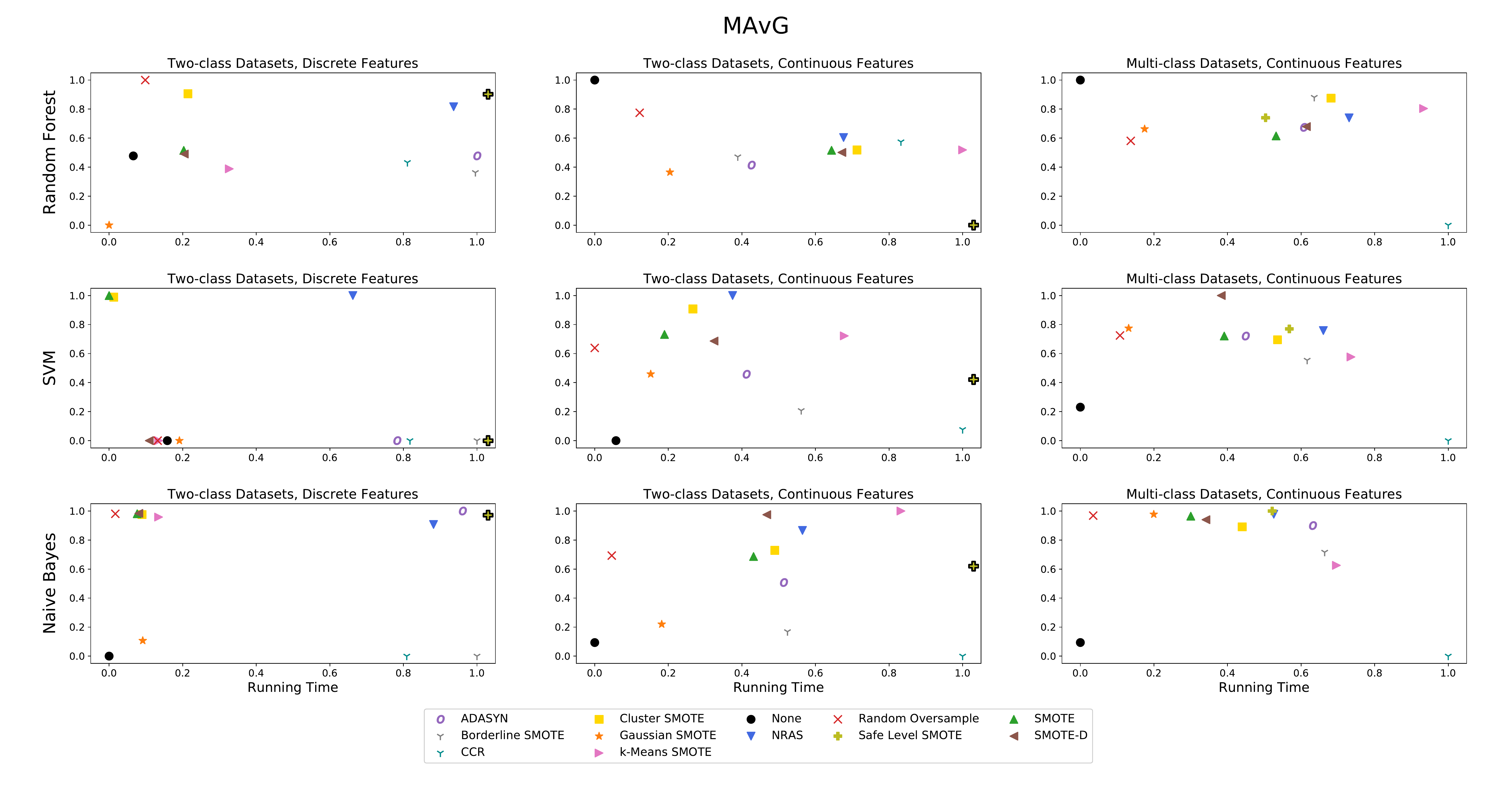}
\caption{The relative run times and MAvG scores for the different dataset groups, classifiers and sampling methods.}
\label{fig:MAvG}
\end{figure}

\begin{figure}[h]
\includegraphics[width=16cm]{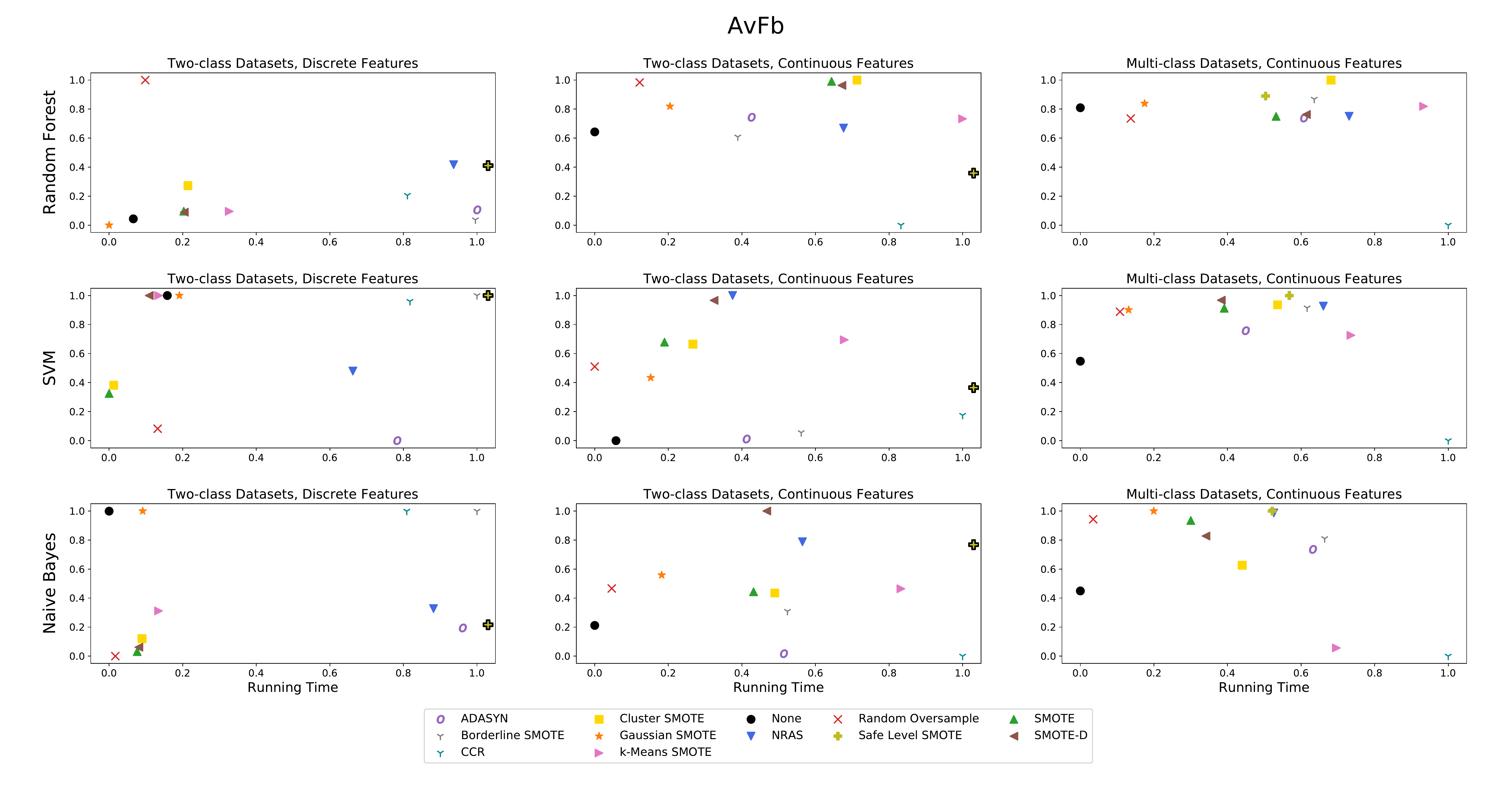}
\caption{The relative run times and AvFb scores for the different dataset groups, classifiers and sampling methods.}
\label{fig:AvFb}
\end{figure}

\begin{figure}[h]
\includegraphics[width=16cm]{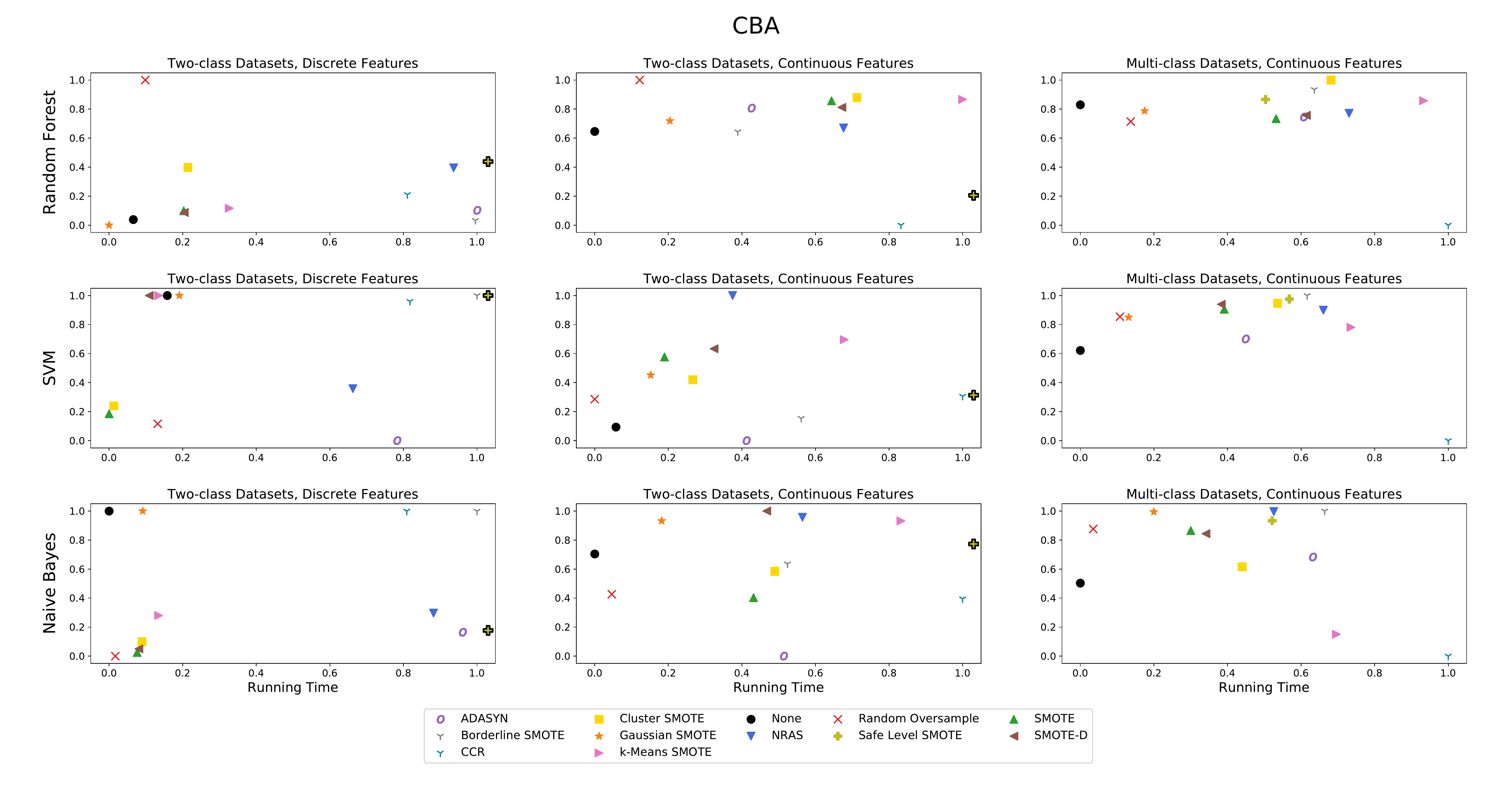}
\caption{The relative run times and CBA scores for the different dataset groups, classifiers and sampling methods.}
\label{fig:cba}
\end{figure}

\begin{figure}[h]
\includegraphics[width=16cm]{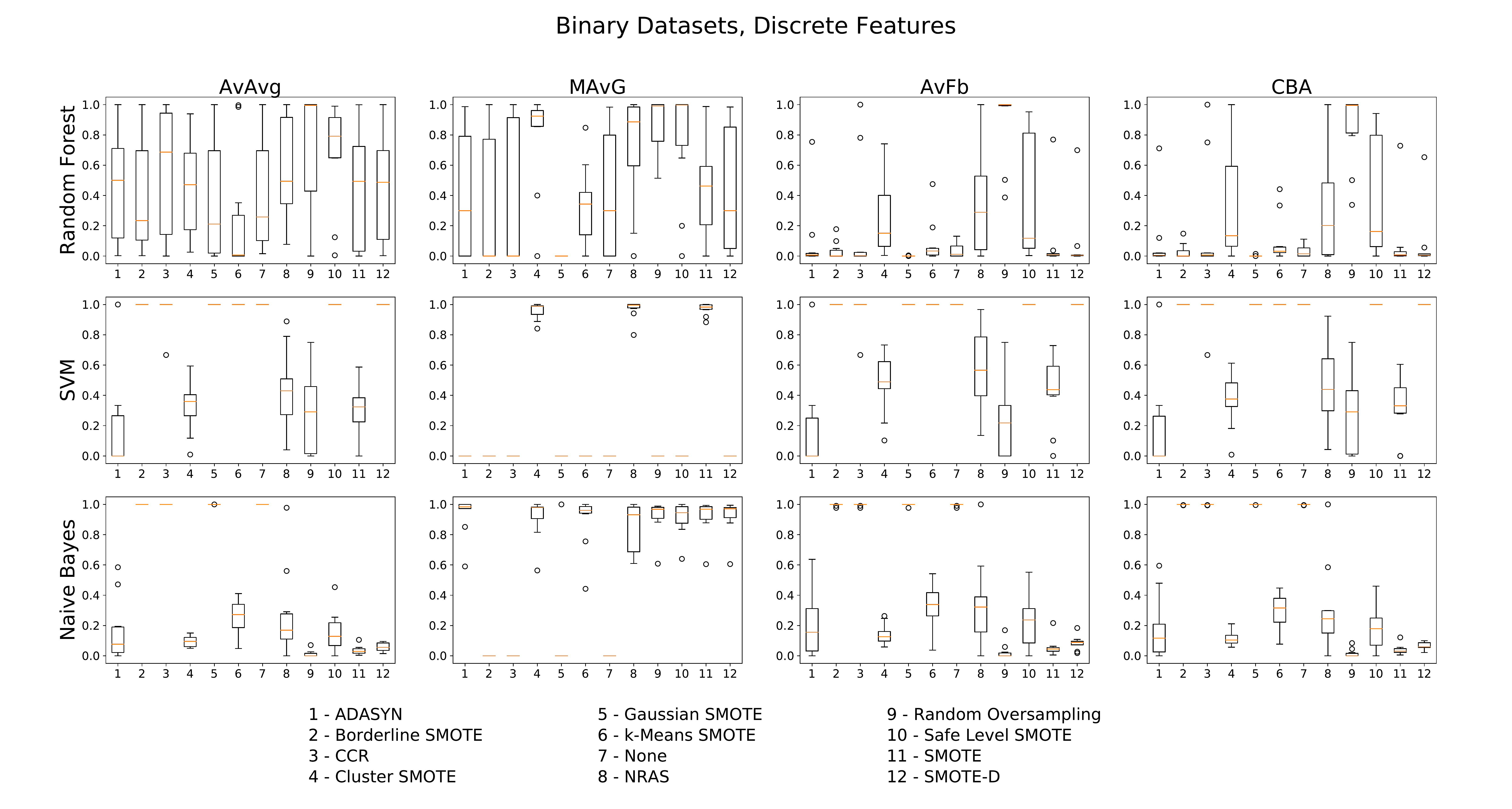}
\caption{The distribution of metric scores for each sampling method and classifier using the Binary Datasets with Discrete Features.}
\label{fig:bd}
\end{figure}

\begin{figure}[h]
\includegraphics[width=16cm]{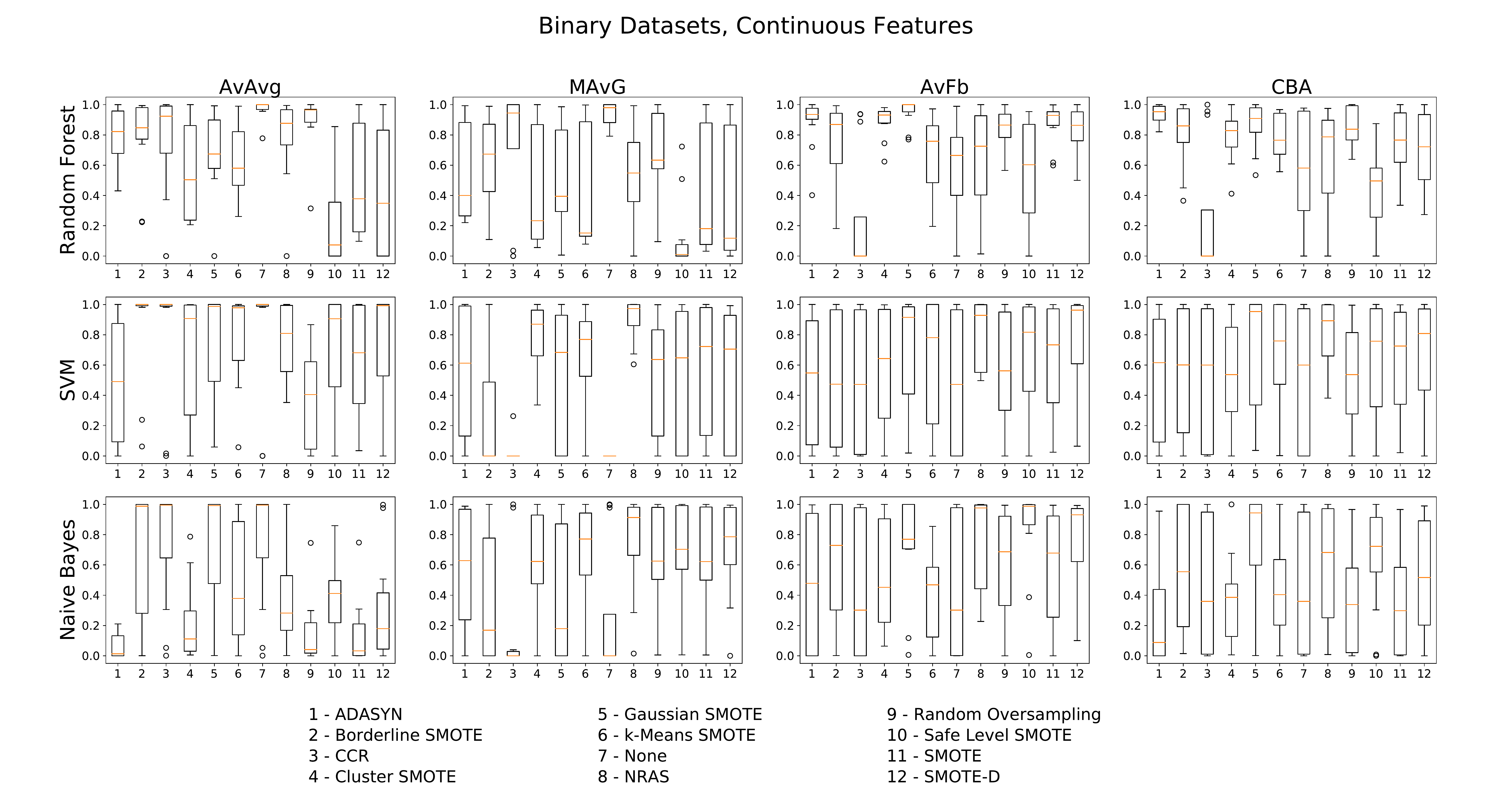}
\caption{The distribution of metric scores for each sampling method and classifier using the Binary Datasets with Continuous Features.}\label{fig:bc}
\end{figure}

\begin{figure}[h]
\includegraphics[width=16cm]{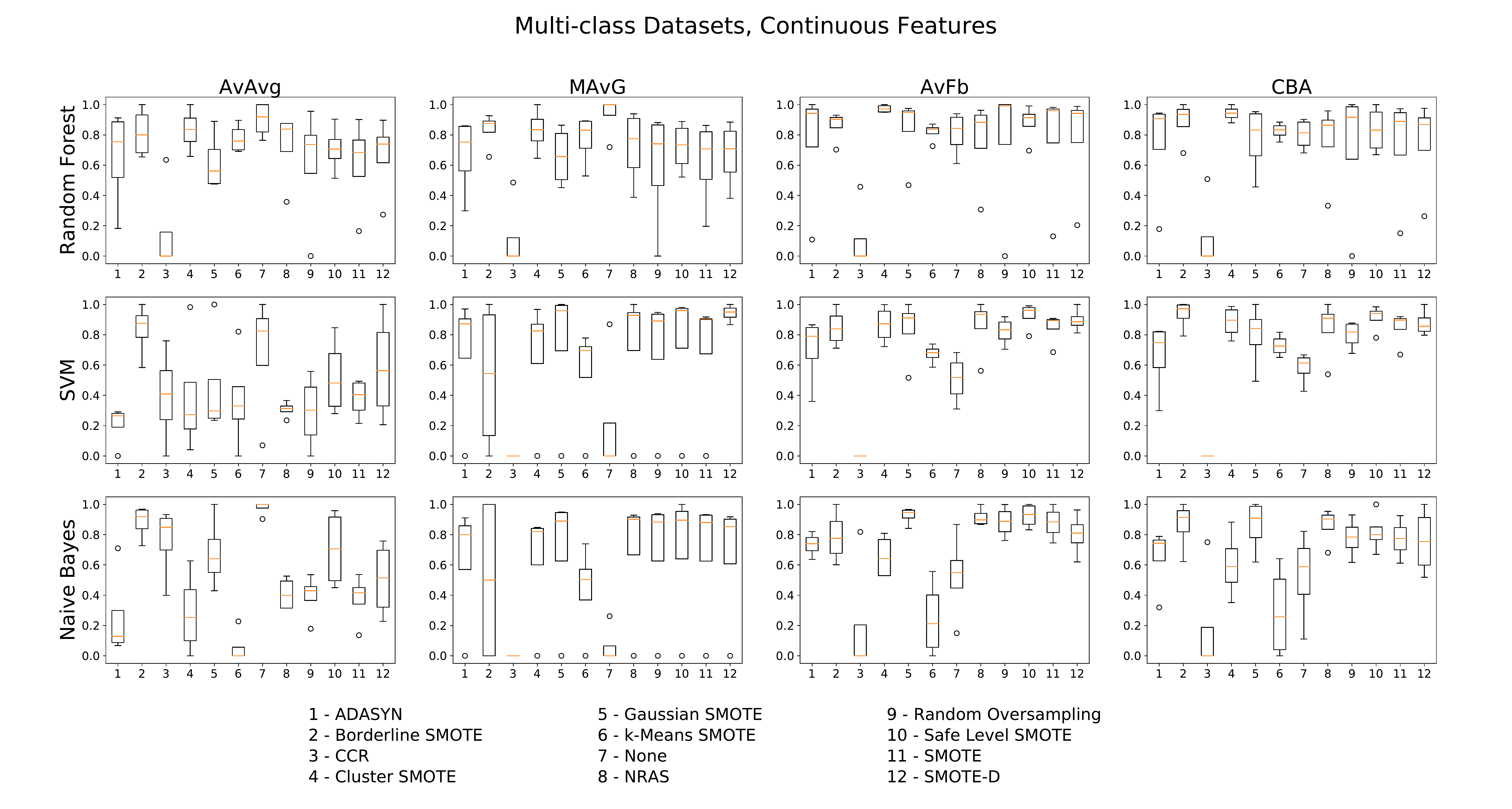}
\caption{The distribution of metric scores for each sampling method and classifier using the Mutli-class Datasets with Continuous Features.}\label{fig:mc}
\end{figure}

\subsection{Experiment 1: Oversampling algorithms for binary big imbalanced data}
\label{sec:exp1}

\noindent \textbf{Overview of learning difficulties.} Binary big imbalanced data has well-defined class roles, hence the learning difficulty comes from the imbalanced ratio combined with the volume of each class and potential presence of instance-level difficulties, such as borderline instances, subconcepts, multiple modalities, or class overlap. Furthermore, due to the distributed nature of the Spark system, an additional challenge may arise when the minority class is significantly smaller (i.e., extreme imbalance ratio) than the majority class. In such a case, the minority class will be further limited in size by being distributed among multiple nodes, leading to the potential problem of learning from a small sample size. Therefore, oversampling methods are a natural choice for tackling such problems, by overcoming limitations in minority class size in each Spark node.  

\smallskip
 \noindent \textbf{Informed vs standard oversampling.} As discussed in previous sections, all the proposed oversampling methods for Spark can be analyzed from the point of how much information regarding class/instance distribution they use. Random Oversampling is the extreme example of using no such information, SMOTE uses only the neighborhood information in a basic way, while methods like CCR or RBO rely on an extensive analysis of instance-level properties. While intuition suggests that informed oversampling will always outperform standard approaches, our results show that this is not always the case. Random oversampling is capable of outperforming multiple informed variants, such as all SMOTE modifications based on clustering or Gaussian spread data. This shows that not all of information about instances is equally important when utilized on the Spark architecture. As Spark uses its own data partitioning algorithm when creating data subsets for each node, the spatial relationship among instances is affected. Thus, oversampling methods based on introducing instances within each cluster / Gaussian mode cannot work efficiently and random oversampling can outperform them. However, oversampling methods that use more advanced information regarding neighborhood of each instance display the best possible performance. This proves that local information is much more useful to big data oversampling than modality-based information for each class.
 
 \smallskip
\noindent \textbf{Comparison among informed oversampling methods.} The four best performing informed oversampling methods, regardless of analyzed metric, are Borderline SMOTE, Safe-level SMOTE, ADASYN, and CCR. All those methods use information about local data difficulty factors, such as class overlapping or homogeneity of the neighborhood, to introduce new artificial instances. Differences in performance of these algorithms on big binary datasets are small and depend on specific characteristics of the considered data set. Borderline SMOTE performs best in cases when there is a high class overlapping and them minority class needs to be reinforced around the uncertainty region (i.e., decision boundary). Safe-level SMOTE performs best for datasets where the disproportion between classes is the main source of learning difficulty. Safe-level SMOTE reinforces homogeneous regions of minority class, leading to balancing in the instance quantity without affecting uncertainty regions (as new instances are introduced in highly certain regions of minority class). ADASYN and CCR offer excellent mechanisms for dealing with rare instances and outliers, hence performing best on difficult datasets with high number of small subconcepts or atypical observations. CCR further augments its performance by cleaning the overlapping regions from majority class observations, leading to excellent performance on small disjuncts. 

 \smallskip
\noindent \textbf{Impact of base classifier selection.} Interestingly, all three base classifiers (Random Forest, SVM, and Naïve Bayes) offer similar correlation with which oversampling method works the best. Random Forest can utilize random oversampling more efficiently, as it can be combined with its ensemble structure to offer improved diversity. Single-model classifiers (SVM and Naïve Bayes) benefit from using guided oversampling much more significantly, as this leads to better separation margins or class probability estimations. In summary, four guided oversampling methods highlighted in the previous point (Borderline-SMOTE, Safe-level SMOTE, ADASYN, and CCR) offer excellent and robust performance regardless of with which classifier they are being paired.

\subsection{Experiment 2: Oversampling algorithms for multi-class big imbalanced data}
\label{sec:exp2}

\noindent \textbf{Differences between binary and multi-class problems.} Multi-class imbalanced problems pose a variety of unique challenges to oversampling algorithms. Our software package offers universal implementations of discussed oversampling methods that can work with binary and multi-class massive datasets. While there are oversampling solutions dedicated specifically to multi-class problems, they usually are characterized by a prohibitive computational cost when dealing with big data (e.g., MC-RBO, extension of RBO). All proposed algorithms have been adapted with one-vs-all approach, allowing them to leverage global and local data properties, without intra-class relationships becoming a choking point. We can observe similar trends with binary data, where informed oversampling approaches significantly outperformed other solutions, although here differences are even more pronounced. This shows that the importance of guided oversampling becomes even more crucial when dealing with multiple classes.

\smallskip
\noindent \textbf{Role of informative oversampling.} While most of the examined oversampling algorithms behave similarly to their binary counterparts, we can see improvement in the performance of Cluster SMOTE. This is the only global informative oversampling that returns predictive accuracy like its local informative counterparts. We can explain this by the fact that in case of multiple classes we are more likely to deal with more complex problems, where modalities in each class are more pronounced. Additionally, by using one-vs-all approach, we create a super-class of majority instances that consists of multiple joint classes. This scenario allows Cluster SMOTE to create pairwise oversampling areas, detecting multiple potential decision boundaries. While its performance is significantly better than in the binary case, local methods such as Borderline SMOTE, ADASYN, and CCR are still the best performing ones. This shows that the importance of local instance-level information is as important in multi-class scenarios as it is in binary ones. 

\smallskip
\noindent \textbf{Behavior of base classifiers in multi-class problems.} While in binary scenarios, all three classifiers returned similar synergies with oversampling methods and we can see that Random Forest and SVM behave differently from Naive Bayes. This is because Random Forest is an explicit ensemble and SVM is an implicit ensemble for multi-class problems (as we train them in one-vs-all mode). Therefore, those compound classifiers are capable of better generalization over multiple skewed distributions, leveraging the additional information provided by oversampling algorithms.

\subsection{Experiment 3: Investigating oversampling trade-off between accuracy and time complexity}
\label{sec:exp3}

When selecting an oversampling algorithm for a given big data problem, it is important to realize the existence of a trade-off between accuracy and run time. We believe that the end user should be able to chose any of the oversampling algorithms present in our software package based on their needs and available resources. To this end, we present the visualizations of the relationships between the analyzed oversampling method accuracy (according to four metrics) and computational time in Figures \ref{fig:AvAvg} - \ref{fig:cba}.

As shown in these results, the combination of data properties and the target classifier may by the deciding factor when choosing a class balancing algorithm. For example, Random Oversampling produced some of the best trade-offs for the Two-Class datasets with the Random Forest classifier but did not perform as well for the Multi-class datasets. In addition, it often performed quite poorly with the SVM and Naive Bayes classifiers on all three dataset groups. Algorithms such as Borderline SMOTE, Cluster SMOTE, Gaussian SMOTE and NRAS showed to have some of the best accuracy and time trade-offs for certain dataset and classifier combinations but no algorithm was the clear overall winner. 

\subsection{Experiment 4: Scalability of oversampling algorithms}

\begin{figure}
	\includegraphics[width=16cm]{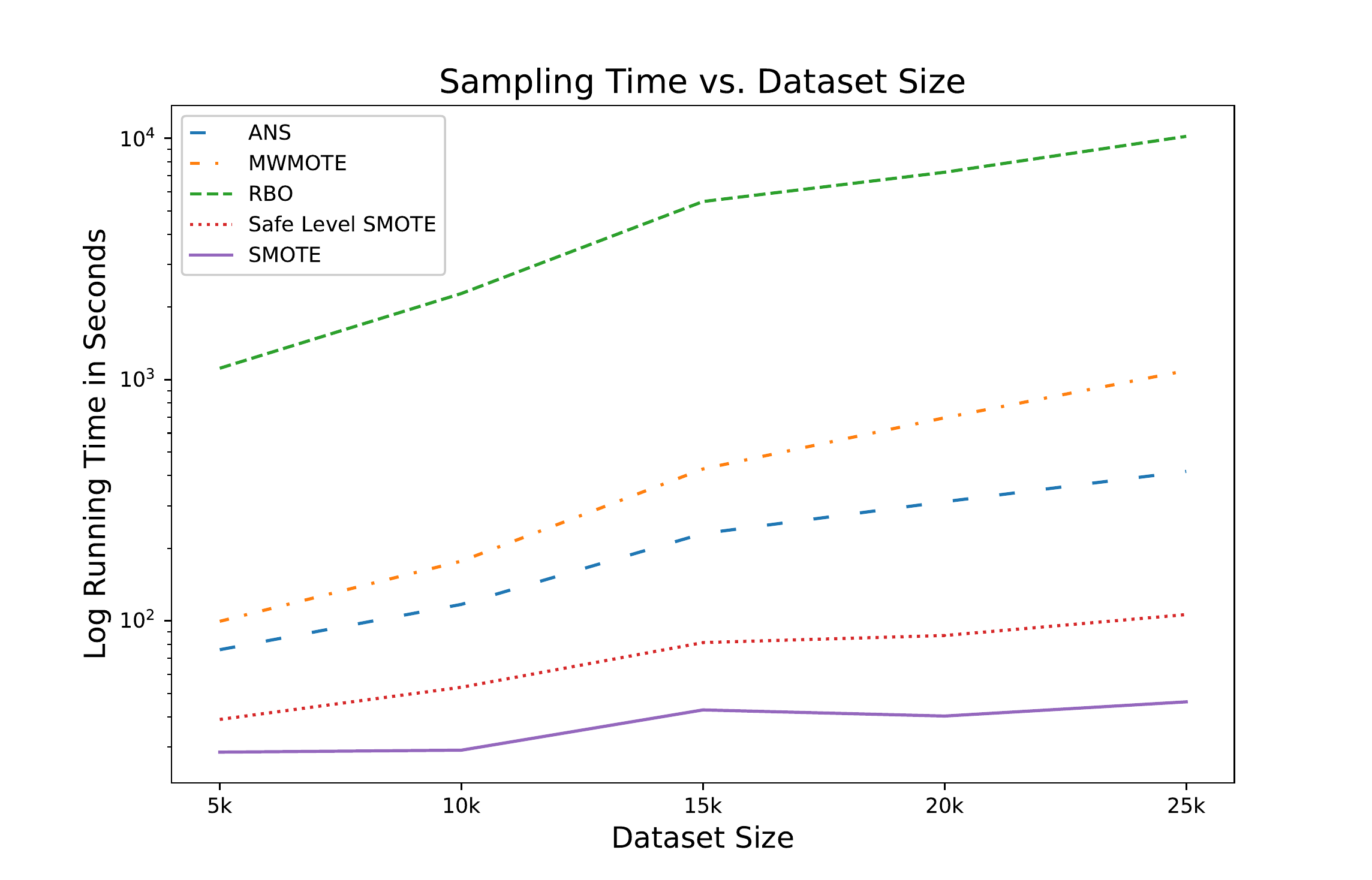}
	\caption{Sampling times for increasing dataset sizes with five sampling methods on the Covertype dataset.}
	\label{fig:slowmethods}
\end{figure}

In addition to the twelve methods presented so far, three other sampling algorithms were also implemented: ANS, MWMOTE, RBO. As these methods have shown competitive results in their original publications and received interested from other researchers, they were added to our slate of oversampling methods. However, during the implementation and initial testing of these methods, it became apparent 
that their lack of scalability made them unsuitable for our main experiments.

To illustrate this issue, Figure \ref{fig:slowmethods} shows the running times for several small subsets of the Covertype dataset with ANS, MWMOTE, RBO and two reference methods of SMOTE and Safe Level SMOTE. Like our main experiments, these were performed with 5-fold cross validation so in each run only 80\% of the data was used for oversampling. The three slower method took significantly longer than the reference methods and required a log scale for the time axis in order to reasonably present these results.

Using linear regression, the sampling times were estimated for using a dataset of 100,000 examples as shown in Table \ref{table:slowprojection}. Compared to standard SMOTE, the others methods were significantly slower with the following approximate time increases: ANS (15x), MWMOTE (40x), RBO (400x).  While ANS and MWMOTE may be feasible for 100,000 example datasets, it was not practicable for this projects as we used 5-fold cross validation and 26 different datasets, resulting in 130 individual experiments per sampling method.

\begin{table}
	\centering
	\renewcommand{\arraystretch}{1.25}
	\begin{adjustbox}{center,max width = 135mm}
		\footnotesize
		\begin{tabular}{ccccc}\hline
			\multicolumn{5}{c}{\textbf{Projected Sampling Times}}\\
			\bottomrule
			\textbf{SMOTE} & \textbf{Safe Level SMOTE} & \textbf{ANS} & \textbf{MWMOTE} & \textbf{RBO}\\
			\bottomrule
			116  & 359 & 1,718 & 4,744 & 44,540\\
			\bottomrule
		\end{tabular}
	\end{adjustbox}
	\caption{Projected sampling times in seconds for oversampling the Covtype dataset with 100k examples.}
	\label{table:slowprojection}
\end{table}

\section{Guidelines for designing oversampling algorithms for imbalanced big data}
\label{sec:fut}

\subsection{Insight into commonly used algorithmic components}
Each of the implemented oversampling methods were based on one or more commonly used algorithmic components. In this section, we provide guidelines on the appropriateness of these components in distributed algorithms designed for processing large datasets. Table \ref{table:algorithmComponents} shows a list of the components present in each oversampling method, sorted by approximate run time (fastest to slowest).

\smallskip
\noindent\textbf{k-NN.} Although the \textit{k}-NN algorithm is included in many machine learning libraries, it is not currently part of the official Spark MLlib package. For this reason, we have used the efficient distributed implementation from spark-knn \cite{fang2015} which is based on a hybrid spill tree. Since \textit{k}-NN has an approximate asymptotic complexly of $\mathcal{O}(n\log n)$, it is practical to be used for large datasets as shown in our results. However, including multiple \textit{k}-NN models can noticeably increase run time and should only be done when truly necessary. In some cases, it may be possible to avoid multiple \textit{k}-NNs by using one model with a larger \textit{k} value and filter the results for use in different sub-tasks.

\smallskip
\noindent\textbf{k-Means.} The \textit{k}-Means clustering algorithm is part of Spark MLlib and is optimized for distributed tasks making it straightforward to use. While \textit{k}-Means has a higher theoretical asymptotic complexity then \textit{k}-NN, our results confirm that the Spark MLlib implementation can run efficiently in practice. When using \textit{k}-Means as part of an oversampling method, the optimal results and run time may be dependent on the number of clusters and total iterations. Discovering these exact values may require an expensive hyperparameter search.

\smallskip
\noindent\textbf{Dependent Loop.} A dependent loop occurs when each iteration requires results from a previous iteration, preventing trivial parallelism. While this approach can work well in serial algorithms, it becomes problematic with distributed computing and is incomparable with the MapReduce design. Ideally, MapReduce decomposes tasks so that all independent work can be done in parallel. By adding a dependency in a loop, MapReduce cannot break down these tasks further and so each loop must be run in a serial manner. If there are more CPUs than loops, a number of these CPUs will go idle. If these loops are being run per training example, as it is the case for a number of the presented algorithms, there is no guarantee that each loop will take the same amount of time, potentially leading to significant bottlenecks. These dependent loops should be avoided whenever possible as they limit scalability and make run time less predictable.

\smallskip
\noindent\textbf{Probability.} Oversampling methods that use probability select examples based on rank ordering. Like dependent loops, this method does not translate well from serial to distributed environments. Probabilistic sampling requires the calculation of a probability/likelihood score, sorting the values and indexing the resulting list for example selection. With MapReduce, data is distributed across multiple computation nodes and the probability based methods require collecting all of the results back to the driver to ensure correct results which requires a serial step.

\smallskip
\noindent\textbf{Linear Regression.}
Like \textit{k}-Means, Spark MLlib includes linear regression so its use in oversampling methods is trivial. Although it is not prohibitively slow, including linear regression does add to the execution time and is one of the reasons NRAS is slower than similar algorithms.

\smallskip
\noindent\textbf{Neighbors by Radius.}
Many of the neighborhood calculations done with oversampling algorithms only consider a fixed number of neighbors as done with a \textit{k}-NN search. However, several of the presented algorithms instead need to find neighbors within a radius. One downside of using a radius based search is that there is no limit of how many neighbors may be returned. This can lead to potential bottlenecks with computational time and memory use as each computational node may be processing different amounts of data depending on the density of these radius based neighborhoods. This approach should be avoided if possible but could be somewhat mitigated by setting a hard limit on the number of neighbors to return.

\smallskip
\noindent\textbf{\bm{$\mathcal{O}(n^{2})$} Complexity.}
Using any sub-components that have an asymptotic complexity of $\mathcal{O}(n^{2})$ or worse should be avoided at all costs. While such methods may work for small datasets, the lack of scalability makes them unsuitable for big data problems.

\begin{table}
	\centering
	\renewcommand{\arraystretch}{1.25}
	\begin{adjustbox}{center,max width = 135mm}
		\footnotesize
		\begin{tabular}{lccccccc}\hline
			\multicolumn{7}{c}{\textbf{Algorithmic Components Present in Oversampling Methods}}\\
			\bottomrule
			\textbf{Sampling Method} & \textbf{\textit{k}NN Count} & \textbf{\textit{k}-Means} & \textbf{Dependent Loop} & \textbf{Probability} & \textbf{Linear Regression} & \textbf{Neighbors by Radius} & \bm{$\geq$ $\mathcal{O}(n^{2})$}  \\
			\bottomrule
			ADASYN & 1 & \ding{53} & \ding{53} & \ding{53} & \ding{53} & \ding{53} & \ding{53}\\
			SMOTE & 1 & \ding{53} & \ding{53} & \ding{53} & \ding{53} & \ding{53} &\ding{53} \\
			Gaussian SMOTE & 1 & \ding{53} & \ding{53} & \ding{53} & \ding{53} & \ding{53} & \ding{53}\\
			SMOTE-D & 1 & \ding{53} & \ding{53} & \ding{51} & \ding{53} & \ding{53} & \ding{53}\\
			NRAS & 1 & \ding{53} & \ding{53} & \ding{53} & \ding{51} & \ding{53} & \ding{53}\\
			Cluster SMOTE & 1 & \ding{51} & \ding{53} & \ding{53} & \ding{53} & \ding{53} & \ding{53}\\
			\textit{k}-Means SMOTE & 1 & \ding{51} & \ding{53} & \ding{53} & \ding{53} & \ding{53} & \ding{51} \\
			CCR & 1 & \ding{53} & \ding{51} & \ding{53} & \ding{53} & \ding{51} & \ding{53}\\
			Safe Level SMOTE & 2 & \ding{53} & \ding{53} & \ding{53} & \ding{53} & \ding{53} & \ding{53}\\
			Borderline SMOTE & 2 & \ding{53} & \ding{53} & \ding{53} & \ding{53} & \ding{53} & \ding{53}\\
			ANS & 3 & \ding{53} & \ding{51} & \ding{53} & \ding{53} & \ding{51} & \ding{53}\\
			MWMOTE & 3 & \ding{51} & \ding{53} & \ding{51} & \ding{53} & \ding{53} & \ding{51}\\
			RBO & 0 & \ding{53} & \ding{51} & \ding{53} & \ding{53} & \ding{53} & \ding{51}\\
			\bottomrule
		\end{tabular}
	\end{adjustbox}
	\caption{A listing of high level algorithmic components present in each oversampling method. The list is sorted by approximate running time, fastest to slowest.}
	\label{table:algorithmComponents}
\end{table}

\subsection{Design choices and future directions for big data oversampling algorithms.}

\noindent\textbf {Local over global data characteristics.} Informative oversampling methods can be divided into two groups: (i) ones utilizing global data characteristics; and (ii) ones utilizing local data characteristics. Global oversampling algorithms are usually based on finding modalities in data, such as subconcepts or underlying clusters. Local oversampling algorithms rely on analysis of the neighborhood of each minority instance and using this information to guide the artificial instance generation and placement. From our experimental study, we can see that global methods (such as Cluster SMOTE) returned highly unsatisfactory performance, while local methods were constantly among the best performing ones. We contribute this fact to the way Spark architecture creates data subsets for each node, leading to decreased importance of global and spatial data characteristics, while the local ones still hold. Therefore, we recommend to focus on local instance-level difficulties when designing novel oversampling algorithms for imbalanced big data. 

\smallskip
\noindent\textbf {Trade-off on instance-level characteristics analysis.} While local instance-level characteristics can be highly beneficial to the design of oversampling methods, there is a plethora of information that can be considered here. The more detailed information on instance neighborhoods, region homogeneity, or class overlap that is taken into consideration, leading to a higher potential for effective generation of artificial instances. However, extraction of such information over big data comes at the expense of a significant computational cost. This can be seen with RBO that uses advanced analysis of binary and multi-class potential generated by instances and classes to optimize the oversampling process. While RBO returns excellent performance, its complexity leads to a lack of scalability to big data, making it an unfeasible algorithm for Spark. Therefore, we recommend utilizing low-complexity analysis of local data properties. Even simple small neighborhood analysis or sketching of local distributions will lead to better predictive accuracy, while still being scalable to massive datasets. 

\smallskip
\noindent\textbf {Potential in hybrid solutions.} The experimental study presented in this paper, as well as our previous works and other findings from the literature, point to the superiority of oversampling to undersampling when handling imbalanced big data on the Spark architecture. These findings point to the fact that undersampling tends to remove important instances (especially for multi-class problems) and that the consolidation of node-based undersampled data does not lead to good generalization capabilities. However, one of the best performing methods considered in this study (CCR) employs a form of undersampling / data cleaning by translating majority observations. This allows us to conclude that while undersampling on its own may deliver underwhelming results, it potentially should be considered as a component for enhancing oversampling methods. Several works on small-scale datasets point out to the usefulness of oversampling for cleaning uncertainty regions and this approach can be effectively translated to big data scenarios. We recommend exploring the direction of hybrid oversampling that leverages undersampling / data cleaning steps to improve the empowerment of the minority classes. At the same time, we recommend using undersampling that considers local data characteristics, as it returns much more favorable results for the Spark architecture than random approach. 

\smallskip
\noindent\textbf {Challenging hyperparameter tuning.} Another potential bottleneck for oversampling methods lies in the difficulty of hyperparameter tuning for massive datasets. While using either grid search or one of the dedicated parameter tuning methods is possible on Spark, it relates to significant computational resource consumption. This can be prohibitive, especially with on-demand computing using commercially available Spark hardware. Let us again use RBO as an example – apart from its high computational complexity, it has a significant number of hyperparameters to be tuned. This is another potential choke point for deploying such an algorithm in real-world applications. Other oversampling methods presented in this study use much fewer parameters, making their tuning a more feasible task. We recommend designing oversampling algorithms for imbalanced big data with the smallest possible number of parameters (ideally non-parametric) or enhancing them with self-tuning solutions that can automatically adapt to the provided data without needing tedious user-guided tuning.

\section{Conclusion}
\label{sec:con}

In this paper, we have presented a holistic view on imbalanced big data oversampling. We have shown that the problem of learning from binary and multi-class skewed problems is highly pervasive in the big data domain, posing unique learning difficulties for machine learning systems. Existing oversampling methods cannot be directly utilized for high-performance architectures due to their specific computing requirements. We have shown how to adapt 14 popular oversampling algorithms to the Spark architecture, making them suitable for high-performance distributed computing. Even with these adaptations, we have observed that not all oversampling methods translate well to big data problems and the complexity of some of them becomes prohibitive and does not scale well with data set size. We identified crucial components of modern oversampling algorithms and showed their impact on the classification performance and time complexity. This allowed us to identify specific oversampling components that translate well to high-performance distributed architectures and that have desired properties from the predictive performance viewpoint. We have also created guidelines and future directions for designing novel oversampling algorithms for imbalanced big data that can be adopted by the research community. To facilitate future research and reproducibility in this domain, we have prepared an efficient library for Spark with implementations of the discussed oversampling methods.


\bibliographystyle{elsarticle-num-names}
\bibliography{references}   

\end{document}